\documentclass{article}

\usepackage{microtype}
\usepackage{graphicx}
\usepackage{subfigure}
\usepackage{booktabs} 

\usepackage{hyperref}

\usepackage[accepted]{icml2024}

\usepackage{amsmath}
\usepackage{amssymb}
\usepackage{mathtools}
\usepackage{amsthm}
\usepackage{bm}
\usepackage{multirow}
\usepackage{multicol}
\usepackage[capitalize,noabbrev]{cleveref}

\theoremstyle{plain}

\theoremstyle{definition}

\theoremstyle{remark}

\usepackage[textsize=tiny]{todonotes}
\pdfoutput=1

\icmltitlerunning{Stereo Risk: A Continuous Modeling Approach to Stereo Matching}

\begin{document}

\twocolumn[
\icmltitle{Stereo Risk: A Continuous Modeling Approach to Stereo Matching}



\icmlsetsymbol{equal}{*}

\begin{icmlauthorlist}
\icmlauthor{Ce Liu}{equal,nju,eth}
\icmlauthor{Suryansh Kumar}{equal,tam}
\icmlauthor{Shuhang Gu}{uestc}
\icmlauthor{Radu Timofte}{wuz}
\icmlauthor{Yao Yao}{nju}
\icmlauthor{Luc Van Gool}{eth,ku,ins}
\end{icmlauthorlist}

\icmlaffiliation{nju}{Nanjing University, China.}
\icmlaffiliation{eth}{ETH Z\"urich, Switzerland.}
\icmlaffiliation{tam}{VCCM, Texas A\&M University, USA.}
\icmlaffiliation{uestc}{UESTC, China.}
\icmlaffiliation{wuz}{University of W\"urzburg, Germany.}
\icmlaffiliation{ins}{INSAIT, Bulgaria}
\icmlaffiliation{ku}{KU Leuven, Belgium.}
\icmlcorrespondingauthor{Yao Yao}{yaoyao@nju.edu.cn}

\icmlkeywords{Stereo Matching, Continuous Disparity, Risk}

\vskip 0.3in
]



\printAffiliationsAndNotice{\icmlEqualContribution} 

\begin{abstract}

We introduce Stereo Risk, a new deep-learning approach to solve the classical stereo-matching problem in computer vision. As it is well-known that stereo matching boils down to a per-pixel disparity estimation problem, the popular state-of-the-art stereo-matching approaches widely rely on regressing the scene disparity values, yet via discretization of scene disparity values. Such discretization often fails to capture the nuanced, continuous nature of scene depth.
Stereo Risk departs from the conventional discretization approach by formulating the scene disparity as an optimal solution to a continuous risk minimization problem, hence the name ``stereo risk''. We demonstrate that $L^1$ minimization of the proposed continuous risk function enhances stereo-matching performance for deep networks, particularly for disparities with multi-modal probability distributions. Furthermore, to enable the end-to-end network training of the non-differentiable $L^1$ risk optimization, we exploited the implicit function theorem, ensuring a fully differentiable network. A comprehensive analysis demonstrates our method's theoretical soundness and superior performance over the state-of-the-art methods across various benchmark datasets, including KITTI 2012, KITTI 2015, ETH3D, SceneFlow, and Middlebury 2014.

\end{abstract}
\begin{figure}[t]
\centering
\subfigure[Image]{
\begin{minipage}[b]{0.22\textwidth}
\includegraphics[width=1.0\textwidth]{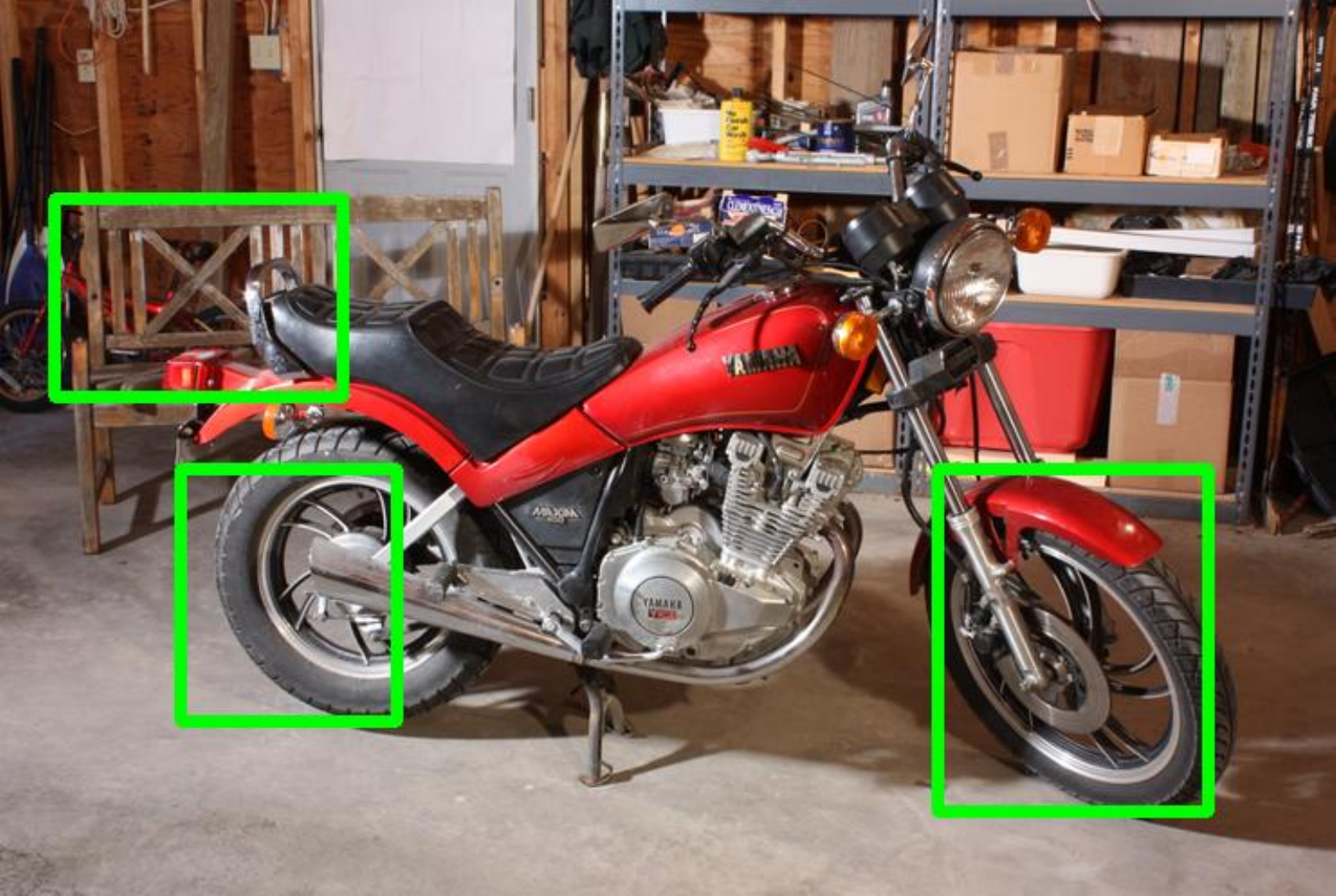}
\end{minipage}
}
\subfigure[IGEV]{
\begin{minipage}[b]{0.22\textwidth}
\includegraphics[width=1.0\textwidth]{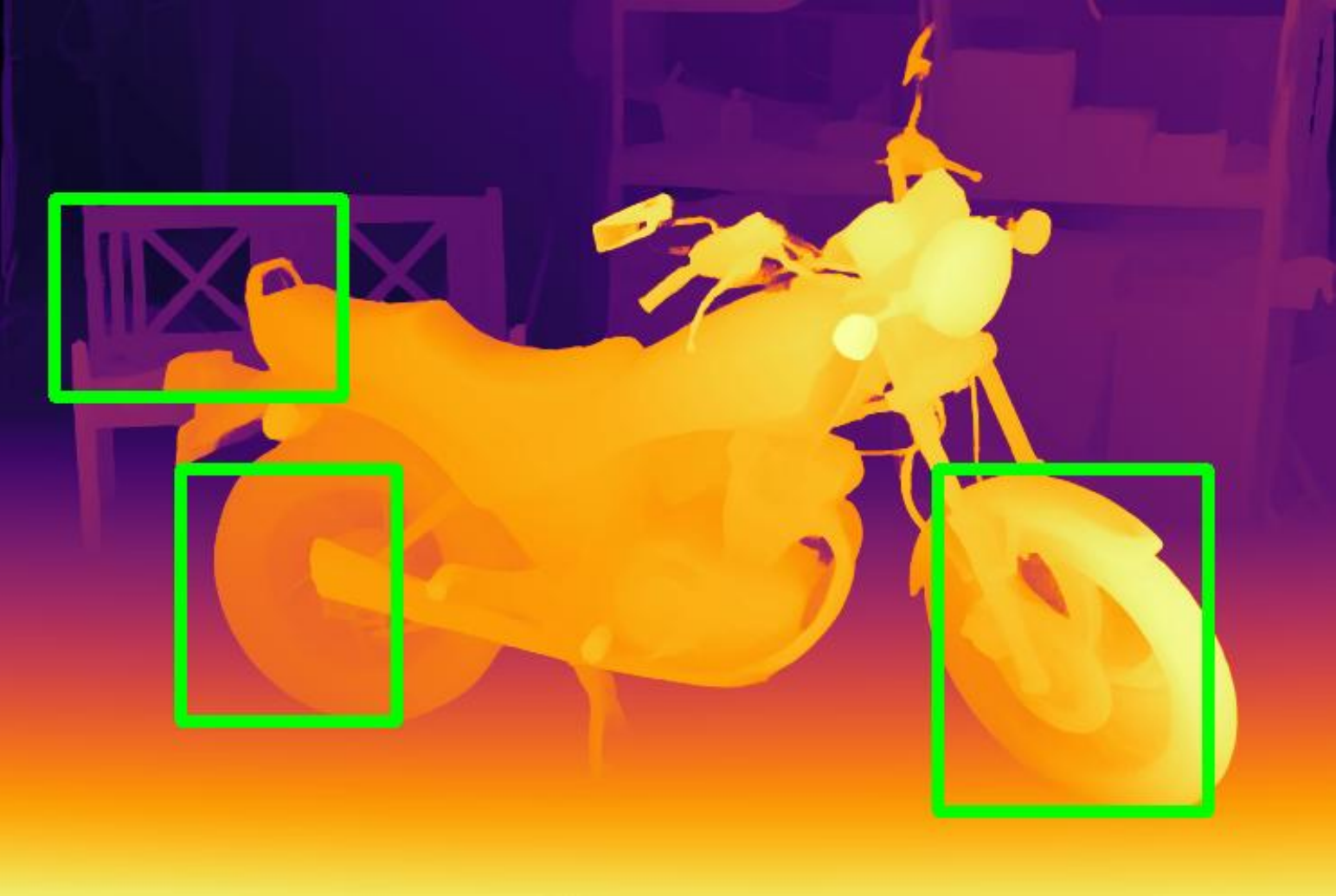}
\end{minipage}
}

\subfigure[DLNR]{
\begin{minipage}[b]{0.22\textwidth}
\includegraphics[width=1.0\textwidth]{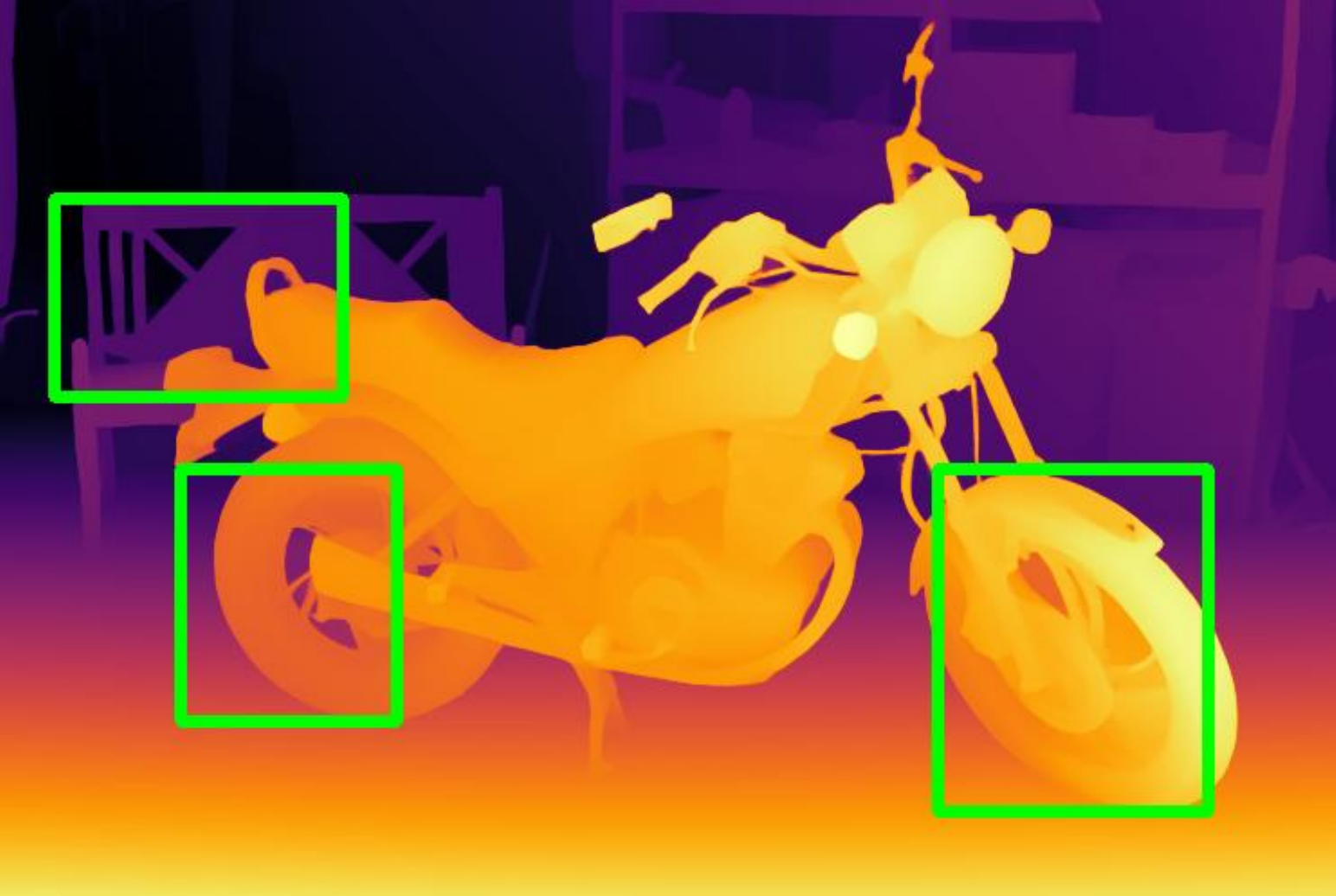}
\end{minipage}
}
\subfigure[Ours]{
\begin{minipage}[b]{0.22\textwidth}
\includegraphics[width=1.0\textwidth]{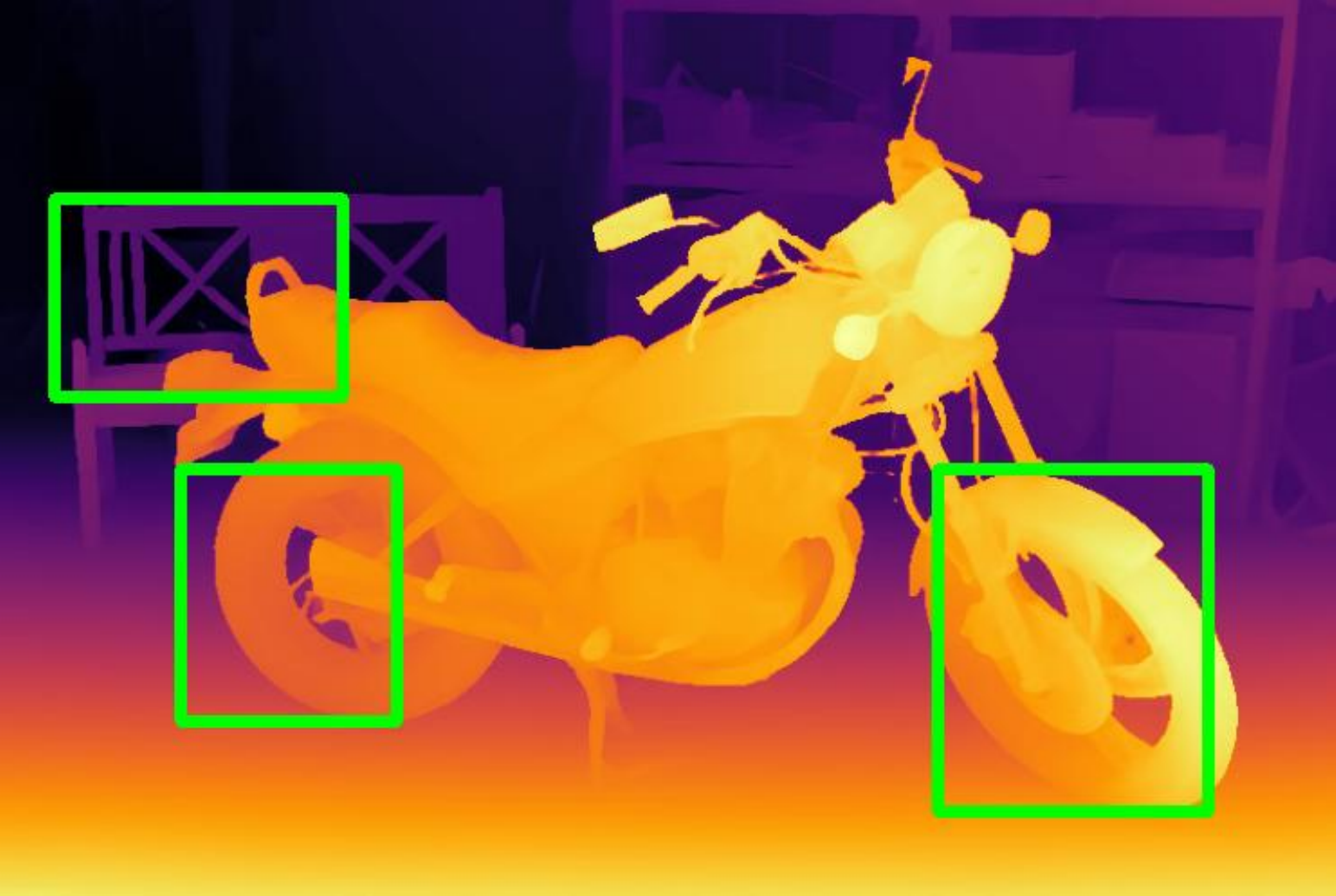}
\end{minipage}
}
\caption{ \textbf{Qualitative Comparison.} Comparison with state-of-the-art methods such as IGEV~\cite{xu2023iterative}, DLNR~\cite{zhao2023high} on Middlebury dataset. All methods are trained only on SceneFlow~\cite{mayer2016large}, and evaluated at quarter resolution. It can be observed that our method generalizes and predicts high-frequency details better than state-of-the-art methods.}
\label{fig:teaser}
\end{figure}

\section{Introduction}

Stereo matching is one of the most important problems in the field of computer vision \cite{hoff1989surfaces, kang1995, scharstein2002taxonomy, szeliski2022computer}. It involves the analysis of a pair of rectified stereo images, captured concurrently, with the objective of determining the pixel-level displacement map from the left image to the corresponding location in the right image, a representation commonly referred to as a ``disparity map''. In the context of rectified image pairs, the stereo matching problem can be conceptualized as a well-structured one-dimensional search problem in the image space \cite{szeliski2022computer}. The utility of stereo camera systems is underscored by their efficacy and cost-effectiveness, leading to their widespread adoption in diverse commercial and industrial applications. Notably, these applications encompass domains such as autonomous navigation \cite{fan2020computer, bimbraw2015autonomous}, smartphone technology \cite{meuleman2022floatingfusion, luo2020wavelet, pang2018zoom}, and various forms of robotic vision based automation systems  \cite{kim2021stereo, liu2023va, liu2023single, jain2023enhanced, jain2024learning, kaya2023multi}.

Classical stereo matching methods---often categorized as local methods, use a predefined support window to find suitable matches between stereo image pair \cite{scharstein2002taxonomy, hirschmuller2007stereo}. Yet, approaches that optimize for all disparity values using a global cost function were observed to provide better results \cite{kolmogorov2001computing, klaus2006segment, bleyer2011patchmatch, yamaguchi2014efficient}.
In recent years, the proliferation of high-quality, large scale synthetic ground-truth datasets, the availability of high-performance GPUs, and advancements in deep learning architectures have paved the way for deep-learning based stereo matching models trained within supervised settings. These models have shown a substantial improvement in accuracy compared to classical methods \cite{Kendall_2017_ICCV, chang2018pyramid, Zhang2019GANet, lipson2021raft}. Nevertheless, one fundamental challenge still remains, i.e., how to model \emph{continuous} scene disparity values given only a limited number of candidate pixels to match? After all, the scene is continuous in nature.


Numerous recent studies have undertaken the challenge of predicting continuous scene disparities, classifiable into two main categories: \textbf{\textit{(i)} Regression-based approaches} predict a real-valued offset by neural networks for each hypothesis of discrete disparity. The offset is then added to the discrete disparity hypothesis as the final continuous prediction. Typical examples include RAFT-Stereo \cite{lipson2021raft}, CDN \cite{garg2020wasserstein}, and more recently IGEV \cite{xu2023iterative} and DLNR \cite{zhao2023high}. \textbf{\textit{(ii)} Classification-based approaches} first estimate the categorical distribution\footnote{A categorical distribution is a discrete probability distribution that describes the possible results of a random variable that can take on the K possible categories, with the probability of each category separately specified.} for the discrete disparity hypotheses and then take the expectation value of the distribution as the final disparity, which can be any arbitrary real value even though the categorical distribution is discrete \cite{Kendall_2017_ICCV, chang2018pyramid, Zhang2019GANet}.

In this paper, we aim to address the importance of continuous disparity modeling in stereo matching, given the categorical distribution of disparity hypotheses. We introduce a new perspective on the disparity prediction problem by framing it as a search problem of finding the minimum risk \cite{LehmCase98, vapnik1991principles, berger2013statistical} of disparity values. Specifically, we define stereo risk as the average prediction error concerning all possible values of the ground-truth disparity. Since the ground-truth disparity is unavailable when making the prediction, we approximate it using the disparity hypotheses with a categorical distribution. We search for a disparity value as our prediction that achieves minimal overall risk involved with it. Furthermore, we show that the commonly used disparity expectation can be viewed as a specific instance of the $L^2$ error function of the proposed risk formulation framework. Yet,  $L^2$ error function approach, despite easy to optimize, is sensitive to multi-modal distribution and  leads to overly smooth solutions \cite{Chen_2019_ICCV, Tosi_2021_CVPR}. Thus, we introduce $L^1$ error function approach to risk minimization, offering potential benefit over $L^2$ limitations.



Nevertheless, our choice to use $L^1$ risk for model training leads to one practical problem, i.e., it's closed-form solution remains elusive. As a result, we embark on the pursuit of a solution by means of derivative computations applied to our novel risk function, followed by its continuous optimization. Our approach involves interpolating the disparity categorical distribution leading to defining a continuous probability density function. Subsequently, we introduce a binary search algorithm designed to efficiently identify the optimal disparity that minimizes the proposed risk. To facilitate end-to-end network training, we introduce the use the implicit function theorem \cite{krantz2002implicit} to compute the backward gradient of the final disparity concerning the categorical distribution. All these methodological choice ensures the better model training while optimizing the proposed risk.


\begin{figure*}[t]
\centering
\subfigure[Image]{
\begin{minipage}[b]{0.3\textwidth}
\includegraphics[width=\textwidth]
{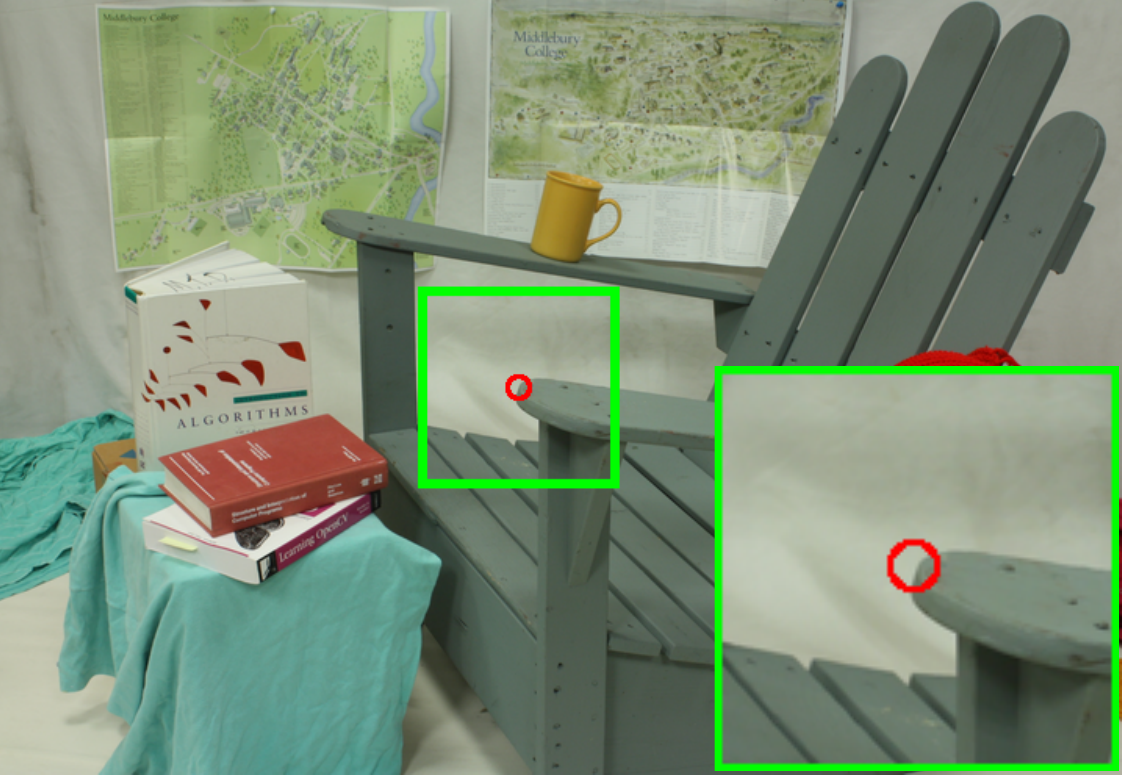}
\end{minipage}
}
\subfigure[Expectation]{
\begin{minipage}[b]{0.3\textwidth}
\includegraphics[width=\textwidth]
{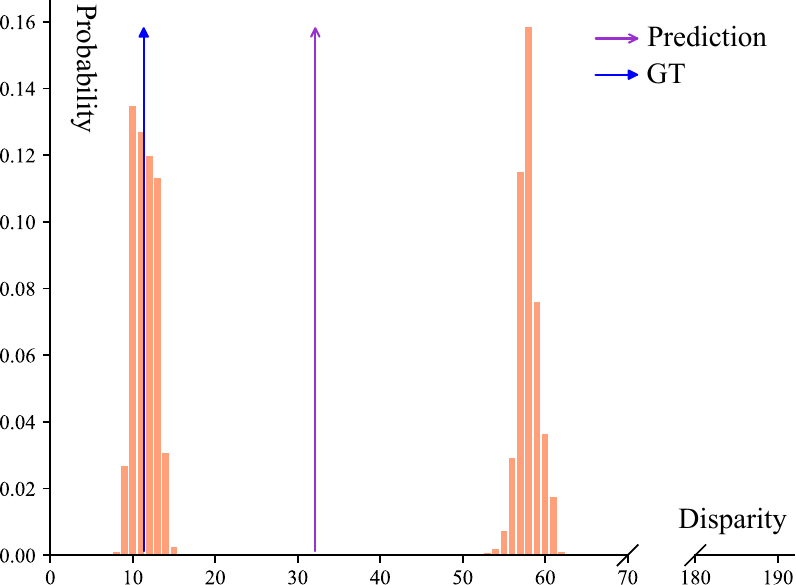}
\end{minipage}
}
\subfigure[Ours]{
\begin{minipage}[b]{0.3\textwidth}
\includegraphics[width=\textwidth]
{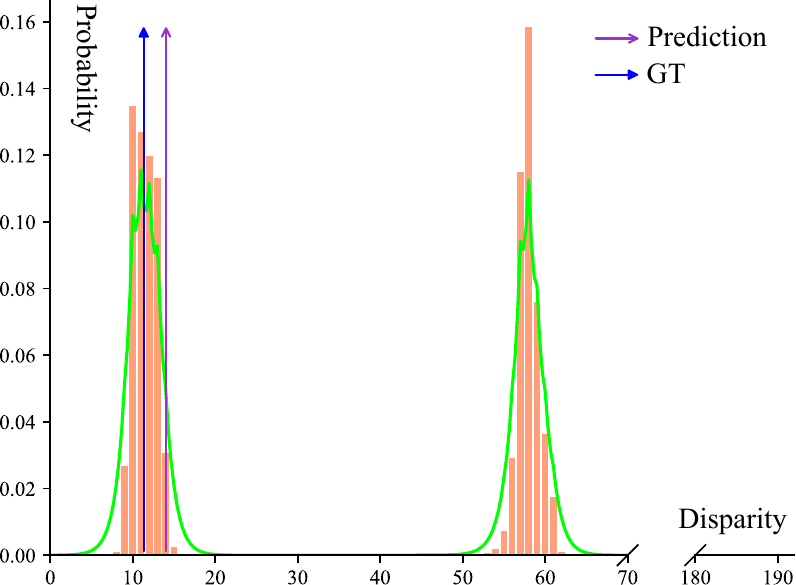}
\end{minipage}
}
\caption{\small \textbf{Difference between the expectation based approach and our method.} In (a) the pixel in the red circle is located at the boundary of the chair, thus the disparity distribution has multiple modes. (b) and (c) shows the discrete distribution of disparity hypotheses in orange bars. In (b) the prediction obtained by averaging is blurred and far from any of the modes. In (c) we obtained the optimal solution under $L^1$ norm, which is more robust and closer to the ground truth. The green curve is the interpolated probability density.}
\label{fig:hist}
\end{figure*}

Upon evaluations, our approach shows superior performance compared to many state-of-the-art methods on benchmark datasets such as SceneFlow \cite{mayer2016large}, KITTI 2012 \& 2015 \cite{geiger2012we, menze2015object}. Moreover, our approach achieves significantly better cross-domain generalization, as observed on Middlebury \cite{scharstein2002taxonomy}, ETH 3D \cite{schoeps2017cvpr}, and KITTI 2012 \& 2015. An example of qualitative comparison is given in Fig. \ref{fig:teaser}. Ablation studies confirm the effectiveness of risk minimization, not only within the proposed network but also in the context of general stereo-matching networks, such as ACVNet \cite{xu2022attention} and PCWNet \cite{shen2022pcw}.  We believe our work not only advances stereo matching in computer vision but also holds promise for its integration to robotics and control via risk analysis.

\section{Related Work}

\noindent
\textbf{\textit{(i)} Deep Stereo Matching.}
In recent years, there has been a substantial enhancement in the accuracy of stereo matching due to the adoption of deep learning-based methodologies. As a result, the pursuit of designing robust and efficient network architectures for stereo matching has emerged as a prominent area of research. For instance, Zbontar et al. \cite{Zbontar_2015_CVPR} harnessed deep convolutional networks to acquire discriminative features for image patches. DispNetCorr \cite{mayer2016large} introduced explicit correlation within networks to construct cost volumes. GCNet \cite{Kendall_2017_ICCV} employed volume concatenation and refined it through 3D convolution. PSM-Net \cite{chang2018pyramid} leveraged spatial pyramid pooling \cite{zhao2017pyramid} and stacked hourglass networks \cite{newell2016stacked} to capture contextual information. STTR \cite{Li_2021_ICCV} extended the flexibility of disparity range by employing transformers \cite{vaswani2017attention, dosovitskiy2021an}. Furthermore, considerations pertaining to the uniqueness constraint were addressed using optimal transport \cite{cuturi2013sinkhorn}. ACVNet \cite{xu2022attention} incorporated attention mechanisms to weight matching costs, further contributing to the advancement of stereo matching methodologies.

Another line of research is to improve efficiency. In GANet \cite{Zhang2019GANet} the computationally costly 3D convolutions are replaced by the differentiable semi-global aggregation \cite{hirschmuller2007stereo}. GWCNet \cite{guo2019group} constructs the cost volume by group-wise correlation. AANet \cite{xu2020aanet} proposes the adaptive cost aggregation to replace the 3D convolution for efficiency. AnyNet \cite{wang2019anytime}, DeepPruner \cite{Duggal2019ICCV}, HITNet \cite{tankovich2021hitnet}, CasMVSNet \cite{gu2020cascade}, PCWNet \cite{shen2022pcw} and Bi3D \cite{badki2020bi3d} prune the range of disparity in the iterative manner. RAFT-Stereo \cite{lipson2021raft}, CREStereo \cite{li2022practical}, IGEV \cite{xu2023iterative} and DLNR \cite{zhao2023high} use recurrent neural networks \cite{cho2014learning} to predict and refine the disparity iteratively. 

Inspired by CasMVSNet \cite{gu2020cascade}, our network consists of two stages one to predict and other to refine the disparity map. This hierarchical design reduces the time and memory cost, while keeping the matching accuracy. 

\noindent
\textbf{\textit{(ii)} Continuous Disparity by Classification.}
In deep networks featuring cost volumes, the prevalent method for predicting disparity from these volumes involves the weighted average operation, commonly referred to as expectation. For instance,
\cite{Chen_2019_ICCV} find the average operation suffers from the over-smoothing problem, introducing the concept of a single-modal weighted average. \cite{garg2020wasserstein} propose to predict a continuous offset to shift the distribution modes of disparity. Furthermore, they generate multi-modal ground truth disparity distributions and supervise the network to learn the distribution by Wasserstein distance. SMD-Net \cite{Tosi_2021_CVPR} exploit bimodal mixture densities as output representation for disparities. UniMVSNet \cite{unimvsnet} aimed to unify the benefit of both classification and regression by introducing a novel representation and a unified focal loss. Yang et al. \cite{yang2022non} tackled the multi-modal issue by utilizing the top-K hypotheses for disparity. On the contrary, we propose to minimize the risk under $L^1$ norm to capture continuous disparity and solve the multi-modal problem. Moreover, our approach can be trained in an end-to-end manner.


\noindent
\textbf{\textit{(iii)} Robustness and Cross-Domain Generalization.}
Existing real-world stereo datasets are small and insufficient to train deep neural networks for possible variations at test time, thereby making network robust and apt for cross-domain generalization. In this spirit, \citet{Tonioni_2017_ICCV, tonioni2019learning, Tonioni_2019_CVPR} fine tune the stereo matching networks on the target domain via unsupervised loss. \citet{Liu_2020_StereoGAN} jointly optimize networks for domain translation and stereo matching during training. \citet{zhang2020domain, Song_2021_CVPR} normalize features to reduce domain shifts. \citet{cai2020matching, liu2022local} design robust features for stereo matching. \citet{liu2022graftnet} shows that the cost volume built by cosine similarity generalizes better. \citet{zhang2022revisiting} apply the stereo contrastive loss and selective whitening loss to improve feature consistency. \citet{chang2023domain} proposed the hierarchical visual transformation to learn invariant robust representation from synthetic images. Our approach can be combined with above methods to further improve the robustness. Yet, we present a novel perspective to improve robustness by $L^1$ risk minimization.



\section{Method}
\subsection{Probability Modeling for Continuous Disparity}
For each pixel in the left image, suppose the possible disparities are in the range of $[d_{\mathtt{min}}, d_{\mathtt{max}}]$. Conventional stereo matching algorithms typically calculate a cost function that can equate to a probability mass function (PMF)
%
%
with a finite set of disparities $\mathbf{d}=[d_1, ..., d_N]^T$. It computes a discrete distribution $\mathbf{p}^m=[p^m_1, ..., p^m_N]^T$, where $d_{i}\in[d_{\mathtt{min}}, d_{\mathtt{max}}]$ and $p^m_i$ is the probability that the ground truth disparity is $d_i$. The $\mathbf{p}^m$  must satisfy $p^m_i\geq0$ and $\sum_ip^m_i=1$. 

The discrete formulation reasons the probability only at a finite set of disparities. Yet, in real-world applications, the ground-truth disparity is continuous. Thus, we propose to interpolate the discrete distribution via Laplacian kernel, and compute the probability density function of disparity $x\in\mathbb{R}$ as
\begin{equation}
    p(x;\mathbf{p}^m) = \sum_i^N k(x, d_i) p^m_i,
\end{equation}
here $k(x, d_i)$ is defined as $\frac{1}{2\sigma}\exp{-\frac{|x-d_i|}{\sigma}}$, and $\sigma$ is the hyper-parameter for bandwidth. The above density function is valid as $p(x;\mathbf{p}^m)\geq0$ for $\forall ~x\in\mathbb{R}$ and $\int p(x;\mathbf{p}^m)dx=1$. An illustration of the interpolation is shown in Fig. \ref{fig:hist} (c). The orange bars represent the given discrete distribution $\mathbf{p}^m$, and the green curve is the interpolated density function. Later, we show that such a continuous modeling enable us to compute derivative of the proposed stereo risk function.

\subsection{Risk in Stereo Matching}
To choose a value as the final prediction, we propose to minimize the following  risk:
\begin{equation}
    \mathtt{argmin}_y F(y, \mathbf{p}^m) = \mathtt{argmin}_y \int \mathcal{L}(y,x) p(x; \mathbf{p}^m) dx
    \label{eq:general_risk}
\end{equation}
where $F(y, \mathbf{p}^m)$ is called as the risk at $y$, and $\mathcal{L}(y, x)$ is the error function between $y$ and $x$. By risk we mean that if we take $y$ as predicted disparity, how much error there shall be with respect to the ground truth. Since the exact ground truth is unavailable at the time for making the prediction, we average the error across all possible ground-truth disparities with the distribution $p(x; \mathbf{p}^m)$.

Previous methods usually compute the expectation value of $x$ as the final prediction for the disparity:
\begin{equation}
y=\int xp(x;\mathbf{p}^m)dx.
\end{equation}
We want to point out that  we can arrive at the same prediction by using squared $L^2$ norm loss as $\mathcal{L}(y,x)$ in Eq.\eqref{eq:general_risk}, i.e., $\mathtt{argmin}_y F(y, \mathbf{p}^m)=\int xp(x;\mathbf{p}^m)dx$ if $\mathcal{L}(y,x) = (y-x)^2$.
%
Nevertheless, it is well known that the $L^2$ norm is not robust to outliers. As an example, in Fig. \ref{fig:hist} (b) it can be observed that the estimated expectation is inaccurate when there are multiple modes in the distribution. And therefore, we resort to $L^1$ norm of $\mathcal{L}(y,x)$ in Eq.\eqref{eq:general_risk}, i.e.,
\begin{equation}
    \mathtt{argmin}_y F(y, \mathbf{p}^m) = \mathtt{argmin}_y \int |y-x| p(x; \mathbf{p}^m) dx.
    \label{eq:l1_risk}
\end{equation}
Given the distribution $p(x; \mathbf{p}^m)$ of the disparity, the optimal $y$ will minimize the $L^1$ error with respect to all possible disparities weighted by the corresponding probability density. 
As shown in Fig. \ref{fig:hist} (c), our final prediction is more robust to the incorrect modes and closer to the ground truth.

\begin{figure*}
    \centering
    \includegraphics[width=0.9\textwidth]{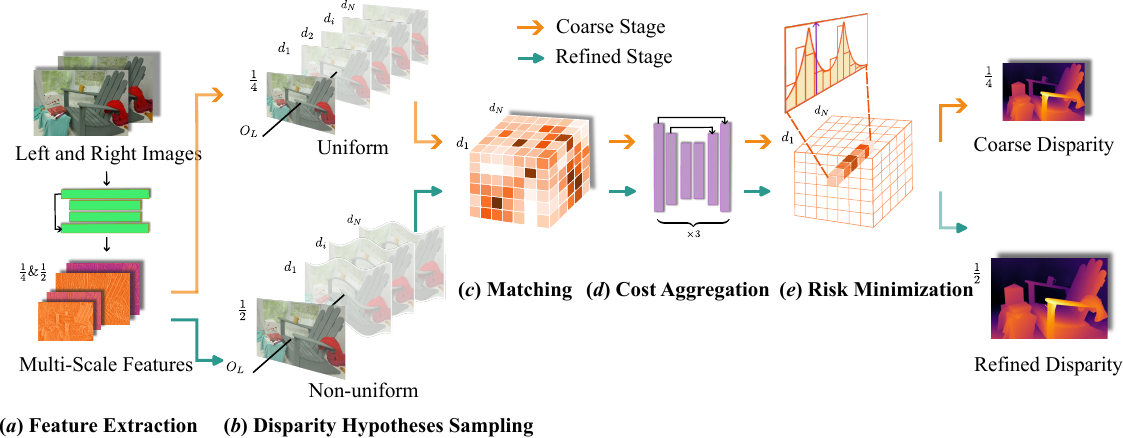}
    \caption{\small \textbf{Overall pipeline (Left to Right).} We first extract multi-scale features from left and right images respectively. The subsequent procedures are divided into two stages. In the coarse stage ---shown in orange arrow, we sample  disparity hypotheses uniformly and match on 1/4-resolution features. While in the refined stage---shown in green arrow, to match 1/2-resolution features efficiently. Disparity hypotheses are sampled centering around the disparity predicted from the coarse stage. In both stages, we first construct cost volumes by concatenation, and then apply the stacked hourglass networks to aggregate the matching cost, and finally search for the disparity that minimizes the proposed $L^1$ risk in Eq.(\ref{eq:l1_risk}).  
    }
    \label{fig:pipeline}
\end{figure*}

\subsection{Differentiable Stereo Risk Minimization}
Obtaining a minimal risk solution to Eq.(\ref{eq:l1_risk}) seems difficult as it is challenging to derive its closed form formulation. 
So, performing end-to-end learning with deep network seems difficult. To this end, we put forward an approach that enable end-to-end learning of the network as follows:


\textbf{\textit{(i)} Forward Prediction.} Given a discrete distribution $\mathbf{p}^m$, we find the optimal $y$ for Eq.(\ref{eq:l1_risk}) based on the following two observations. Firstly, the target function $F(y, \mathbf{p}^m)$ is convex with respect to $y$, hence we compute the optimal solution where $\partial F / \partial y = 0$, i.e., 
\begin{align}
    &G(y, \mathbf{p}^m) \triangleq  \frac{\partial F(y, \mathbf{p}^m)}{\partial y} \\
    &= \sum_ip^m_i\mathtt{Sign}(y-d_i)(1-\exp{-\frac{|y-d_i|}{\sigma}}) = 0
    \label{eq:fisrt_derivative}
\end{align}
where $\mathtt{Sign}()$ is the sign function (a slight abuse of notation). $\mathtt{Sign}()$ can be thought of as an indicator function, i.e., it is 1 if $y>d_i$ and $-1$ otherwise. Secondly, the second-order derivative $\partial^2 F / \partial^2 y \geq 0$, indicating that the first-order derivative is a non-decreasing function. We find the optimal disparity, i.e., the zero point of $G(y, \mathbf{p}^m)$, by binary search, as shown in Algorithm \ref{alg:forward}. In all our experiments, we set $\sigma=1.1$ and $\tau = 0.1$. For $N$ disparity hypotheses, the binary search algorithm can find the optimal solution with time complexity of $O(\log N)$ \citep{cormen2009}.

\begin{algorithm}[tb]
\caption{Forward Prediction}\label{alg:forward}
\begin{algorithmic}
\INPUT $\tau > 0$, $\sigma>0$, $\mathbf{d}=[d_1, ..., d_N]$, $d_1<d_2<\cdots<d_N$, and $\mathbf{p}^m=[p^m_1,...,p^m_N]$ 
\STATE $d^l \gets d_1$ 
\STATE $d^r \gets d_N$ 
\STATE $g \gets \tau + 1$ 
\WHILE{$|g| > \tau$}
\STATE $d^m \gets (d^l + d^r) / 2.0$ 
\STATE $g \gets \sum_ip^m_i\mathtt{Sign}(d^m-d_i)(1-\exp{-\frac{|d^m-d_i|}{\sigma}})$ 
\IF{$g>0$}  
    \STATE $d^r \gets d^m$
\ELSE
    \STATE $d^l \gets d^m$ 
\ENDIF
\ENDWHILE
\OUTPUT $d^m$ 
\end{algorithmic}
\end{algorithm}

\textbf{\textit{(ii)} Backward Propagation.} Our approach to forward prediction for solving Eq.(\ref{eq:l1_risk}) contains non-differentiable operations ---refer Algorithm \ref{alg:forward}. Yet, to enable end-to-end training, we have to compute $dy / d\mathbf{p}^m$ to propagate the gradient backward. Now, since $G(y, \mathbf{p}^m)\equiv 0$ at the optimal $y$,  we obtain the following via use of Implicit Function Theorem \citep{krantz2002implicit}:
\begin{equation}
dG(y,\mathbf{p}^m) =  \frac{\partial G}{\partial y} dy + \frac{\partial G}{\partial \mathbf{p}^m} d\mathbf{p}^m = 0.  
\end{equation}
By organizing the terms, we obtain
\begin{align}
    \frac{dy}{d\mathbf{p}^m} &= -\frac{\partial G /\partial \mathbf{p}^m}{\partial G / \partial y} \\
    &= [\ldots, \frac{\sigma\mathtt{Sign}(d_i-y)(1-\exp{-\frac{|y-d_i|}{\sigma}})}{\sum_jp^m_j\exp{-\frac{|y-d_j|}{\sigma}}},\ldots]^T, 
    \label{eq:implicit_gradient}
\end{align}
showing the back-propagation computation. Here, we clip the denominator $\sum_jp^m_j\exp{-\frac{|y-d_j|}{\sigma}}$ in the above equation to be no less than $0.1$ to avoid large gradients. 

\begin{table*}
\begin{center}
\scriptsize
\caption{Comparison with state-of-the-art on SceneFlow test set. The \textcolor{red}{$1^{st}$} and \textcolor{blue}{$2^{nd}$} bests are in red and blue, respectively. \textbf{Ours} in bold. 
}
\label{tab:sceneflow_sota}
\begin{small}
\begin{sc}
\begin{tabular*}{1.0\textwidth}{l@{\extracolsep{\fill}}cccccc}
\toprule
Method & Param (M) & Time (s) & EPE $\downarrow$ &$>0.5$px $\downarrow$  &$>1$px $\downarrow$ &$>2$px $\downarrow$\\
\midrule
CFNet \cite{shen2021cfnet} & 21.98 & 0.13 & 1.04 & 15.91 & 10.30 & 6.89\\ 
PCWNet \cite{shen2022pcw}& 34.27 & 0.25 & 0.90 & 17.59 & 8.08 & 4.57 \\
ACVNet \cite{xu2022attention} & 6.84 & 0.16 & \textcolor{blue}{0.47} & 9.70 & \textcolor{blue}{5.00} & \textcolor{blue}{2.74}\\
DLNR \cite{zhao2023high} & 54.72 & 0.44 & 0.53 & 8.75 & 5.44 & 3.44\\
IGEV \cite{xu2023iterative} & 12.60 & 0.36 & \textcolor{blue}{0.47} & \textcolor{blue}{8.51} & 5.21 & 3.26\\
\hline
\textbf{Ours} & \textbf{11.96} & \textbf{0.35} &  \textbf{\textcolor{red}{0.43}} & \textbf{\textcolor{red}{8.10}} & \textbf{\textcolor{red}{4.22}} & \textbf{\textcolor{red}{2.34}}\\
\bottomrule
\end{tabular*}
\end{sc}
\end{small}
\end{center}
\end{table*}

\begin{table*}
\begin{center}
\scriptsize
\caption{Comparison with state-of-the-art methods on KITTI 2012 Benchmark. $\dagger$ denotes using extra data for pre-training. The \textcolor{red}{first} and \textcolor{blue}{second} bests are in red and blue respectively. \textbf{Our method} in bold. 
The results are obtained from KITTI official website.}
\label{tab:kitti12_sota}
\begin{small}
\begin{sc}
\begin{tabular*}{1.0\textwidth}{l@{\extracolsep{\fill}}cccccc}
\toprule
\multirow{2}{*}{Method} & \multirow{2}{*}{Param (M)} & \multirow{2}{*}{Time (s)} & \multicolumn{2}{c}{$>2$px} &\multicolumn{2}{c}{$>3$px}\\
 & & & Noc & All & Noc & All\\
\midrule
LEAStereo \cite{cheng2020hierarchical} & 1.81 & & 1.90 & 2.39 & 1.13 & 1.45\\
CFNet \cite{shen2021cfnet} & 21.98 &  0.12 & 1.90 & 2.43& 1.23 &1.58 \\
ACVNet \cite{xu2022attention} & 6.84 & 0.15 & 1.83 & 2.34 & 1.13 & 1.47\\
ACFNet \cite{chen2021cost} &  & & 1.83 & 2.35 &1.17 & 1.54\\
NLCA-Net v2 \cite{rao2022rethinking} & & & 1.83 & 2.34 & 1.11 & 1.46 \\
CAL-Net \cite{chen2021cost} &  & & 1.74 & 2.24 &1.19 & 1.53\\
CREStereo \cite{li2022practical} $\dagger$&  & & 1.72 & \textcolor{blue}{2.18} & 1.14& 1.46 \\
LaC+GANet \cite{liu2022local} & 9.43 & & 1.72 & 2.26 & 1.05 & \textcolor{blue}{1.42}\\
IGEV \cite{xu2023iterative}$\dagger$ & 12.60 & 0.32 & 1.71 & \textcolor{red}{2.17} & 1.12 & 1.44\\
PCWNet \cite{shen2022pcw} & 34.27 & 0.23 & \textcolor{blue}{1.69} & \textcolor{blue}{2.18} & \textcolor{blue}{1.04} & \textcolor{red}{1.37}\\
\hline
\textbf{Ours} & \textbf{11.96} & \textbf{0.32} &  \textbf{\textcolor{red}{1.58}} & \textbf{2.20} & \textbf{\textcolor{red}{1.00}} &\textbf{1.44}\\
\bottomrule
\end{tabular*}
\end{sc}
\end{small}
\end{center}
\end{table*}

\begin{table*}[t]
\begin{center}
\scriptsize
\caption{Comparison with state-of-the-art methods on KITTI 2015 Benchmark. $\dagger$ denotes using extra data for pre-training. The \textcolor{red}{first} and \textcolor{blue}{second} bests are in red and blue respectively. \textbf{Our method} in bold. 
The results are obtained from KITTI official website.}
\label{tab:kitti15_sota}
\begin{small}
\begin{sc}
\begin{tabular*}{1.0\textwidth}{l@{\extracolsep{\fill}}cccccccc}
\toprule
\multirow{2}{*}{Method} & \multirow{2}{*}{Param (M)} & \multirow{2}{*}{Time (s)} & \multicolumn{3}{c}{All} &\multicolumn{3}{c}{Noc}\\
 & & & D1\_bg & D1\_fg & D1\_all & D1\_bg & D1\_fg & D1\_all\\
\midrule
LEAStereo \cite{cheng2020hierarchical} & 1.81 & & 1.40 & 2.91 & 1.65 & 1.29 & 2.65  &  	1.51\\
CFNet \cite{shen2021cfnet} & 21.98 & 0.12  & 1.54 &3.56 & 1.88 &1.43 & 3.25 & 1.73 \\
ACVNet \cite{xu2022attention} & 6.84 & 0.15 & \textcolor{red}{1.37}  & 3.07 & 1.65  & \textcolor{blue}{1.26} & 2.84 & 1.52\\
ACFNet \cite{chen2021cost} &  & & 1.51 & 3.80 &1.89  & 1.36 & 3.49 & 1.72\\
NLCA-Net v2 \cite{rao2022rethinking} & & & 1.41 & 3.56 & 1.77 & 1.28 & 3.22 & 1.60 \\
CAL-Net \cite{chen2021cost} &  & & 1.59 & 3.76 &1.95 & 1.45 & 3.42 & 1.77 \\
CREStereo \cite{li2022practical} $\dagger$ &  & &1.45  & 2.86 & 1.69 &1.33 & 2.60 & 1.54  \\
LaC+GANet \cite{liu2022local} & 9.43 & & 1.44  & 2.83 & 1.67 & \textcolor{blue}{1.26} & 2.64 & \textcolor{blue}{1.49}\\
IGEV \cite{xu2023iterative} $\dagger$ & 12.60 & 0.32 & \textcolor{blue}{1.38}  & 2.67 & \textcolor{red}{1.59} & 1.27 & 2.62 & \textcolor{blue}{1.49}\\
DLNR \cite{zhao2023high} & 54.72 & 0.39 & 1.60 & \textcolor{red}{2.59} & 1.76 & 1.45 & \textcolor{red}{2.39} & 1.61 \\
PCWNet \cite{shen2022pcw} & 34.27 & 0.23 & \textcolor{red}{1.37} & 3.16 & 1.67 & \textcolor{blue}{1.26} & 2.93 & 1.53 \\
CroCo  \cite{philippe2023croco}$\dagger$ & 417.15 & & \textcolor{blue}{1.38} & \textcolor{blue}{2.65} & \textcolor{red}{1.59} & 1.30 & \textcolor{blue}{2.56} & 1.51 \\
\midrule
\textbf{Ours} & \textbf{11.96} & \textbf{0.32} &  \textbf{1.40} &  \textbf{2.76} & \textbf{\textcolor{blue}{1.63}} & \textbf{\textcolor{red}{1.25}} & \textbf{2.62} & \textbf{\textcolor{red}{1.48}}\\
\bottomrule
\end{tabular*}
\end{sc}
\end{small}
\end{center}
\end{table*}

\subsection{Network Architecture}
To find the disparity value, we match the image patches of left and right images by constructing stereo cost volumes as done in \citet{kendall2017end, chang2018pyramid}. Yet, an exhaustive matching requires extensive memory and computation. So, for efficiency, we adopt a cascade structure following \citet{gu2020cascade}. Categorically, we first sample the disparity hypothesis by a coarse matching at low-resolution image features. This reduces the search space to a large extent. Next, we refine the sampled hypothesis at high-resolution image features. 

Fig. \ref{fig:pipeline} shows the overall network architecture details. For clarity on our design choices, we explain the network components in five module as follows\footnote{More details are provided in the Appendix}:


\noindent
\textbf{\textit{(a)} Feature Extraction.} Given an input image, the module aims to output multi-scale 2D feature maps. Specifically, we first use a ResNet \cite{he2016deep} to extract 2D feature maps of resolution 1/4 and 1/2 with respect to the input image. The ResNet contains 4 stages of transformation with 3, 16, 3, 3 residual blocks, respectively. The spatial resolution is downsampled before the beginning of the first and third stages of transformation. Next, we apply the spatial pyramid pooling \cite{zhao2017pyramid} on the 1/4-resolution feature map from the fourth stage to enlarge the receptive field. In the end, we upsample the enhanced feature map from 1/4 to 1/2 and fuse it with the 1/2-resolution feature map from ResNet. The final outputs are the feature maps of 1/4 and 1/2 resolution. We apply the same network and weights to extract features from left and right images.

\noindent
\textbf{\textit{(b)} Disparity Hypotheses Sampling.} The disparity hypotheses provide the candidates of pixel pairs to match. In the coarse stage, we uniformly sample 192 hypotheses in the range 0 to maximum possible disparity. In the refined stage, we reduce the sampling space according to the predicted disparity from the coarse stage. Concretely, for each pixel we sample 16 hypotheses between the minimum and maximum disparity in the local window of size $12\times12$. 

\noindent
\textbf{\textit{(c)} Matching.} We match the 2D feature maps from the left and right images according to the sampled disparity hypothesis. The features at each pair of candidates pixels for matching will be concatenated along the channel dimension, which forms a 4D stereo cost volume (feature$\times$ disparity$\times$height$\times$width). In the coarse stage, we match the feature map of 1/4 resolution for efficiency. To capture high-frequency details, we match the  1/2-resolution feature map in the refined stage.

\noindent
\textbf{\textit{(d)} Cost Aggregation.} We use the stacked hourglass architecture \cite{newell2016stacked} to transform the stereo cost volume and aggregate the matching cost. For the coarse and refined stages, the structures are same except for the number of feature channels. Specifically, the network consists of three 3D hourglasss as in \cite{chang2018pyramid}. Each hourglass first downsamples the volume hierarchically to 1/2 and 1/4 resolution with respect to the input volume, and then upsample in sequence to recover the resolution. This procedure helps aggregate information across various scales. The final output is a volume that represents the discrete distribution of disparity hypotheses.

\noindent
\textbf{\textit{(e)} Risk Minimization.} This module applies Alg. \ref{alg:forward} to compute the optimal continuous disparity for each pixel given the discrete distribution of disparity hypotheses. At train time, we additionally compute the gradient according to Eq.(\ref{eq:implicit_gradient}) to enable backward propagation.

\subsection{Loss Function}
Given the predicted disparity $x^{\mathtt{pred}}\in\mathbb{R}$ and the ground-truth disparity $x^{\mathtt{gt}}\in\mathbb{R}$, we compute the smooth $L^1$ loss as
\begin{equation}
    \mathcal{L}(x^{\mathtt{gt}}, x^{\mathtt{pred}}) = 
    \begin{cases}
    0.5(x^{\mathtt{gt}}-x^{\mathtt{pred}})^2, ~~\text{if $|x^\mathtt{gt}-x^{\mathtt{pred}}| < 1.0$}\\
    |x^\mathtt{gt}-x^{\mathtt{pred}}| - 0.5, ~~\text{otherwise}
    \end{cases}
\end{equation}
We apply the above loss function to the predicted disparities from both the coarse and refined stages, and obtain $\mathcal{L}_{\mathtt{coarse}}$ and $\mathcal{L}_{\mathtt{refined}}$, respectively. The total loss is thus defined as $\mathcal{L}=0.1 *\mathcal{L}_{\mathtt{coarse}}+1.0*\mathcal{L}_{\mathtt{refined}}$.

\section{Experiments and Results}
\textbf{Implementation Details.} We implement our method in PyTorch 2.0.1 (Python 3.11.2) with CUDA 11.8. The software is evaluated on a machine with GeForce-RTX-3090 GPU.\\
\textbf{Datasets.} We perform experiments on four datasets namely SceneFlow \cite{mayer2016large}, KITTI 2012 \& 2015 \cite{geiger2012we, menze2015object}, Middlebury 2014 \cite{scharstein2002taxonomy}, and ETH 3D \cite{schoeps2017cvpr}. \textbf{(a) SceneFlow} is a synthetic dataset containing 35,454 image pairs for training, and 4,370 image pairs for test. \textbf{(b) KITTI 2012 \& 2015} are captured for autonomous driving. There are 194 training image pairs and 195 test image pairs in KITTI 2012. For KITTI 2015, there are 200 training image pairs and 200 test image pairs. \textbf{(c) Middlebury 2014} is an indoor dataset including 15 image pairs for training. \textbf{(d) ETH 3D} is a gray-scale dataset providing 27 image pairs for training.  \\
\textbf{Training Details.} We train our network on SceneFlow. The weight is initialized randomly. We use AdamW optimizer \cite{loshchilov2018decoupled} with weight decay $10^{-5}$. The learning rate decreases from $2\times10^{-4}$ to $2\times10^{-8}$ according to the one cycle learning rate policy. We train the network for $2\times 10^{5}$ iterations. The images are randomly cropped to $320\times736$. For KITTI 2012 \& 2015 benchmarks, we further fine tune the network on the training image pairs for $2.5\times10^{3}$ iterations. The learning rate goes from $5\times10^{-5}$ to $5\times10^{-9}$ over iterations. More details are provided in the Appendix.

\begin{table}[thb]
\begin{center}
\scriptsize
\caption{Cross-domain evaluation on Middlebury train set of quarter resolution. $\dagger$ denotes using extra data for pre-training. The \textcolor{red}{first} and \textcolor{blue}{second} bests are in red and blue respectively.  
All methods are trained on SceneFlow and evaluated on Middlebury train set without fine-tuning.
}
\label{tab:zero_mid}
\begin{small}
\begin{sc}
\begin{tabular}{lcccc}
\toprule
\multirow{2}{*}{Method}  & \multicolumn{2}{c}{$>0.5$px} &\multicolumn{2}{c}{$>1$px}\\
 & Noc & All & Noc & All\\
\midrule
CFNet  & 29.50 & 34.30 & 17.85 & 22.16	 \\
ACVNet  & 39.04 & 42.97 & 22.68 & 26.49\\
DLNR  & 19.43 & \textcolor{blue}{23.75} & \textcolor{blue}{10.16} & \textcolor{blue}{13.76} \\
IGEV $\dagger$  & \textcolor{red}{19.05} & \textcolor{red}{23.33} & 10.44 & 14.05\\
PCWNet   & 33.33 & 38.00 & 16.80 & 21.36\\
\midrule
\textbf{Ours} & \textbf{\textcolor{blue}{19.22}} & \textbf{\textcolor{red}{23.33}} & \textbf{\textcolor{red}{9.32}} & \textbf{\textcolor{red}{12.63}}\\
\bottomrule
\end{tabular}
\end{sc}
\end{small}
\end{center}
\end{table}

\subsection{In-Domain Evaluation}
Tab.(\ref{tab:sceneflow_sota}), Tab.(\ref{tab:kitti12_sota}) and Tab.(\ref{tab:kitti15_sota}) provide statistical comparison results with the competing methods on SceneFlow, KITTI 2012, and KITTI 2015 bechmarks, respectively. All the methods have been trained and fine-tuned on the corresponding training set. For SceneFlow test set, our proposed approach shows the best results over all the evaluation metrics. Particularly, we reduce $>1$px error from 5.00 to 4.22, and  $>0.5$px error from 8.51 to 8.10. For KITTI 2012 \& 2015 benchmarks, the matching accuracy of our approach in the non-occluded regions rank the first among the published methods. Especially, in KITTI 2012, we reduce the $>2$px error in non-occluded regions by 0.11.

\begin{table}[thb]
\begin{center}
\scriptsize
\caption{Cross-domain evaluation on ETH 3D train set. $\dagger$ denotes using extra data for pre-training. The \textcolor{red}{first} and \textcolor{blue}{second} bests are in red and blue respectively. 
All methods are trained on SceneFlow and evaluated on ETH 3D train set without fine-tuning.}
\label{tab:zero_eth3d}
\begin{small}
\begin{sc}
\begin{tabular}{lcccccc}
\toprule
\multirow{2}{*}{Method}  & \multicolumn{2}{c}{$>0.5$px} &\multicolumn{2}{c}{$>1$px}\\
 & Noc & All & Noc & All\\
\midrule
CFNet  & 15.57 & 16.24 & 5.30 & 5.59 \\
ACVNet  & 21.83 & 22.64 & 8.13 & 8.81\\
DLNR  & 18.66 & 19.07 & 13.11 & 13.39 \\
IGEV $\dagger$ & \textcolor{blue}{9.83} & \textcolor{blue}{10.39} & \textcolor{blue}{3.60} & \textcolor{blue}{4.05}\\
PCWNet & 18.25 & 18.88 & 5.17 & 5.43\\
\midrule
\textbf{Ours} &  \textbf{\textcolor{red}{7.90}}  & \textbf{\textcolor{red}{8.59}} & \textbf{\textcolor{red}{2.41}} & \textbf{\textcolor{red}{2.71}}\\
\bottomrule
\end{tabular}
\end{sc}
\end{small}
\end{center}
\end{table}

\begin{table*}
\begin{center}
\scriptsize
\caption{Ablation studies on Middlebury training set of quarter resolution. The \textcolor{red}{first} and \textcolor{blue}{second} bests are in red and blue respectively. \textbf{Our method} in bold. 
All methods are trained on SceneFlow and evaluated on Middlebury training set without fine-tuning.}
\label{tab:ablation}
\begin{small}
\begin{sc}
\begin{tabular*}{1.0\textwidth}{l@{\extracolsep{\fill}}cccccccc}
\hline
\multirow{2}{*}{Backbone} &  \multirow{2}{*}{Training}&\multirow{2}{*}{Test} & \multirow{2}{*}{Param (M)} & \multirow{2}{*}{Time (s)} & \multicolumn{2}{c}{$>1$px} &\multicolumn{2}{c}{$>2$px}\\
& & & & & Noc & All & Noc & All\\
\hline
\multirow{2}{*}{ACVNet} & Expectation & Expectation &6.84 & 0.12 & 22.68 & 26.49 & 13.54 & 16.49\\
 & Expectation &L1-Risk & 6.84 &  0.18 & 22.32 & 26.14 & 13.13 & 16.05\\
 \hline
\multirow{2}{*}{PCWNet} & Expectation & Expectation & 34.27 & 0.19 & 16.80 & 21.36 & 8.93 & 12.62  \\
 & Expectation & L1-Risk & 34.27  & 0.26 &  16.53 & 21.08 &  8.65 & 12.30 \\
 \hline
\multirow{4}{*}{\textbf{Ours}} & Expectation & Expectation & 11.96 & 0.17 & 9.88 & 13.27 & 4.92 & 7.29 \\
 & Expectation & L1-Risk & 11.96 & 0.25 & \textcolor{blue}{9.83} &13.22 & 4.90& 7.27 \\
& L1-Risk & Expectation & 11.96 &  0.17 & \textcolor{blue}{9.83} & \textcolor{blue}{13.19} & \textcolor{blue}{4.79} & \textcolor{blue}{7.06}\\ 
& \textbf{L1-Risk} & \textbf{L1-Risk} & \textbf{11.96} & \textbf{0.25} & \textcolor{red}{\textbf{9.32}} & \textcolor{red}{\textbf{12.63}} & \textbf{\textcolor{red}{4.49}} & \textbf{\textcolor{red}{6.70}}\\
\hline
\end{tabular*}
\end{sc}
\end{small}
\end{center}
\end{table*}

\subsection{Cross-Domain Generalization}
For this experiment, we compare the methods when dealing with environments never seen in the train set. Specifically, all methods are trained only on SceneFlow training set, and then evaluated on the Middlebury, ETH 3D and KITTI 2012 \& 2015 train set, ``without'' fine-tuning. 

The statistical comparison results are shown in Tab.(\ref{tab:zero_mid}), Tab.(\ref{tab:zero_eth3d}), Tab.(\ref{tab:zero_kitti12}), and Tab.(\ref{tab:zero_kitti15}). Our proposed approach achieves the first or the second best accuracies under all the evaluation metrics on the 4 real-world datasets. Particularly, for Middlebury, we reduce the $>1$px error from 13.76 to 12.63. Furthermore, on ETH 3D we reduce $>0.5$px error from 10.39 to 8.59, and $>1$px error from 4.05 to 2.71. Thus, our approach result seems more resilient to cross-domain setting and generalizes better than competing methods. The qualitative comparison is provided in the Appendix.


\begin{table}
\begin{center}
\caption{\small Cross-domain evaluation on KITTI 2012 train set. $\dagger$ denotes using extra data for pre-training. The \textcolor{red}{first} and \textcolor{blue}{second} bests are in red and blue respectively. 
All methods are trained on SceneFlow and evaluated on KITTI 2012 train set without fine-tuning.}
\label{tab:zero_kitti12}
\begin{small}
\begin{sc}
\begin{tabular}{lcccc}
\toprule
\multirow{2}{*}{Method} & \multicolumn{2}{c}{$>2$px} &\multicolumn{2}{c}{$>3$px}\\
 & Noc & All & Noc & All\\
\midrule
CFNet  & 7.08 & 7.97 & 4.66 & 5.31 \\
ACVNet  & 20.34 & 21.44 & 14.22 & 15.18\\
DLNR  & 12.01 & 12.81 & 8.83 & 9.46\\
IGEV  & 7.55 & 8.44 & 5.03 & 5.70\\
PCWNet & \textcolor{blue}{6.63} & \textcolor{blue}{7.49} & \textcolor{blue}{4.08} & \textcolor{blue}{4.68}\\
\midrule
\textbf{Ours} &  \textbf{\textcolor{red}{5.82}}  &  \textbf{\textcolor{red}{6.70}} & \textbf{\textcolor{red}{3.84}} & \textbf{\textcolor{red}{4.43}}\\
\bottomrule
\end{tabular}
\end{sc}
\end{small}
\end{center}
\end{table}

\subsection{Ablation Studies}
We performed ablations to analyze risk minimization effects in disparity prediction.  All the models are trained on SceneFlow and tested on Middlebury without fine-tuning.


\textbf{\textit{(a)} Effect of Risk Minimization.} We compare the expectation, i.e., Eq.\eqref{eq:general_risk} and the $L^1$-norm risk minimization for disparity prediction at train and test time. We present the comparison results in Tab.(\ref{tab:ablation}). Even with expectation minimization at train time, we slightly improve the matching accuracy with $L^1$-norm risk minimization at test time.
Yet, if we use the $L^1$-norm risk minimization at both train time and test time, the best accuracy is achieved under all metrics. 

\textbf{\textit{(b)} Performance with Different Networks.} We replace the disparity prediction method in ACVNet \cite{xu2022attention} and PCWNet \cite{shen2022pcw} from expectation i.e., Eq.\eqref{eq:general_risk} to $L^1$-norm risk minimization \textit{only} during test. The results are shown in Tab.(\ref{tab:ablation}). Our proposed method improves the accuracy under all metrics \textit{without} re-training.

\begin{table}
\begin{center}
\scriptsize
\caption{Cross-domain evaluation on KITTI 2015 train set. $\dagger$ denotes using extra data for pre-training. The \textcolor{red}{first} and \textcolor{blue}{second} bests are in red and blue respectively.  
All methods are trained on SceneFlow and evaluated on KITTI 2015 train set without fine-tuning.}
\label{tab:zero_kitti15}
\begin{small}
\begin{sc}
\begin{tabular}{lccc}
\toprule
\multirow{2}{*}{Method} & \multicolumn{3}{c}{All} \\
 & D1\_bg & D1\_fg & D1\_all \\
\midrule
CFNet  & 4.77 & \textcolor{red}{13.26} & 6.07  \\
ACVNet & 12.35  & 19.97 & 13.52 \\
DLNR & 18.67 & 14.86 & 18.08 \\
IGEV & \textcolor{blue}{4.01}  & 15.58 & \textcolor{blue}{5.79} \\
PCWNet & 4.25 & 14.40 & 5.81  \\
\midrule
\textbf{Ours}& \textbf{\textcolor{red}{3.68}} & \textbf{\textcolor{blue}{13.52}}  & \textbf{\textcolor{red}{5.19}}\\
\bottomrule
\end{tabular}
\end{sc}
\end{small}
\end{center}
\end{table}

\subsection{Network Processing Time \& Paremeters}
We present the networks' inference time and number of parameters in  Tab.(\ref{tab:sceneflow_sota}), Tab.(\ref{tab:kitti12_sota}), Tab.(\ref{tab:kitti15_sota}), and Tab.(\ref{tab:ablation})---cf. Time (s) column. For a fair comparison, all networks are evalutated on the same machine with a GeForce-RTX-3090 GPU. Our network outperforms many state of the arts on inference time, including IGEV and DLNR. Moreover, our network has fewer learnable parameters than PCWNet, IGEV and DLNR. In addition, our proposed $L^1$-norm risk minimization module doesn't require extra learnable parameters. The running time is shown in Tab.(\ref{tab:ablation}). By changing the disparity prediction method from expectation minimization to our proposed approach, the running time increases slightly.

\section{Conclusion}
The paper concludes that continuous end-to-end trainable model for stereo matching is possible with $L^1$ risk minimization formulation. It is shown that the proposed approach is beneficial to multi-modal disparity distributions and outliers and generalizes better on cross-domain stereo images. Stereo Risk is unique in a way that it provides a new way of solving stereo-matching with well-thought-out theoretical arc \cite{vapnik1991principles} and improved results, enabling adaptations from fields such as robotics and control engineering.  
%

\section*{Impact Statement}
We include the following statement, in accordance with the ICML 2024 guidelines outlined at \href{ICLM 2024}{https://icml.cc/Conferences/2024/CallForPapers} in the Impact Statement section.

\noindent
\textbf{Stereo matching}, a key technology in computer vision, promises significant advancements in areas like autonomous vehicles, medical imaging, virtual reality, and robotics, enhancing safety, efficiency, and immersive experiences. The future of stereo matching will significantly impact society, specifically coming from the automation industry and therefore, it requires a balanced approach in its usage that maximizes benefits while mitigating risks, ensuring its development aligns with societal values and needs.

\section*{Acknowledgements}
This work was partly supported by the Alexander von Humboldt Foundation and NSFC 62441204.
\bibliography{arXiv}

\begin{thebibliography}{78}
\providecommand{\natexlab}[1]{#1}
\providecommand{\url}[1]{\texttt{#1}}
\expandafter\ifx\csname urlstyle\endcsname\relax
  \providecommand{\doi}[1]{doi: #1}\else
  \providecommand{\doi}{doi: \begingroup \urlstyle{rm}\Url}\fi

\bibitem[Badki et~al.(2020)Badki, Troccoli, Kim, Kautz, Sen, and Gallo]{badki2020bi3d}
Badki, A., Troccoli, A., Kim, K., Kautz, J., Sen, P., and Gallo, O.
\newblock Bi3d: Stereo depth estimation via binary classifications.
\newblock In \emph{Proceedings of the IEEE/CVF Conference on Computer Vision and Pattern Recognition}, pp.\  1600--1608, 2020.

\bibitem[Berger(2013)]{berger2013statistical}
Berger, J.~O.
\newblock \emph{Statistical decision theory and Bayesian analysis}.
\newblock Springer Science \& Business Media, 2013.

\bibitem[Bimbraw(2015)]{bimbraw2015autonomous}
Bimbraw, K.
\newblock Autonomous cars: Past, present and future a review of the developments in the last century, the present scenario and the expected future of autonomous vehicle technology.
\newblock In \emph{2015 12th international conference on informatics in control, automation and robotics (ICINCO)}, volume~1, pp.\  191--198. IEEE, 2015.

\bibitem[Bleyer et~al.(2011)Bleyer, Rhemann, and Rother]{bleyer2011patchmatch}
Bleyer, M., Rhemann, C., and Rother, C.
\newblock Patchmatch stereo-stereo matching with slanted support windows.
\newblock In \emph{Bmvc}, volume~11, pp.\  1--11, 2011.

\bibitem[Cai et~al.(2020)Cai, Poggi, Mattoccia, and Mordohai]{cai2020matching}
Cai, C., Poggi, M., Mattoccia, S., and Mordohai, P.
\newblock Matching-space stereo networks for cross-domain generalization.
\newblock In \emph{2020 International Conference on 3D Vision (3DV)}, pp.\  364--373. IEEE, 2020.

\bibitem[Chang \& Chen(2018)Chang and Chen]{chang2018pyramid}
Chang, J.-R. and Chen, Y.-S.
\newblock Pyramid stereo matching network.
\newblock In \emph{Proceedings of the IEEE conference on computer vision and pattern recognition}, pp.\  5410--5418, 2018.

\bibitem[Chang et~al.(2023)Chang, Yang, Zhang, and Wang]{chang2023domain}
Chang, T., Yang, X., Zhang, T., and Wang, M.
\newblock Domain generalized stereo matching via hierarchical visual transformation.
\newblock In \emph{Proceedings of the IEEE/CVF Conference on Computer Vision and Pattern Recognition}, pp.\  9559--9568, 2023.

\bibitem[Chen et~al.(2019)Chen, Chen, and Cheng]{Chen_2019_ICCV}
Chen, C., Chen, X., and Cheng, H.
\newblock On the over-smoothing problem of cnn based disparity estimation.
\newblock In \emph{Proceedings of the IEEE/CVF International Conference on Computer Vision (ICCV)}, October 2019.

\bibitem[Chen et~al.(2021)Chen, Li, Wang, Zhang, Li, and Wang]{chen2021cost}
Chen, S., Li, B., Wang, W., Zhang, H., Li, H., and Wang, Z.
\newblock Cost affinity learning network for stereo matching.
\newblock In \emph{ICASSP 2021-2021 IEEE International Conference on Acoustics, Speech and Signal Processing (ICASSP)}, pp.\  2120--2124. IEEE, 2021.

\bibitem[Cheng et~al.(2020)Cheng, Zhong, Harandi, Dai, Chang, Li, Drummond, and Ge]{cheng2020hierarchical}
Cheng, X., Zhong, Y., Harandi, M., Dai, Y., Chang, X., Li, H., Drummond, T., and Ge, Z.
\newblock Hierarchical neural architecture search for deep stereo matching.
\newblock \emph{Advances in Neural Information Processing Systems}, 33:\penalty0 22158--22169, 2020.

\bibitem[Cho et~al.(2014)Cho, Merrienboer, Gulcehre, Bougares, Schwenk, and Bengio]{cho2014learning}
Cho, K., Merrienboer, B., Gulcehre, C., Bougares, F., Schwenk, H., and Bengio, Y.
\newblock Learning phrase representations using rnn encoder-decoder for statistical machine translation.
\newblock In \emph{EMNLP}, 2014.

\bibitem[Chuah et~al.(2022)Chuah, Tennakoon, Hoseinnezhad, Bab-Hadiashar, and Suter]{chuah2022cvpr}
Chuah, W., Tennakoon, R., Hoseinnezhad, R., Bab-Hadiashar, A., and Suter, D.
\newblock Itsa: An information-theoretic approach to automatic shortcut avoidance and domain generalization in stereo matching networks.
\newblock In \emph{Proceedings of the IEEE/CVF Conference on Computer Vision and Pattern Recognition (CVPR)}, pp.\  13022--13032, June 2022.

\bibitem[Cormen et~al.(2009)Cormen, Leiserson, Rivest, and Stein]{cormen2009}
Cormen, T.~H., Leiserson, C.~E., Rivest, R.~L., and Stein, C.
\newblock \emph{Introduction to Algorithms, Third Edition}.
\newblock The MIT Press, 3rd edition, 2009.
\newblock ISBN 0262033844.

\bibitem[Cuturi(2013)]{cuturi2013sinkhorn}
Cuturi, M.
\newblock Sinkhorn distances: Lightspeed computation of optimal transport.
\newblock \emph{Advances in neural information processing systems}, 26, 2013.

\bibitem[Dosovitskiy et~al.(2021)Dosovitskiy, Beyer, Kolesnikov, Weissenborn, Zhai, Unterthiner, Dehghani, Minderer, Heigold, Gelly, Uszkoreit, and Houlsby]{dosovitskiy2021an}
Dosovitskiy, A., Beyer, L., Kolesnikov, A., Weissenborn, D., Zhai, X., Unterthiner, T., Dehghani, M., Minderer, M., Heigold, G., Gelly, S., Uszkoreit, J., and Houlsby, N.
\newblock An image is worth 16x16 words: Transformers for image recognition at scale.
\newblock In \emph{International Conference on Learning Representations}, 2021.

\bibitem[Duggal et~al.(2019)Duggal, Wang, Ma, Hu, and Urtasun]{Duggal2019ICCV}
Duggal, S., Wang, S., Ma, W.-C., Hu, R., and Urtasun, R.
\newblock Deeppruner: Learning efficient stereo matching via differentiable patchmatch.
\newblock In \emph{ICCV}, 2019.

\bibitem[Fan et~al.(2020)Fan, Wang, Bocus, and Pitas]{fan2020computer}
Fan, R., Wang, L., Bocus, M.~J., and Pitas, I.
\newblock Computer stereo vision for autonomous driving.
\newblock \emph{arXiv preprint arXiv:2012.03194}, 2020.

\bibitem[Garg et~al.(2020)Garg, Wang, Hariharan, Campbell, Weinberger, and Chao]{garg2020wasserstein}
Garg, D., Wang, Y., Hariharan, B., Campbell, M., Weinberger, K.~Q., and Chao, W.-L.
\newblock Wasserstein distances for stereo disparity estimation.
\newblock \emph{Advances in Neural Information Processing Systems}, 33:\penalty0 22517--22529, 2020.

\bibitem[Geiger et~al.(2012)Geiger, Lenz, and Urtasun]{geiger2012we}
Geiger, A., Lenz, P., and Urtasun, R.
\newblock Are we ready for autonomous driving? the kitti vision benchmark suite.
\newblock In \emph{2012 IEEE conference on computer vision and pattern recognition}, pp.\  3354--3361. IEEE, 2012.

\bibitem[Gu et~al.(2020)Gu, Fan, Zhu, Dai, Tan, and Tan]{gu2020cascade}
Gu, X., Fan, Z., Zhu, S., Dai, Z., Tan, F., and Tan, P.
\newblock Cascade cost volume for high-resolution multi-view stereo and stereo matching.
\newblock In \emph{Proceedings of the IEEE/CVF conference on computer vision and pattern recognition}, pp.\  2495--2504, 2020.

\bibitem[Guo et~al.(2019)Guo, Yang, Yang, Wang, and Li]{guo2019group}
Guo, X., Yang, K., Yang, W., Wang, X., and Li, H.
\newblock Group-wise correlation stereo network.
\newblock In \emph{Proceedings of the IEEE Conference on Computer Vision and Pattern Recognition}, pp.\  3273--3282, 2019.

\bibitem[He et~al.(2016)He, Zhang, Ren, and Sun]{he2016deep}
He, K., Zhang, X., Ren, S., and Sun, J.
\newblock Deep residual learning for image recognition.
\newblock In \emph{Proceedings of the IEEE conference on computer vision and pattern recognition}, pp.\  770--778, 2016.

\bibitem[Hirschmuller(2007)]{hirschmuller2007stereo}
Hirschmuller, H.
\newblock Stereo processing by semiglobal matching and mutual information.
\newblock \emph{IEEE Transactions on pattern analysis and machine intelligence}, 30\penalty0 (2):\penalty0 328--341, 2007.

\bibitem[Hoff \& Ahuja(1989)Hoff and Ahuja]{hoff1989surfaces}
Hoff, W. and Ahuja, N.
\newblock Surfaces from stereo: Integrating feature matching, disparity estimation, and contour detection.
\newblock \emph{IEEE transactions on pattern analysis and machine intelligence}, 11\penalty0 (2):\penalty0 121--136, 1989.

\bibitem[Jain et~al.(2023)Jain, Kumar, and Van~Gool]{jain2023enhanced}
Jain, N., Kumar, S., and Van~Gool, L.
\newblock Enhanced stable view synthesis.
\newblock In \emph{Proceedings of the IEEE/CVF Conference on Computer Vision and Pattern Recognition}, pp.\  13208--13217, 2023.

\bibitem[Jain et~al.(2024)Jain, Kumar, and Van~Gool]{jain2024learning}
Jain, N., Kumar, S., and Van~Gool, L.
\newblock Learning robust multi-scale representation for neural radiance fields from unposed images.
\newblock \emph{International Journal of Computer Vision}, 132\penalty0 (4):\penalty0 1310--1335, 2024.

\bibitem[Kang et~al.(1995)Kang, Webb, Zitnick, and Kanade]{kang1995}
Kang, S.~B., Webb, J., Zitnick, C., and Kanade, T.
\newblock A multibaseline stereo system with active illumination and real-time image acquisition.
\newblock In \emph{Proceedings of IEEE International Conference on Computer Vision}, pp.\  88--93, 1995.
\newblock \doi{10.1109/ICCV.1995.466802}.

\bibitem[Kaya et~al.(2023)Kaya, Kumar, Oliveira, Ferrari, and Van~Gool]{kaya2023multi}
Kaya, B., Kumar, S., Oliveira, C., Ferrari, V., and Van~Gool, L.
\newblock Multi-view photometric stereo revisited.
\newblock In \emph{Proceedings of the IEEE/CVF Winter Conference on Applications of Computer Vision}, pp.\  3126--3135, 2023.

\bibitem[Kendall et~al.(2017{\natexlab{a}})Kendall, Martirosyan, Dasgupta, Henry, Kennedy, Bachrach, and Bry]{Kendall_2017_ICCV}
Kendall, A., Martirosyan, H., Dasgupta, S., Henry, P., Kennedy, R., Bachrach, A., and Bry, A.
\newblock End-to-end learning of geometry and context for deep stereo regression.
\newblock In \emph{Proceedings of the IEEE International Conference on Computer Vision (ICCV)}, Oct 2017{\natexlab{a}}.

\bibitem[Kendall et~al.(2017{\natexlab{b}})Kendall, Martirosyan, Dasgupta, Henry, Kennedy, Bachrach, and Bry]{kendall2017end}
Kendall, A., Martirosyan, H., Dasgupta, S., Henry, P., Kennedy, R., Bachrach, A., and Bry, A.
\newblock End-to-end learning of geometry and context for deep stereo regression.
\newblock In \emph{Proceedings of the IEEE international conference on computer vision}, pp.\  66--75, 2017{\natexlab{b}}.

\bibitem[Kim et~al.(2021)Kim, Lee, Kim, Kim, Lee, and Choi]{kim2021stereo}
Kim, W.-S., Lee, D.-H., Kim, Y.-J., Kim, T., Lee, W.-S., and Choi, C.-H.
\newblock Stereo-vision-based crop height estimation for agricultural robots.
\newblock \emph{Computers and Electronics in Agriculture}, 181:\penalty0 105937, 2021.

\bibitem[Klaus et~al.(2006)Klaus, Sormann, and Karner]{klaus2006segment}
Klaus, A., Sormann, M., and Karner, K.
\newblock Segment-based stereo matching using belief propagation and a self-adapting dissimilarity measure.
\newblock In \emph{18th International Conference on Pattern Recognition (ICPR'06)}, volume~3, pp.\  15--18. IEEE, 2006.

\bibitem[Kolmogorov \& Zabih(2001)Kolmogorov and Zabih]{kolmogorov2001computing}
Kolmogorov, V. and Zabih, R.
\newblock Computing visual correspondence with occlusions using graph cuts.
\newblock In \emph{Proceedings Eighth IEEE International Conference on Computer Vision. ICCV 2001}, volume~2, pp.\  508--515. IEEE, 2001.

\bibitem[Krantz \& Parks(2002)Krantz and Parks]{krantz2002implicit}
Krantz, S.~G. and Parks, H.~R.
\newblock \emph{The implicit function theorem: history, theory, and applications}.
\newblock Springer Science \& Business Media, 2002.

\bibitem[Lehmann \& Casella(1998)Lehmann and Casella]{LehmCase98}
Lehmann, E.~L. and Casella, G.
\newblock \emph{Theory of Point Estimation}.
\newblock Springer-Verlag, New York, NY, USA, second edition, 1998.

\bibitem[Li et~al.(2022)Li, Wang, Xiong, Cai, Yan, Yang, Liu, Fan, and Liu]{li2022practical}
Li, J., Wang, P., Xiong, P., Cai, T., Yan, Z., Yang, L., Liu, J., Fan, H., and Liu, S.
\newblock Practical stereo matching via cascaded recurrent network with adaptive correlation.
\newblock In \emph{Proceedings of the IEEE/CVF conference on computer vision and pattern recognition}, pp.\  16263--16272, 2022.

\bibitem[Li et~al.(2021)Li, Liu, Drenkow, Ding, Creighton, Taylor, and Unberath]{Li_2021_ICCV}
Li, Z., Liu, X., Drenkow, N., Ding, A., Creighton, F.~X., Taylor, R.~H., and Unberath, M.
\newblock Revisiting stereo depth estimation from a sequence-to-sequence perspective with transformers.
\newblock In \emph{Proceedings of the IEEE/CVF International Conference on Computer Vision (ICCV)}, pp.\  6197--6206, October 2021.

\bibitem[Lipson et~al.(2021)Lipson, Teed, and Deng]{lipson2021raft}
Lipson, L., Teed, Z., and Deng, J.
\newblock Raft-stereo: Multilevel recurrent field transforms for stereo matching.
\newblock In \emph{2021 International Conference on 3D Vision (3DV)}, pp.\  218--227. IEEE, 2021.

\bibitem[Liu et~al.(2022{\natexlab{a}})Liu, Yu, and Long]{liu2022local}
Liu, B., Yu, H., and Long, Y.
\newblock Local similarity pattern and cost self-reassembling for deep stereo matching networks.
\newblock In \emph{Proceedings of the AAAI Conference on Artificial Intelligence}, volume~36, pp.\  1647--1655, 2022{\natexlab{a}}.

\bibitem[Liu et~al.(2022{\natexlab{b}})Liu, Yu, and Qi]{liu2022graftnet}
Liu, B., Yu, H., and Qi, G.
\newblock Graftnet: Towards domain generalized stereo matching with a broad-spectrum and task-oriented feature.
\newblock In \emph{Proceedings of the IEEE/CVF Conference on Computer Vision and Pattern Recognition}, pp.\  13012--13021, 2022{\natexlab{b}}.

\bibitem[Liu et~al.(2023{\natexlab{a}})Liu, Kumar, Gu, Timofte, and Van~Gool]{liu2023single}
Liu, C., Kumar, S., Gu, S., Timofte, R., and Van~Gool, L.
\newblock Single image depth prediction made better: A multivariate gaussian take.
\newblock In \emph{Proceedings of the IEEE/CVF Conference on Computer Vision and Pattern Recognition}, pp.\  17346--17356, 2023{\natexlab{a}}.

\bibitem[Liu et~al.(2023{\natexlab{b}})Liu, Kumar, Gu, Timofte, and Van~Gool]{liu2023va}
Liu, C., Kumar, S., Gu, S., Timofte, R., and Van~Gool, L.
\newblock Va-depthnet: A variational approach to single image depth prediction.
\newblock \emph{arXiv preprint arXiv:2302.06556}, 2023{\natexlab{b}}.

\bibitem[Liu et~al.(2020)Liu, Yang, Sun, Wang, and Li]{Liu_2020_StereoGAN}
Liu, R., Yang, C., Sun, W., Wang, X., and Li, H.
\newblock Stereogan: Bridging synthetic-to-real domain gap by joint optimization of domain translation and stereo matching.
\newblock In \emph{The IEEE Conference on Computer Vision and Pattern Recognition (CVPR)}, June 2020.

\bibitem[Loshchilov \& Hutter(2019)Loshchilov and Hutter]{loshchilov2018decoupled}
Loshchilov, I. and Hutter, F.
\newblock Decoupled weight decay regularization.
\newblock In \emph{International Conference on Learning Representations}, 2019.

\bibitem[Luo et~al.(2020)Luo, Li, Lin, Chen, Lee, Choi, Yoo, and Polley]{luo2020wavelet}
Luo, C., Li, Y., Lin, K., Chen, G., Lee, S.-J., Choi, J., Yoo, Y.~F., and Polley, M.~O.
\newblock Wavelet synthesis net for disparity estimation to synthesize dslr calibre bokeh effect on smartphones.
\newblock In \emph{Proceedings of the IEEE/CVF Conference on Computer Vision and Pattern Recognition}, pp.\  2407--2415, 2020.

\bibitem[Mayer et~al.(2016)Mayer, Ilg, Hausser, Fischer, Cremers, Dosovitskiy, and Brox]{mayer2016large}
Mayer, N., Ilg, E., Hausser, P., Fischer, P., Cremers, D., Dosovitskiy, A., and Brox, T.
\newblock A large dataset to train convolutional networks for disparity, optical flow, and scene flow estimation.
\newblock In \emph{Proceedings of the IEEE conference on computer vision and pattern recognition}, pp.\  4040--4048, 2016.

\bibitem[Menze \& Geiger(2015)Menze and Geiger]{menze2015object}
Menze, M. and Geiger, A.
\newblock Object scene flow for autonomous vehicles.
\newblock In \emph{Proceedings of the IEEE conference on computer vision and pattern recognition}, pp.\  3061--3070, 2015.

\bibitem[Meuleman et~al.(2022)Meuleman, Kim, Tompkin, and Kim]{meuleman2022floatingfusion}
Meuleman, A., Kim, H., Tompkin, J., and Kim, M.~H.
\newblock Floatingfusion: Depth from tof and image-stabilized stereo cameras.
\newblock In \emph{European Conference on Computer Vision}, pp.\  602--618. Springer, 2022.

\bibitem[Newell et~al.(2016)Newell, Yang, and Deng]{newell2016stacked}
Newell, A., Yang, K., and Deng, J.
\newblock Stacked hourglass networks for human pose estimation.
\newblock In \emph{Computer Vision--ECCV 2016: 14th European Conference, Amsterdam, The Netherlands, October 11-14, 2016, Proceedings, Part VIII 14}, pp.\  483--499. Springer, 2016.

\bibitem[Pang et~al.(2018)Pang, Sun, Yang, Ren, Xiao, Zeng, and Lin]{pang2018zoom}
Pang, J., Sun, W., Yang, C., Ren, J., Xiao, R., Zeng, J., and Lin, L.
\newblock Zoom and learn: Generalizing deep stereo matching to novel domains.
\newblock In \emph{Proceedings of the IEEE Conference on Computer Vision and Pattern Recognition}, pp.\  2070--2079, 2018.

\bibitem[Peng et~al.(2022)Peng, Wang, Wang, Lai, and Wang]{unimvsnet}
Peng, R., Wang, R., Wang, Z., Lai, Y., and Wang, R.
\newblock Rethinking depth estimation for multi-view stereo: A unified representation.
\newblock In \emph{Proceedings of the IEEE Conference on Computer Vision and Pattern Recognition (CVPR)}, 2022.

\bibitem[Rao et~al.(2022)Rao, Dai, Shen, and He]{rao2022rethinking}
Rao, Z., Dai, Y., Shen, Z., and He, R.
\newblock Rethinking training strategy in stereo matching.
\newblock \emph{IEEE Transactions on Neural Networks and Learning Systems}, pp.\  1--14, 2022.

\bibitem[Scharstein \& Szeliski(2002)Scharstein and Szeliski]{scharstein2002taxonomy}
Scharstein, D. and Szeliski, R.
\newblock A taxonomy and evaluation of dense two-frame stereo correspondence algorithms.
\newblock \emph{International journal of computer vision}, 47:\penalty0 7--42, 2002.

\bibitem[Sch\"ops et~al.(2017)Sch\"ops, Sch\"onberger, Galliani, Sattler, Schindler, Pollefeys, and Geiger]{schoeps2017cvpr}
Sch\"ops, T., Sch\"onberger, J.~L., Galliani, S., Sattler, T., Schindler, K., Pollefeys, M., and Geiger, A.
\newblock A multi-view stereo benchmark with high-resolution images and multi-camera videos.
\newblock In \emph{Conference on Computer Vision and Pattern Recognition (CVPR)}, 2017.

\bibitem[Shen et~al.(2021)Shen, Dai, and Rao]{shen2021cfnet}
Shen, Z., Dai, Y., and Rao, Z.
\newblock Cfnet: Cascade and fused cost volume for robust stereo matching.
\newblock In \emph{Proceedings of the IEEE/CVF Conference on Computer Vision and Pattern Recognition}, pp.\  13906--13915, 2021.

\bibitem[Shen et~al.(2022)Shen, Dai, Song, Rao, Zhou, and Zhang]{shen2022pcw}
Shen, Z., Dai, Y., Song, X., Rao, Z., Zhou, D., and Zhang, L.
\newblock Pcw-net: Pyramid combination and warping cost volume for stereo matching.
\newblock In \emph{European Conference on Computer Vision}, pp.\  280--297. Springer, 2022.

\bibitem[Song et~al.(2021)Song, Yang, Zhu, Zhou, Wang, and Shi]{Song_2021_CVPR}
Song, X., Yang, G., Zhu, X., Zhou, H., Wang, Z., and Shi, J.
\newblock Adastereo: A simple and efficient approach for adaptive stereo matching.
\newblock In \emph{Proceedings of the IEEE/CVF Conference on Computer Vision and Pattern Recognition (CVPR)}, pp.\  10328--10337, June 2021.

\bibitem[Szeliski(2022)]{szeliski2022computer}
Szeliski, R.
\newblock \emph{Computer vision: algorithms and applications}.
\newblock Springer Nature, 2022.

\bibitem[Tankovich et~al.(2021)Tankovich, Hane, Zhang, Kowdle, Fanello, and Bouaziz]{tankovich2021hitnet}
Tankovich, V., Hane, C., Zhang, Y., Kowdle, A., Fanello, S., and Bouaziz, S.
\newblock Hitnet: Hierarchical iterative tile refinement network for real-time stereo matching.
\newblock In \emph{Proceedings of the IEEE/CVF Conference on Computer Vision and Pattern Recognition}, pp.\  14362--14372, 2021.

\bibitem[Tonioni et~al.(2017)Tonioni, Poggi, Mattoccia, and Di~Stefano]{Tonioni_2017_ICCV}
Tonioni, A., Poggi, M., Mattoccia, S., and Di~Stefano, L.
\newblock Unsupervised adaptation for deep stereo.
\newblock In \emph{Proceedings of the IEEE International Conference on Computer Vision (ICCV)}, Oct 2017.

\bibitem[Tonioni et~al.(2019{\natexlab{a}})Tonioni, Rahnama, Joy, Stefano, Ajanthan, and Torr]{tonioni2019learning}
Tonioni, A., Rahnama, O., Joy, T., Stefano, L.~D., Ajanthan, T., and Torr, P.~H.
\newblock Learning to adapt for stereo.
\newblock In \emph{Proceedings of the IEEE/CVF Conference on Computer Vision and Pattern Recognition}, pp.\  9661--9670, 2019{\natexlab{a}}.

\bibitem[Tonioni et~al.(2019{\natexlab{b}})Tonioni, Tosi, Poggi, Mattoccia, and Stefano]{Tonioni_2019_CVPR}
Tonioni, A., Tosi, F., Poggi, M., Mattoccia, S., and Stefano, L.~D.
\newblock Real-time self-adaptive deep stereo.
\newblock In \emph{Proceedings of the IEEE/CVF Conference on Computer Vision and Pattern Recognition (CVPR)}, June 2019{\natexlab{b}}.

\bibitem[Tosi et~al.(2021)Tosi, Liao, Schmitt, and Geiger]{Tosi_2021_CVPR}
Tosi, F., Liao, Y., Schmitt, C., and Geiger, A.
\newblock Smd-nets: Stereo mixture density networks.
\newblock In \emph{Proceedings of the IEEE/CVF Conference on Computer Vision and Pattern Recognition (CVPR)}, pp.\  8942--8952, June 2021.

\bibitem[Vapnik(1991)]{vapnik1991principles}
Vapnik, V.
\newblock Principles of risk minimization for learning theory.
\newblock \emph{Advances in neural information processing systems}, 4, 1991.

\bibitem[Vaswani et~al.(2017)Vaswani, Shazeer, Parmar, Uszkoreit, Jones, Gomez, Kaiser, and Polosukhin]{vaswani2017attention}
Vaswani, A., Shazeer, N., Parmar, N., Uszkoreit, J., Jones, L., Gomez, A.~N., Kaiser, L.~u., and Polosukhin, I.
\newblock Attention is all you need.
\newblock In Guyon, I., Luxburg, U.~V., Bengio, S., Wallach, H., Fergus, R., Vishwanathan, S., and Garnett, R. (eds.), \emph{Advances in Neural Information Processing Systems}, volume~30. Curran Associates, Inc., 2017.

\bibitem[Wang et~al.(2019)Wang, Lai, Huang, Wang, Van Der~Maaten, Campbell, and Weinberger]{wang2019anytime}
Wang, Y., Lai, Z., Huang, G., Wang, B.~H., Van Der~Maaten, L., Campbell, M., and Weinberger, K.~Q.
\newblock Anytime stereo image depth estimation on mobile devices.
\newblock In \emph{2019 international conference on robotics and automation (ICRA)}, pp.\  5893--5900. IEEE, 2019.

\bibitem[Weinzaepfel et~al.(2023)Weinzaepfel, Lucas, Leroy, Cabon, Arora, Br{\'e}gier, Csurka, Antsfeld, Chidlovskii, and Revaud]{philippe2023croco}
Weinzaepfel, P., Lucas, T., Leroy, V., Cabon, Y., Arora, V., Br{\'e}gier, R., Csurka, G., Antsfeld, L., Chidlovskii, B., and Revaud, J.
\newblock {CroCo v2: Improved Cross-view Completion Pre-training for Stereo Matching and Optical Flow}.
\newblock In \emph{ICCV}, 2023.

\bibitem[Xu et~al.(2022)Xu, Cheng, Guo, and Yang]{xu2022attention}
Xu, G., Cheng, J., Guo, P., and Yang, X.
\newblock Attention concatenation volume for accurate and efficient stereo matching.
\newblock In \emph{Proceedings of the IEEE/CVF Conference on Computer Vision and Pattern Recognition}, pp.\  12981--12990, 2022.

\bibitem[Xu et~al.(2023)Xu, Wang, Ding, and Yang]{xu2023iterative}
Xu, G., Wang, X., Ding, X., and Yang, X.
\newblock Iterative geometry encoding volume for stereo matching.
\newblock In \emph{Proceedings of the IEEE/CVF Conference on Computer Vision and Pattern Recognition}, pp.\  21919--21928, 2023.

\bibitem[Xu \& Zhang(2020)Xu and Zhang]{xu2020aanet}
Xu, H. and Zhang, J.
\newblock Aanet: Adaptive aggregation network for efficient stereo matching.
\newblock In \emph{Proceedings of the IEEE/CVF Conference on Computer Vision and Pattern Recognition}, pp.\  1959--1968, 2020.

\bibitem[Yamaguchi et~al.(2014)Yamaguchi, McAllester, and Urtasun]{yamaguchi2014efficient}
Yamaguchi, K., McAllester, D., and Urtasun, R.
\newblock Efficient joint segmentation, occlusion labeling, stereo and flow estimation.
\newblock In \emph{Computer Vision--ECCV 2014: 13th European Conference, Zurich, Switzerland, September 6-12, 2014, Proceedings, Part V 13}, pp.\  756--771. Springer, 2014.

\bibitem[Yang et~al.(2022)Yang, Alvarez, and Liu]{yang2022non}
Yang, J., Alvarez, J.~M., and Liu, M.
\newblock Non-parametric depth distribution modelling based depth inference for multi-view stereo.
\newblock In \emph{Proceedings of the IEEE/CVF Conference on Computer Vision and Pattern Recognition}, pp.\  8626--8634, 2022.

\bibitem[Zbontar \& LeCun(2015)Zbontar and LeCun]{Zbontar_2015_CVPR}
Zbontar, J. and LeCun, Y.
\newblock Computing the stereo matching cost with a convolutional neural network.
\newblock In \emph{Proceedings of the IEEE Conference on Computer Vision and Pattern Recognition (CVPR)}, June 2015.

\bibitem[Zhang et~al.(2019)Zhang, Prisacariu, Yang, and Torr]{Zhang2019GANet}
Zhang, F., Prisacariu, V., Yang, R., and Torr, P.~H.
\newblock Ga-net: Guided aggregation net for end-to-end stereo matching.
\newblock In \emph{Proceedings of the IEEE Conference on Computer Vision and Pattern Recognition}, pp.\  185--194, 2019.

\bibitem[Zhang et~al.(2020)Zhang, Qi, Yang, Prisacariu, Wah, and Torr]{zhang2020domain}
Zhang, F., Qi, X., Yang, R., Prisacariu, V., Wah, B., and Torr, P.
\newblock Domain-invariant stereo matching networks.
\newblock In \emph{Computer Vision--ECCV 2020: 16th European Conference, Glasgow, UK, August 23--28, 2020, Proceedings, Part II 16}, pp.\  420--439. Springer, 2020.

\bibitem[Zhang et~al.(2022)Zhang, Wang, Bai, Wang, Huang, Chen, Gu, Zhou, Harada, and Hancock]{zhang2022revisiting}
Zhang, J., Wang, X., Bai, X., Wang, C., Huang, L., Chen, Y., Gu, L., Zhou, J., Harada, T., and Hancock, E.~R.
\newblock Revisiting domain generalized stereo matching networks from a feature consistency perspective.
\newblock In \emph{Proceedings of the IEEE/CVF Conference on Computer Vision and Pattern Recognition}, pp.\  13001--13011, 2022.

\bibitem[Zhao et~al.(2017)Zhao, Shi, Qi, Wang, and Jia]{zhao2017pyramid}
Zhao, H., Shi, J., Qi, X., Wang, X., and Jia, J.
\newblock Pyramid scene parsing network.
\newblock In \emph{Proceedings of the IEEE conference on computer vision and pattern recognition}, pp.\  2881--2890, 2017.

\bibitem[Zhao et~al.(2023)Zhao, Zhou, Zhang, Chen, Yang, and Zhao]{zhao2023high}
Zhao, H., Zhou, H., Zhang, Y., Chen, J., Yang, Y., and Zhao, Y.
\newblock High-frequency stereo matching network.
\newblock In \emph{Proceedings of the IEEE/CVF Conference on Computer Vision and Pattern Recognition}, pp.\  1327--1336, 2023.

\end{thebibliography}
\bibliographystyle{icml2024}

\newpage
\appendix
\onecolumn
\section{Training Details}
We train our network on SceneFlow. The weight is initialized randomly. We use AdamW optimizer \citep{loshchilov2018decoupled} with weight decay $10^{-5}$. The learning rate decreases from $2\times10^{-4}$ to $2\times10^{-8}$ according to the one cycle learning rate policy. We train the network for $2\times 10^{5}$ iterations. The images will be randomly cropped to $320\times736$. For KITTI 2012 \& 2015 benchmarks, we further fine tune the network on the training image pairs for $2.5\times10^{3}$ iterations. The learning rate starts from $5\times10^{-5}$ to $5\times10^{-9}$. 
 
 Following RAFT-Stereo \citep{lipson2021raft}, we apply various image augmentations during training to avoid the over-fitting problem. Specifically, the augmentations include \textit{(a)} color transformation, \textit{(b)} occlusion, and \textit{(c)} spatial transformation.  In \textit{(a)} color transformation, we randomly change the brightness, contrast, saturation and hue of the left and right images independently. The brightness and contrast factors are uniformly chosen from [0.6, 1.4]. The saturation factor is uniformly chosen from [0.0, 1.4]. The hue factor is uniformly chosen from [-0.16, 0.16]. In \textit{(b)} occlusion, we randomly select a few rectangular regions in the right image, and set the pixels inside the regions as the mean color of the right image. The number of regions is chosen from \{0, 1, 2, 3\} with probabilities \{0.5, 0.166, 0.166, 0.166\}. The position of the region is uniformly chosen in the right image, and the width and height are uniformly chosen from [50, 100]. In \textit{(c)} spatial transformation, we randomly crop the left and right images to the resolution 320$\times$736. 
\section{Network Structure Details}
In this part, we present more details for the \textit{(i)} feature extraction and \textit{(ii)} cost aggregation.

\textbf{\textit{(i)} Feature Extraction.} Given an input image, the module aims to output multi-scale 2D feature maps. More specifically, we first use a ResNet \citep{he2016deep} to extract 2D feature maps of resolution 1/4 and 1/2 with respect to the input image. The ResNet contains 4 stages of non-linear transformation with 3, 16, 3, 3 residual blocks respectively, where each block is composed of convolutional layers and skip connections. And the spatial resolution is downsampled before the beginning of the first and third stages of transformation. Then we apply the spatial pyramid pooling \citep{zhao2017pyramid} on the 1/4-resolution feature map from the fourth stage of transformation to enlarge the receptive field. In the end, we upsample the enhanced feature map from 1/4 to 1/2 and fuse it with the 1/2-resolution feature map from the first stage of transformation in ResNet. The final outputs are the feature maps of 1/4 and 1/2 resolution. We apply the same network and weights to extract features from left and right images. The details of the network structure and the resolution of the feature maps are shown in Tab.(\ref{tab:feature_extraction_structure}).\\
\begin{table}
\begin{center}
\scriptsize
\caption{\small Network structure for feature extraction. 
}
\label{tab:feature_extraction_structure}
\setlength\tabcolsep{5.5pt}
\begin{tabular*}{0.8\textwidth}{l@{\extracolsep{\fill}}cc}
\hline
Name & Layer Setting & Output Dimension\\
\hline
\multicolumn{3}{c}{ResNet}\\
\hline
Input &  & $H\times W \times 3$\\
\hline
Stem-1 & $3\times3, 32$ & $H \times W \times 32$\\
Stem-2 & $3\times3, 32$ & $H \times W \times 32$\\
Stem-3 & $3\times3, 32$ & $\frac{1}{2} H \times \frac{1}{2} W \times 32$\\
\hline
Stage-1 & $\left[\begin{array}{c} 3\times3, 32 \\ 3\times3, 32 \end{array}\right]\times 3$ & $\frac{1}{2} H \times \frac{1}{2} W \times 32$\\
\hline
Stage-2 & $\left[\begin{array}{c} 3\times3, 64 \\ 3\times3, 64 \end{array}\right]\times 16$ & $\frac{1}{4} H \times \frac{1}{4} W \times 64$\\
\hline
Stage-3 & $\left[\begin{array}{c} 3\times3, 128 \\ 3\times3, 128 \end{array}\right]\times 3$ & $\frac{1}{4} H \times \frac{1}{4} W \times 128$\\
\hline
Stage-4 & $\left[\begin{array}{c} 3\times3, 128 \\ 3\times3, 128 \end{array}\right]\times 3$, $\mathtt{dila}=2$ & $\frac{1}{4} H \times \frac{1}{4} W \times 128$\\
\hline
\multicolumn{3}{c}{Spatial Pyramid Pooling}\\
\hline
\multirow{3}{*}{Branch-1} & $64\times64\quad\mathtt{avg\; pool}$ & \multirow{3}{*}{$\frac{1}{4} H \times \frac{1}{4} W \times 32$}\\
&$3\times3, 32$ &\\
&$\mathtt{bilinear\;interpolation}$ & \\
\hline
\multirow{3}{*}{Branch-2} & $32\times32\quad\mathtt{avg\; pool}$ & \multirow{3}{*}{$\frac{1}{4} H \times \frac{1}{4} W \times 32$}\\
&$3\times3, 32$ &\\
&$\mathtt{bilinear\;interpolation}$ & \\
\hline
\multirow{3}{*}{Branch-3} & $16\times16\quad\mathtt{avg\; pool}$ & \multirow{3}{*}{$\frac{1}{4} H \times \frac{1}{4} W \times 32$}\\
&$3\times3, 32$ &\\
&$\mathtt{bilinear\;interpolation}$ & \\
\hline
\multirow{3}{*}{Branch-4} & $8\times8\quad\mathtt{avg\; pool}$ & \multirow{3}{*}{$\frac{1}{4} H \times \frac{1}{4} W \times 32$}\\
&$3\times3, 32$ &\\
&$\mathtt{bilinear\;interpolation}$ & \\
\hline
\multicolumn{2}{c}{Concat [Stage-2, Stage-4, Branch-1, Branch-2, Branch-3, Branch-4]} &$\frac{1}{4} H \times \frac{1}{4} W \times 32$ \\
\hline
\multirow{2}{*}{Fusion-1} & $3\times3, 128$ &  \multirow{2}{*}{$\frac{1}{4} H \times \frac{1}{4} W \times 32$}\\
& $1\times1, 32$ & \\
\hline
\multicolumn{3}{c}{UpSample}\\
\hline
Up-1 & $\mathtt{nearest\; interpolation}$ & $\frac{1}{2} H \times \frac{1}{2} W \times 32$\\
\hline
\multicolumn{2}{c}{Add [Stage-1, Up-0]} &$\frac{1}{2} H \times \frac{1}{2} W \times 32$ \\
\hline
Fusion-2 & $3\times3, 16$ & $\frac{1}{2} H \times \frac{1}{2} W \times 16$\\
\hline
\end{tabular*}
\end{center}
\end{table}
\textbf{\textit{(ii)} Cost Aggregation.} We use the stacked hourglass architecture \citep{newell2016stacked} to transform the stereo cost volume and aggregate the matching cost. For the coarse and refined stages, the structures are the same except for the number of feature channels. Specifically, the network consists of three 3D hourglasss as in \citet{chang2018pyramid}. Each hourglass first downsamples the volume hierarchically to 1/2 and 1/4 resolution with respect to the input volume, and then upsamples in sequence to recover the resolution. The above procedure helps aggregate the matching information across various scales. The final output is a volume that represents the discrete distribution of disparity hypotheses. We present the details of a single hourglass structure in Tab.(\ref{tab:hourglass_structure}). For an input image with resolution $h\times w$, the $D$, $H$, $W$, $C$ are 192, $h/4$, $w/4$, 32 respectively in the coarse stage. In the refined stage, we set $D$, $H$, $W$, $C$ to be 16, $h/2$, $w/2$, 16 respectively.\\

\begin{table}
\begin{center}
\scriptsize
\caption{\small Network structure for 3D hourglass. 
}
\label{tab:hourglass_structure}
\setlength\tabcolsep{5.5pt}
\begin{tabular*}{0.8\textwidth}{l@{\extracolsep{\fill}}cc}
\hline
Name & Layer Setting & Output Dimension\\
\hline
Input &  & $D\times H\times W \times C$\\
\hline
Conv-1 & $3\times3\times3, 2C$ & $\frac{1}{2}D\times\frac{1}{2}H \times \frac{1}{2}W \times 2C$\\
\hline
Conv-2 & $3\times3\times3, 2C$ & $\frac{1}{2}D\times\frac{1}{2}H \times \frac{1}{2}W \times 2C$\\
\hline
Conv-3 & $3\times3\times3, 4C$ & $\frac{1}{4}D\times\frac{1}{4}H \times \frac{1}{4}W \times 4C$\\
\hline
Conv-4 & $3\times3\times3, 4C$ & $\frac{1}{4}D\times\frac{1}{4}H \times \frac{1}{4}W \times 4C$\\
\hline
\multirow{4}{*}{Atte-4} & $3\times3\times3, C$ & \multirow{4}{*}{$\frac{1}{4}D\times\frac{1}{4}H \times \frac{1}{4}W \times 4C$}\\
&$3\times3\times3, 4C$ &\\
&$\mathtt{sigmoid}$ &\\
& $\mathtt{prod}$ Conv-4& \\
\hline
\multirow{2}{*}{Conv-5} & $\mathtt{deconv}\;3\times3\times3, 2C$ & \multirow{2}{*}{$\frac{1}{2}D\times\frac{1}{2}H \times \frac{1}{2}W \times 2C$}\\
& $\mathtt{add}$ Conv-2& \\
\hline
\multirow{4}{*}{Atte-5} & $3\times3\times3, C$ & \multirow{4}{*}{$\frac{1}{2}D\times\frac{1}{2}H \times \frac{1}{2}W \times 2C$}\\
&$3\times3\times3, 2C$ &\\
&$\mathtt{sigmoid}$ &\\
& $\mathtt{prod}$ Conv-5& \\
\hline
\multirow{2}{*}{Conv-6} & $\mathtt{deconv}\;3\times3\times3, C$ & \multirow{2}{*}{$D\times H \times W \times C$}\\
& $\mathtt{add}$ Input& \\
\hline
\multirow{4}{*}{Atte-6} & $3\times3\times3, C$ & \multirow{4}{*}{$D\times H \times W \times C$}\\
&$3\times3\times3, C$ &\\
&$\mathtt{sigmoid}$ &\\
& $\mathtt{prod}$ Conv-6& \\
\hline
\end{tabular*}
\end{center}
\end{table}

\section{Experiments}
\subsection{Ablation Study for Tolerance}
In this part, we change the value of the tolerance $\tau$ in the binary search algorithm and observe its effects. As shown in Tab.(\ref{tab:tolerance_ablation}), when decreasing the value of $\tau$, the search algorithm will iterate for more times to search for the optimal solution. And the error of the predicted disparity is reduced. When $\tau\geq0.1$, the algorithm achieves the best accuracy. 
\begin{table}[h]
\begin{center}
\scriptsize
\caption{\small Ablation studies for tolerance $\tau$ on Middlebury training set of quarter resolution. The \textcolor{red}{first} and \textcolor{blue}{second} bests are in red and blue respectively. \textbf{Our method} in bold. All settings are trained on SceneFlow and evaluated on Middlebury training set without fine-tuning.}
\label{tab:tolerance_ablation}
\setlength\tabcolsep{5.5pt}
\begin{tabular*}{1.0\textwidth}{c@{\extracolsep{\fill}}cccccc}
\hline
\multirow{2}{*}{Tolerance $\tau$} & \multirow{2}{*}{Number of Iterations} & \multicolumn{2}{c}{$>1$px} &\multicolumn{2}{c}{$>2$px}\\
& & Noc & All & Noc & All\\
\hline
0.3 & 9 & \textcolor{blue}{9.36} & \textcolor{blue}{12.67} & \textcolor{blue}{4.50} & \textcolor{blue}{6.71} \\
 \hline
\textbf{0.1} & \textbf{11} & \textbf{\textcolor{red}{9.32}} & \textbf{\textcolor{red}{12.63}} &  \textbf{\textcolor{red}{4.49}} & \textbf{\textcolor{red}{6.70}}\\
\hline
0.01 & 14 & \textcolor{red}{9.32} & \textcolor{red}{12.63} & \textcolor{red}{4.49} & \textcolor{red}{6.70}\\
\hline
\end{tabular*}
\end{center}
\end{table}

\subsection{Ablation Studies for Huber Loss}
In this part, we evaluate the effects of different loss functions. In Tab.(\ref{tab:loss_ablation}), we evaluate the $L^2$ loss, the $L^1$ loss, and the Huber loss, i.e. a combination of $L^1$ and $L^2$ norm depending on the thresholding value $\beta$.  
The table clearly shows the benefit of using risk minimization loss under $L^1$.

\begin{table}[h]
\begin{center}
\scriptsize
\caption{\small Ablation studies for loss function on Middlebury training set of quarter resolution. The \textcolor{red}{first} and \textcolor{blue}{second} bests are in red and blue respectively. \textbf{Our method} in bold. 
All settings are trained on SceneFlow and evaluated on Middlebury training set without fine-tuning.}
\label{tab:loss_ablation}
\setlength\tabcolsep{5.5pt}
\begin{tabular*}{1.0\textwidth}{c@{\extracolsep{\fill}}cccc}
\hline
\multirow{2}{*}{Loss} & \multicolumn{2}{c}{$>1$px} &\multicolumn{2}{c}{$>2$px}\\
& Noc & All & Noc & All\\
\hline
$L^2$ & 9.83 & 13.19 & 4.79 & 7.06 \\
\hline
$\beta=10.0$ & 9.41 & 12.73 & 4.55 & 6.76 \\
 \hline
$\beta=4.0$ & 9.36 & 12.68 &  4.51 & \textcolor{blue}{6.72}\\
\hline
$\beta=1.0$ & \textcolor{blue}{9.33} & \textcolor{blue}{12.64} & \textcolor{blue}{4.50} & \textcolor{red}{6.70}\\
\hline
$\bm{L^1}$ & \textbf{\textcolor{red}{9.32}} & \textbf{\textcolor{red}{12.63}} & \textbf{\textcolor{red}{4.49}} & \textbf{\textcolor{red}{6.70}} \\
\hline
\end{tabular*}
\end{center}
\end{table}

\subsection{Ablation Studies for Network Architectures}
In this part, we apply our method to the IGEV \citep{xu2023iterative} framework. Specifically, we use our method to compute the initial disparities from the geometry encoding volume. The results are shown in Tab.(\ref{tab:igev_ablation}). Our method improves the accuracy of IGEV.
\begin{table}[h]
\begin{center}
\scriptsize
\caption{\small Ablation studies for IGEV on Middlebury training set of quarter resolution. The \textcolor{red}{first} and \textcolor{blue}{second} bests are in red and blue respectively. \textbf{Our method} in bold. 
All methods are trained on SceneFlow and evaluated on Middlebury training set without fine-tuning.}
\label{tab:igev_ablation}
\setlength\tabcolsep{5.5pt}
\begin{tabular*}{1.0\textwidth}{l@{\extracolsep{\fill}}cccccccc}
\hline
\multirow{2}{*}{Backbone} &  \multirow{2}{*}{Training}&\multirow{2}{*}{Test} & \multirow{2}{*}{Param (M)} & \multirow{2}{*}{Time (s)} & \multicolumn{2}{c}{$>3$px} &\multicolumn{2}{c}{$>4$px}\\
& & & & & Noc & All & Noc & All\\
\hline
\multirow{2}{*}{IGEV \citep{xu2023iterative}} & Expectation & Expectation & 12.60 & 0.34 & \textcolor{blue}{4.47} & \textcolor{blue}{6.64} & \textcolor{blue}{3.46} & \textcolor{red}{5.32}\\
 & \textbf{Expectation} & \textbf{L1-Risk} & \textbf{12.60} &  \textbf{0.38} & \textbf{\textcolor{red}{4.37}} & \textbf{\textcolor{red}{6.63}} & \textbf{\textcolor{red}{3.40}} & \textbf{\textcolor{red}{5.32}}\\
 \hline
 \end{tabular*}
 \end{center}
 \end{table}

\subsection{Ablation Studies for Interpolation Kernel}
 In this part, we change the interpolation kernel from Laplacian to Gaussian and observe the effects. As shown in Tab.(\ref{tab:kernel_ablation}), we find the Laplacian kernel has better accuracy.
 \begin{table}[h]
\begin{center}
\scriptsize
\caption{\small Ablation studies for interpolation kernel on Middlebury training set of quarter resolution. The \textcolor{red}{first} and \textcolor{blue}{second} bests are in red and blue respectively. \textbf{Our method} in bold. 
All settings are trained on SceneFlow and evaluated on Middlebury training set without fine-tuning.}
\label{tab:kernel_ablation}
\setlength\tabcolsep{5.5pt}
\begin{tabular*}{1.0\textwidth}{l@{\extracolsep{\fill}}cccccc}
\hline
\multirow{2}{*}{Kernel} &  \multirow{2}{*}{Param (M)} & \multirow{2}{*}{Time (s)} & \multicolumn{2}{c}{$>1$px} &\multicolumn{2}{c}{$>2$px}\\
& & & Noc & All & Noc & All\\
\hline
Gaussian & 11.96 & 0.25 & \textcolor{blue}{9.35} & \textcolor{blue}{12.66} & \textcolor{blue}{4.50} & \textcolor{blue}{6.71}\\
\textbf{Laplacian} & \textbf{11.96} & \textbf{0.25} & \textcolor{red}{\textbf{9.32}} & \textcolor{red}{\textbf{12.63}} & \textbf{\textcolor{red}{4.49}} & \textbf{\textcolor{red}{6.70}}\\
 \hline
 \end{tabular*}
 \end{center}
 \end{table}

\subsection{Cross-Domain Generalization}
In this part, we apply our method to ITSA \citep{chuah2022cvpr} only at inference time. We use the pre-trained model provided by ITSA, which is trained on synthetic images. As shown in Tab.(\ref{tab:cross_domain_itsa}), when evaluated on real-world datasets, our method can improve the performance on various networks and benchmarks.

\begin{table}[h]
\begin{center}
\scriptsize
\caption{\small Cross-domain evaluation with ITSA. The \textcolor{red}{first} and \textcolor{blue}{second} bests are in red and blue respectively. 
All methods are trained on SceneFlow and evaluated on Middlebury training set without fine-tuning.}
\label{tab:cross_domain_itsa}
\setlength\tabcolsep{5.5pt}
\begin{tabular*}{1.0\textwidth}{l@{\extracolsep{\fill}}cccccc}
\hline
Backbone &  Training & Test  &  KITTI 2012 & KITTI 2015 & Middlebury & ETH3D\\
\hline
\multirow{2}{*}{ITSA-PSMNet} & Expectation & Expectation & 5.2 & 5.8 & 9.6 & 9.8\\
 & Expectation & L1-Risk & 5.0 & 5.6 & 9.0 & 9.7\\
 \hline
 \multirow{2}{*}{ITSA-GwcNet} & Expectation & Expectation & 4.9 & 5.4 & 9.3 & 7.1\\
 & Expectation & L1-Risk & 4.6 & \textcolor{blue}{5.2} & 8.8 & 7.1\\
 \hline
 \multirow{2}{*}{ITSA-CFNet} & Expectation & Expectation & \textcolor{blue}{4.2} & \textcolor{red}{4.7} & \textcolor{blue}{8.5} & \textcolor{blue}{5.1}\\
 & Expectation & L1-Risk & \textcolor{red}{4.1} &\textcolor{red}{4.7} & \textcolor{red}{8.4} & \textcolor{red}{5.0}\\
 \hline
 \end{tabular*}
 \end{center}
 \end{table}

\subsection{Ablation Studies for Training Using $L^1$ Risk}
In this part, we provide more results on Middlebury using L1-risk minimization both at training and test time on several popular stereo-matching network architectures, demonstrating the usefulness and completeness of our approach to stereo matching problem. The results are shown in Tab.(\ref{tab:training_l1_ablation}).
\begin{table}[h]
\begin{center}
\scriptsize
\caption{\small Ablation studies for $L^1$ risk on Middlebury training set of quarter resolution. The \textcolor{red}{first} and \textcolor{blue}{second} bests are in red and blue respectively. \textbf{Our method} in bold. 
All methods are trained on SceneFlow and evaluated on Middlebury training set without fine-tuning.}
\label{tab:training_l1_ablation}
\setlength\tabcolsep{5.5pt}
\begin{tabular*}{1.0\textwidth}{l@{\extracolsep{\fill}}cccccc}
\hline
\multirow{2}{*}{Backbone} &  \multirow{2}{*}{Training}&\multirow{2}{*}{Test}  & \multicolumn{2}{c}{$>1$px} &\multicolumn{2}{c}{$>2$px}\\
& & & Noc & All & Noc & All\\
\hline
\multirow{2}{*}{PSMNet \citep{chang2018pyramid}} & Expectation & Expectation & \textcolor{blue}{15.42} & \textcolor{blue}{21.01} & \textcolor{blue}{7.53} & \textcolor{blue}{12.17}\\
 & \textbf{L1-Risk} & \textbf{L1-Risk} &\textbf{\textcolor{red}{15.27}} & \textbf{\textcolor{red}{20.67}} & \textbf{\textcolor{red}{7.48}} & \textbf{\textcolor{red}{11.92}}\\
 \hline
 \multirow{2}{*}{GCNet \citep{Kendall_2017_ICCV}} & Expectation & Expectation & \textcolor{blue}{19.93} & \textcolor{blue}{25.72} & \textcolor{blue}{11.15} & \textcolor{blue}{16.12}\\
 & \textbf{L1-Risk} & \textbf{L1-Risk} &\textbf{\textcolor{red}{16.31}} & \textbf{\textcolor{red}{22.19}} & \textbf{\textcolor{red}{8.55}} & \textbf{\textcolor{red}{13.45}}\\
 \hline
 \end{tabular*}
 \end{center}
 \end{table}
 
\section{Qualitative Results}
\begin{figure}[b]
\centering
\subfigure[Image]{
\begin{minipage}[b]{0.23\textwidth}
\includegraphics[width=1.0\textwidth]{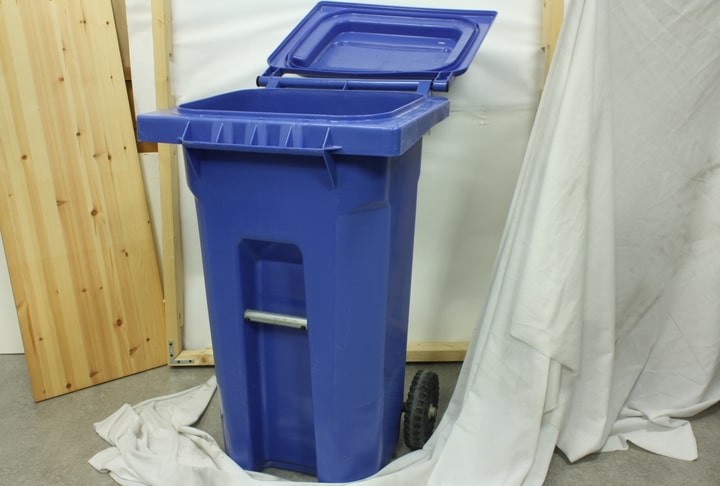}
\includegraphics[width=1.0\textwidth]{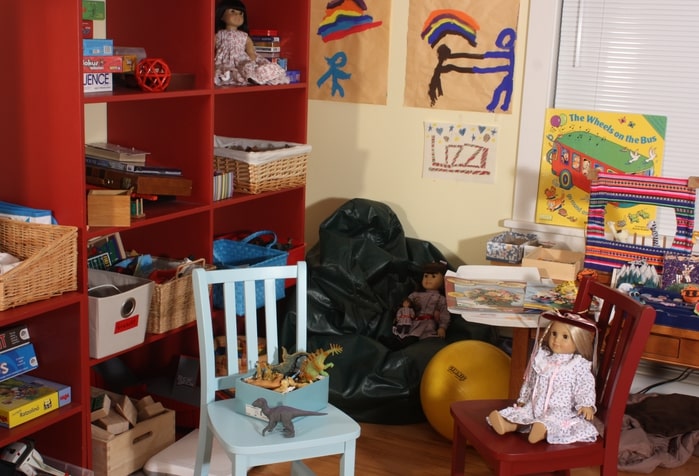}
\includegraphics[width=1.0\textwidth]{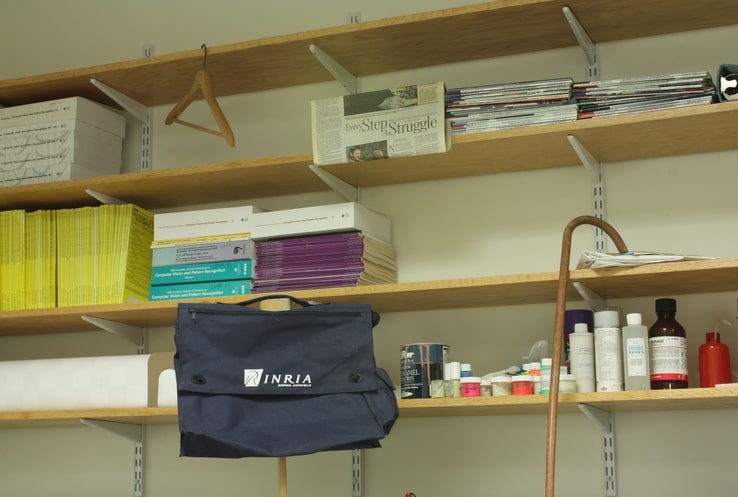}
\end{minipage}
}
\subfigure[IGEV]{
\begin{minipage}[b]{0.23\textwidth}
\includegraphics[width=1.0\textwidth]{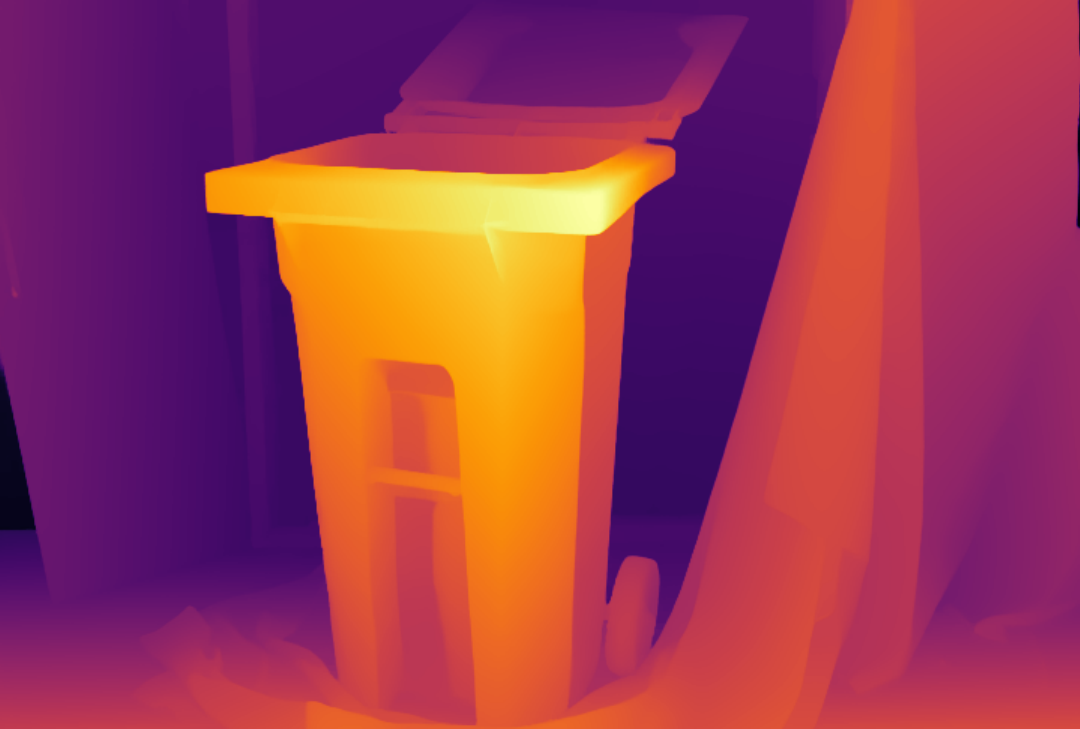}
\includegraphics[width=1.0\textwidth]{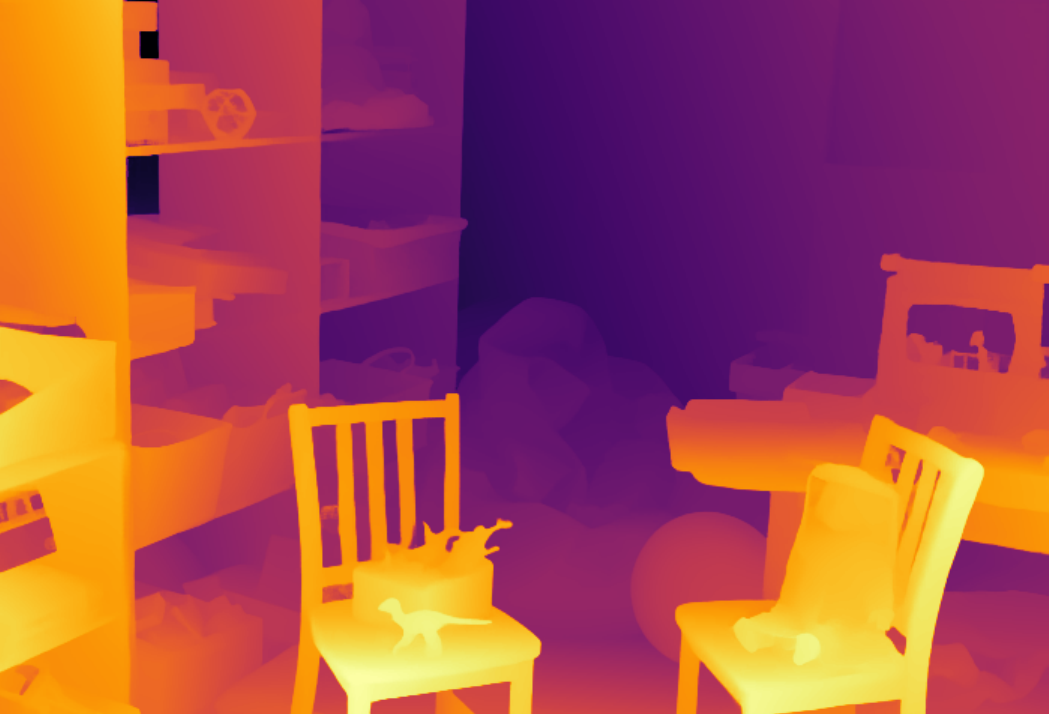}
\includegraphics[width=1.0\textwidth]{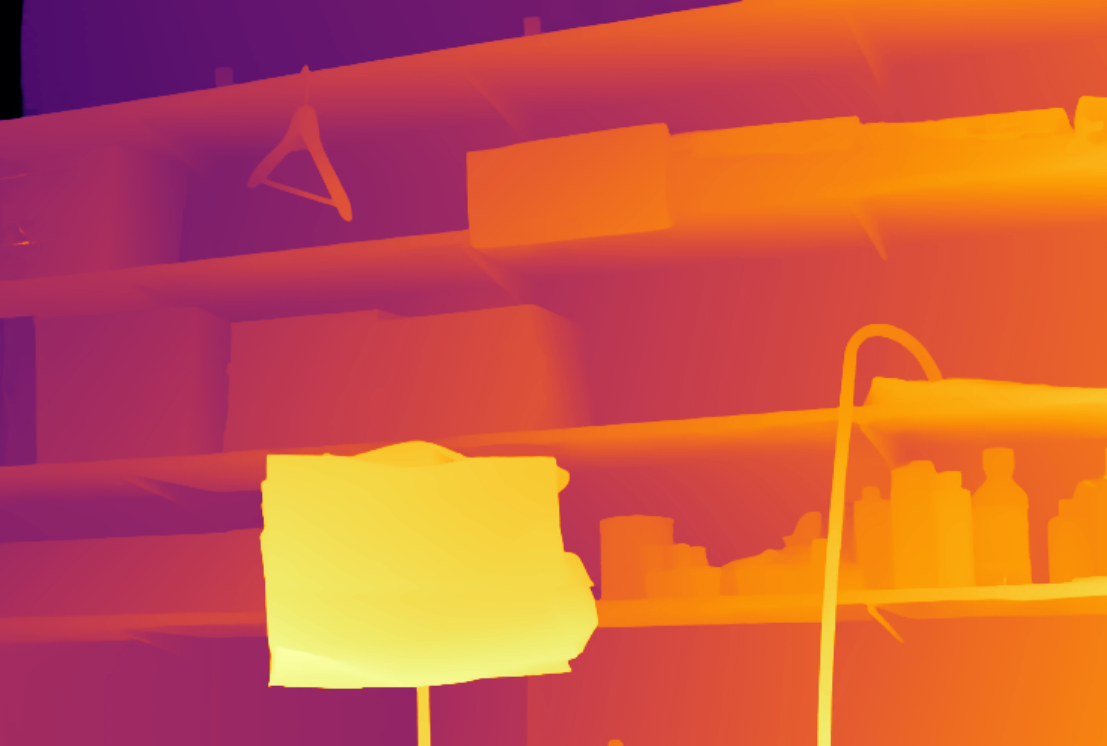}
\end{minipage}
}
\subfigure[DLNR]{
\begin{minipage}[b]{0.23\textwidth}
\includegraphics[width=1.0\textwidth]{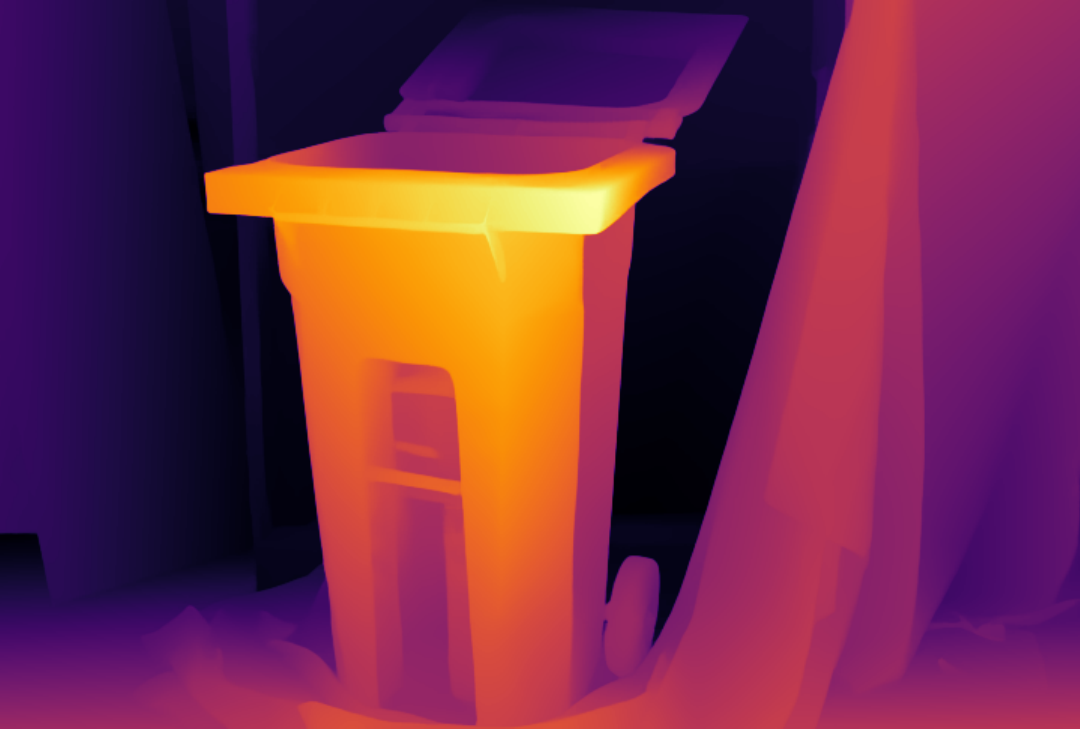}
\includegraphics[width=1.0\textwidth]{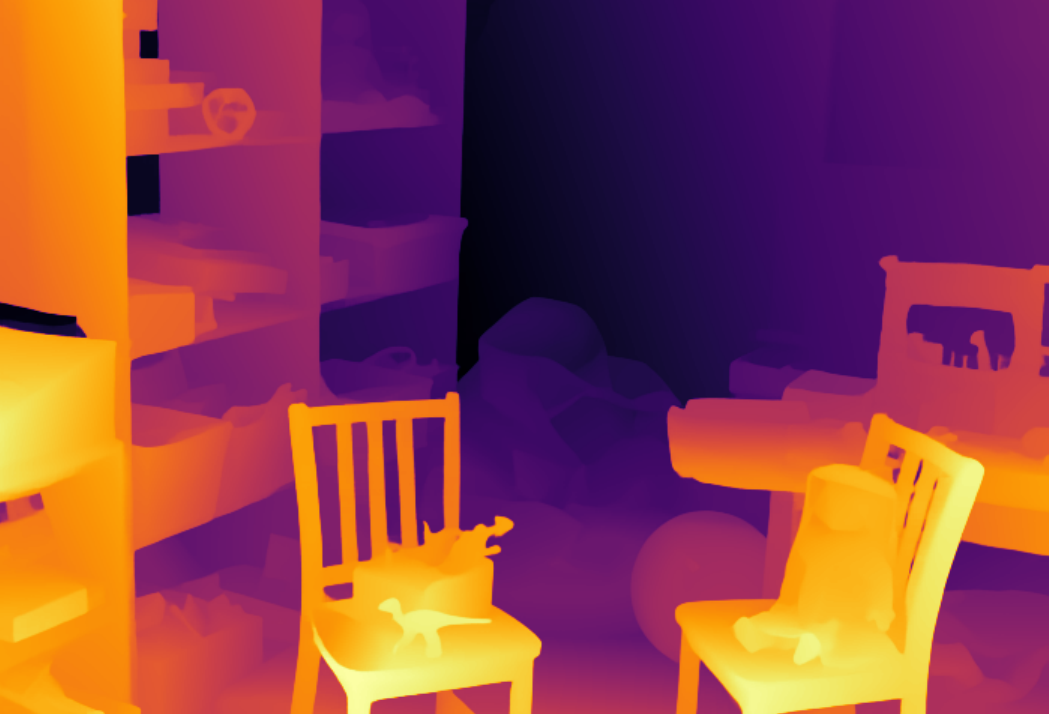}
\includegraphics[width=1.0\textwidth]{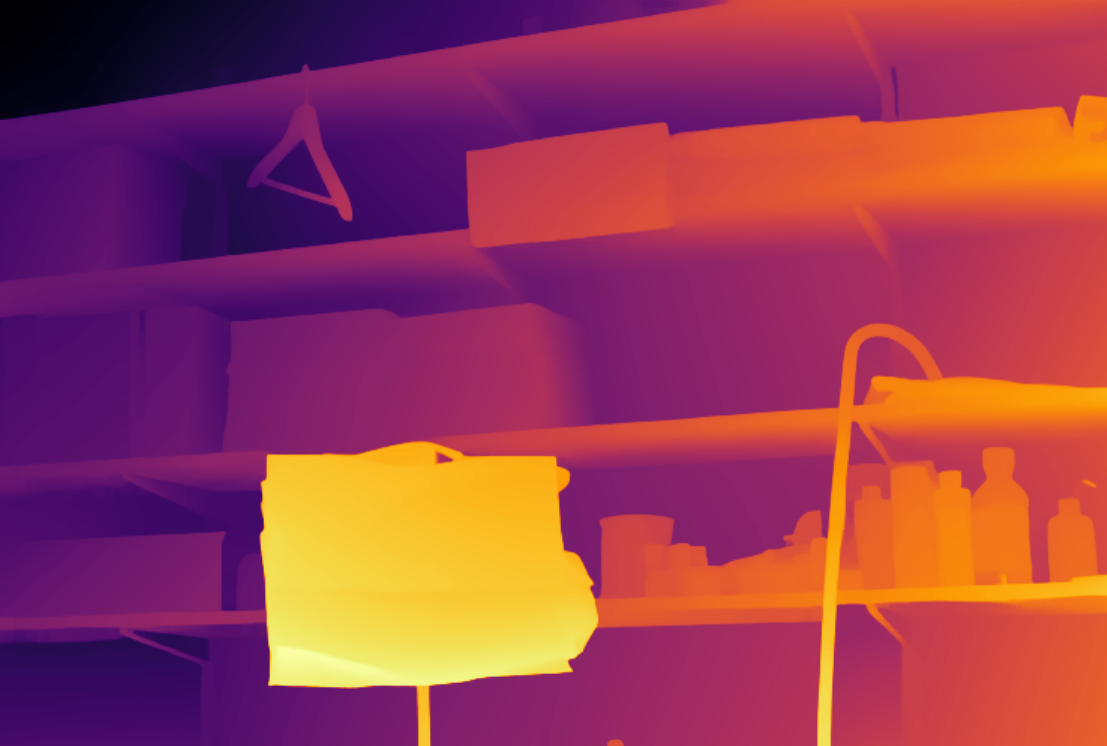}
\end{minipage}
}
\subfigure[Ours]{
\begin{minipage}[b]{0.23\textwidth}
\includegraphics[width=1.0\textwidth]{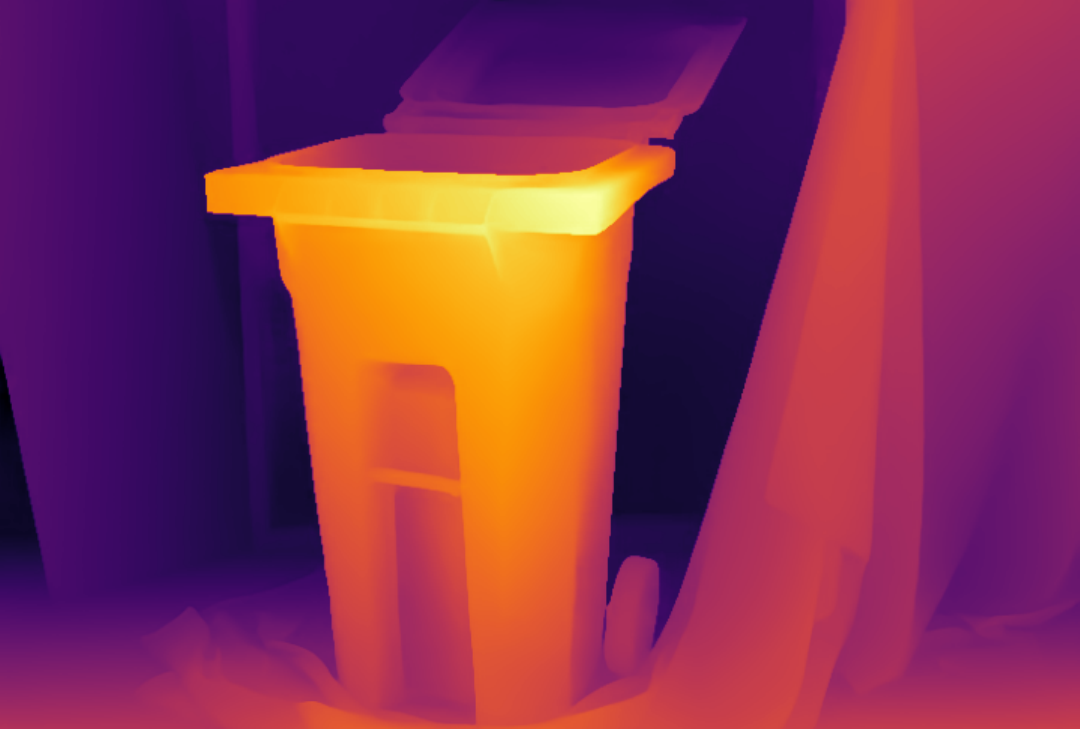}
\includegraphics[width=1.0\textwidth]{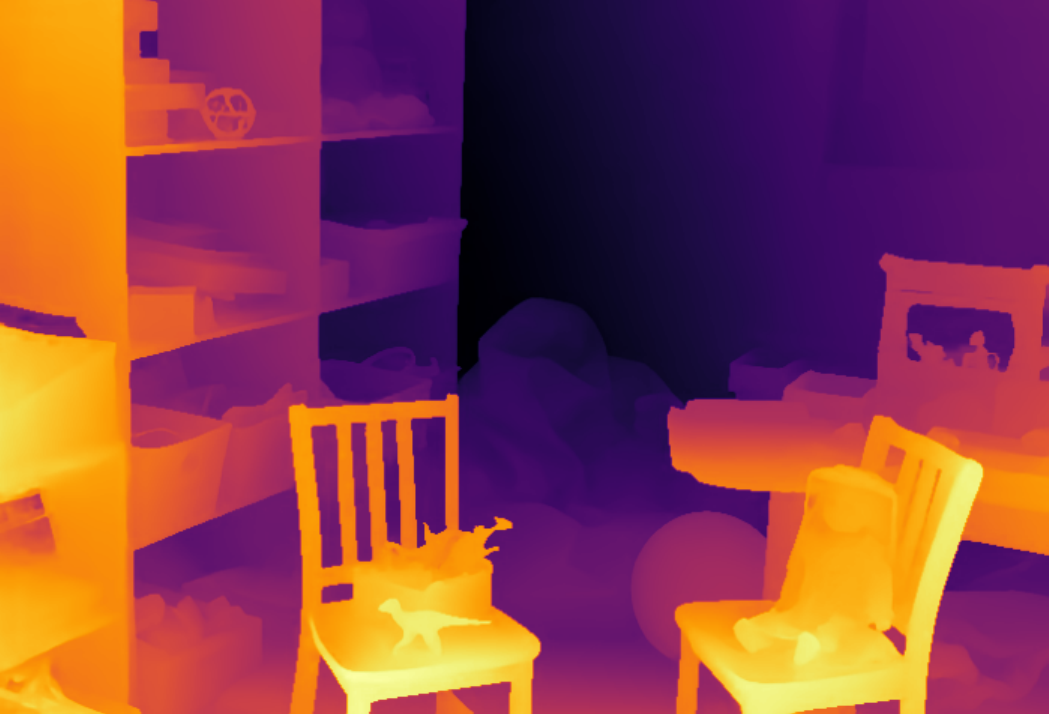}
\includegraphics[width=1.0\textwidth]{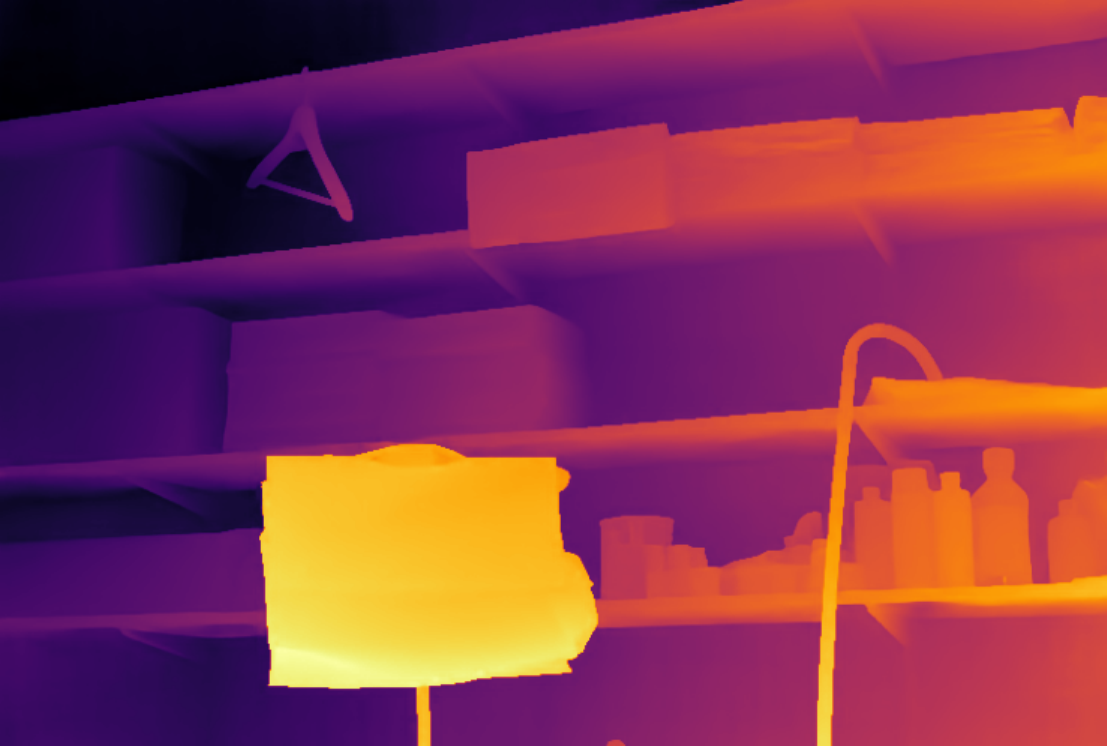}
\end{minipage}
}
\caption{\small \textbf{Qualitative Comparison.} We compare our method with recent state-of-the-art methods such as IGEV~\citep{xu2023iterative}, DLNR~\citep{zhao2023high} on Middlebury~\citep{scharstein2002taxonomy}. All methods are trained only on SceneFlow~\citep{mayer2016large}, and evaluated at quarter resolution.}
\label{fig:sup_qual_mid}
\end{figure}

\begin{figure}
\centering
\subfigure[Image]{
\begin{minipage}[b]{0.23\textwidth}
\includegraphics[width=1.0\textwidth]{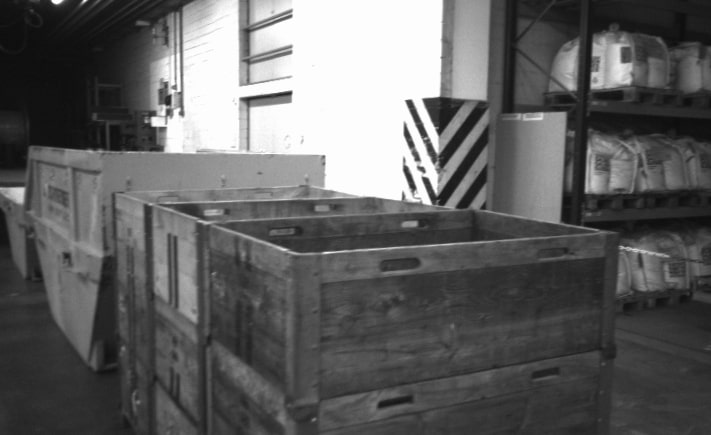}
\includegraphics[width=1.0\textwidth]{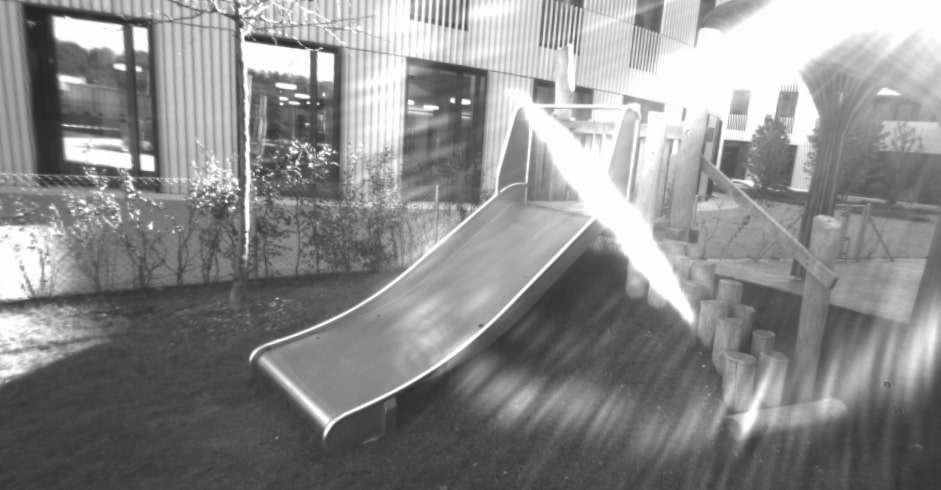}
\includegraphics[width=1.0\textwidth]{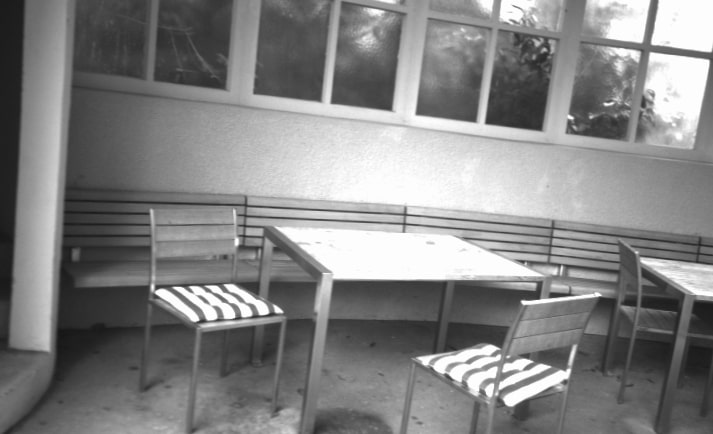}
\end{minipage}
}
\subfigure[IGEV]{
\begin{minipage}[b]{0.23\textwidth}
\includegraphics[width=1.0\textwidth]{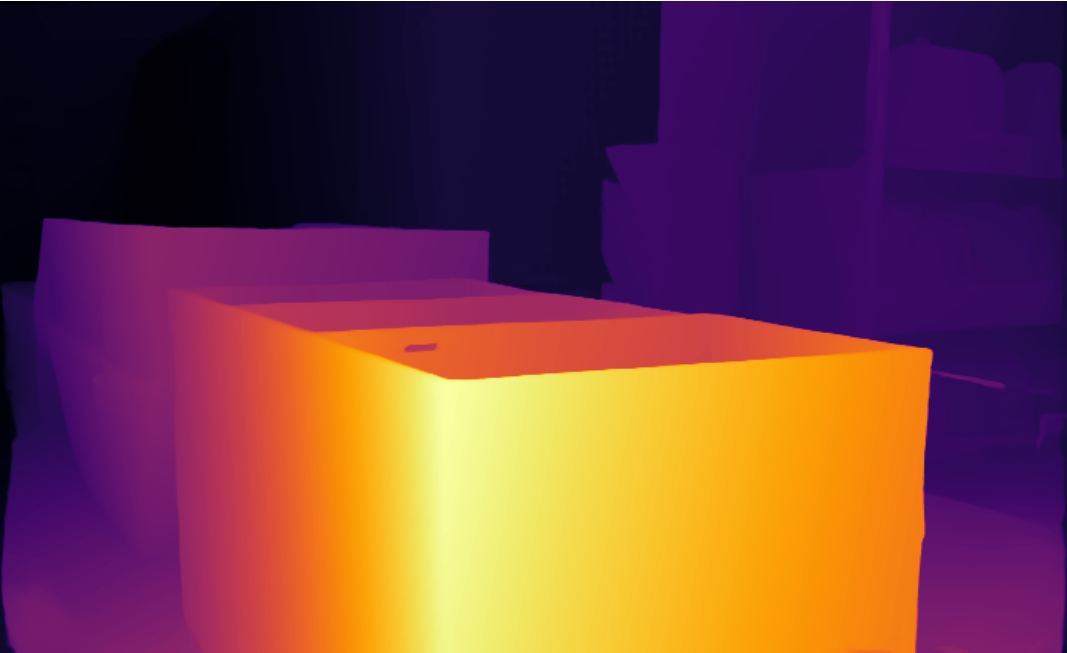}
\includegraphics[width=1.0\textwidth]{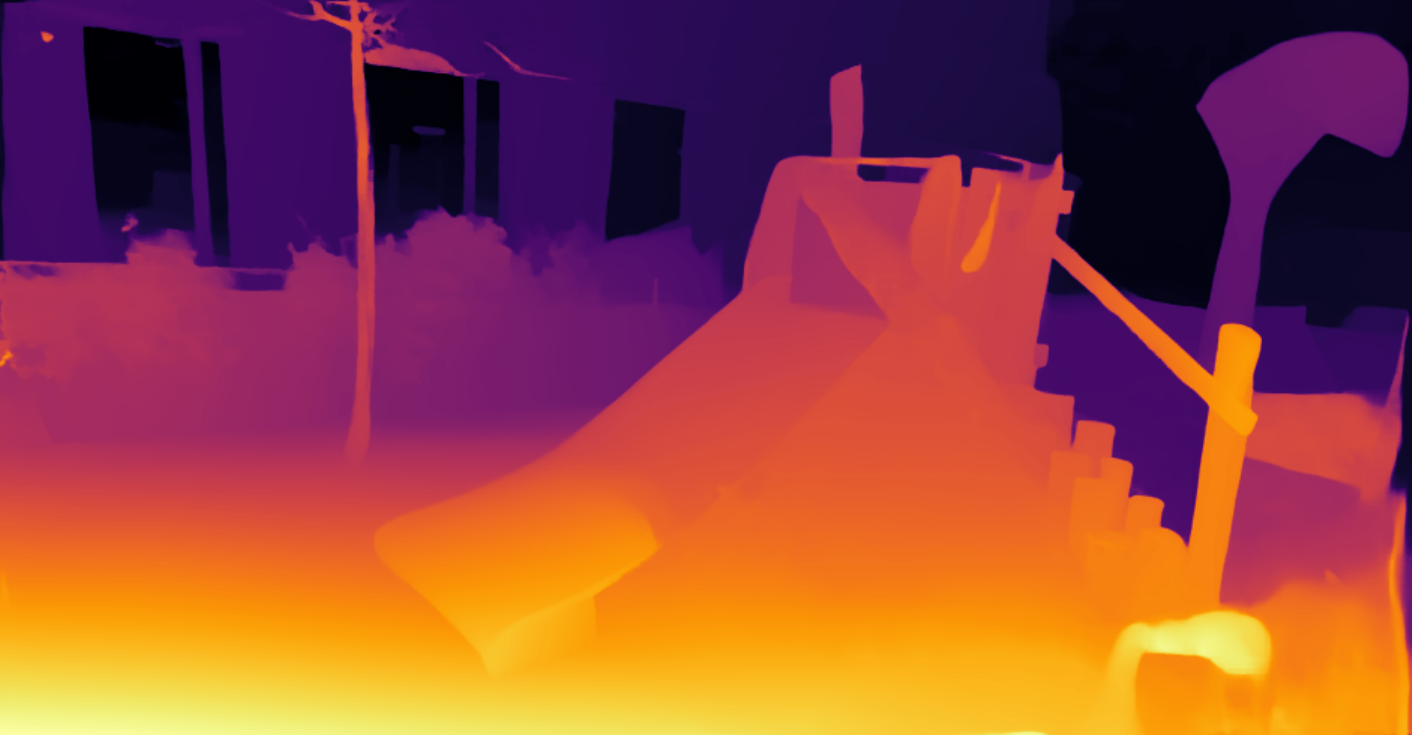}
\includegraphics[width=1.0\textwidth]{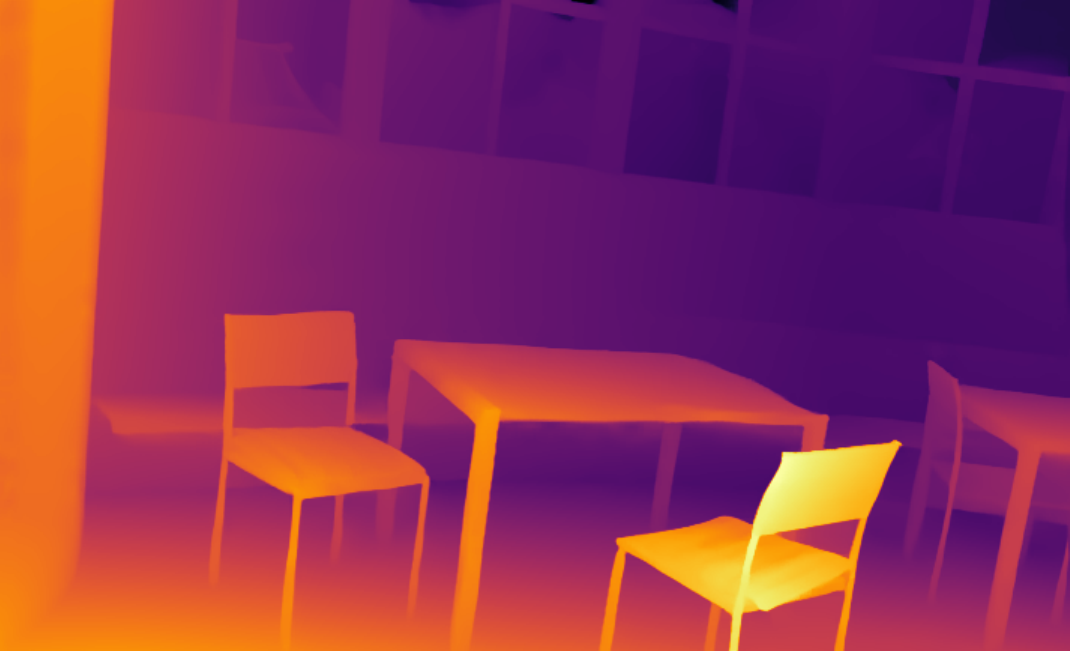}
\end{minipage}
}
\subfigure[PCWNet]{
\begin{minipage}[b]{0.23\textwidth}
\includegraphics[width=1.0\textwidth]{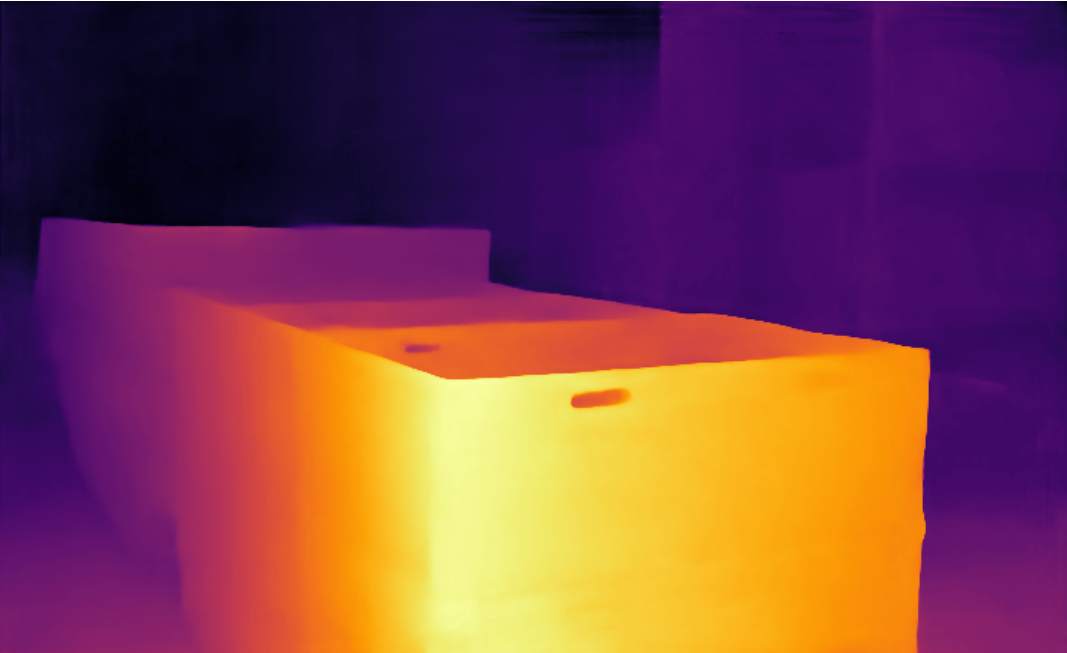}
\includegraphics[width=1.0\textwidth]{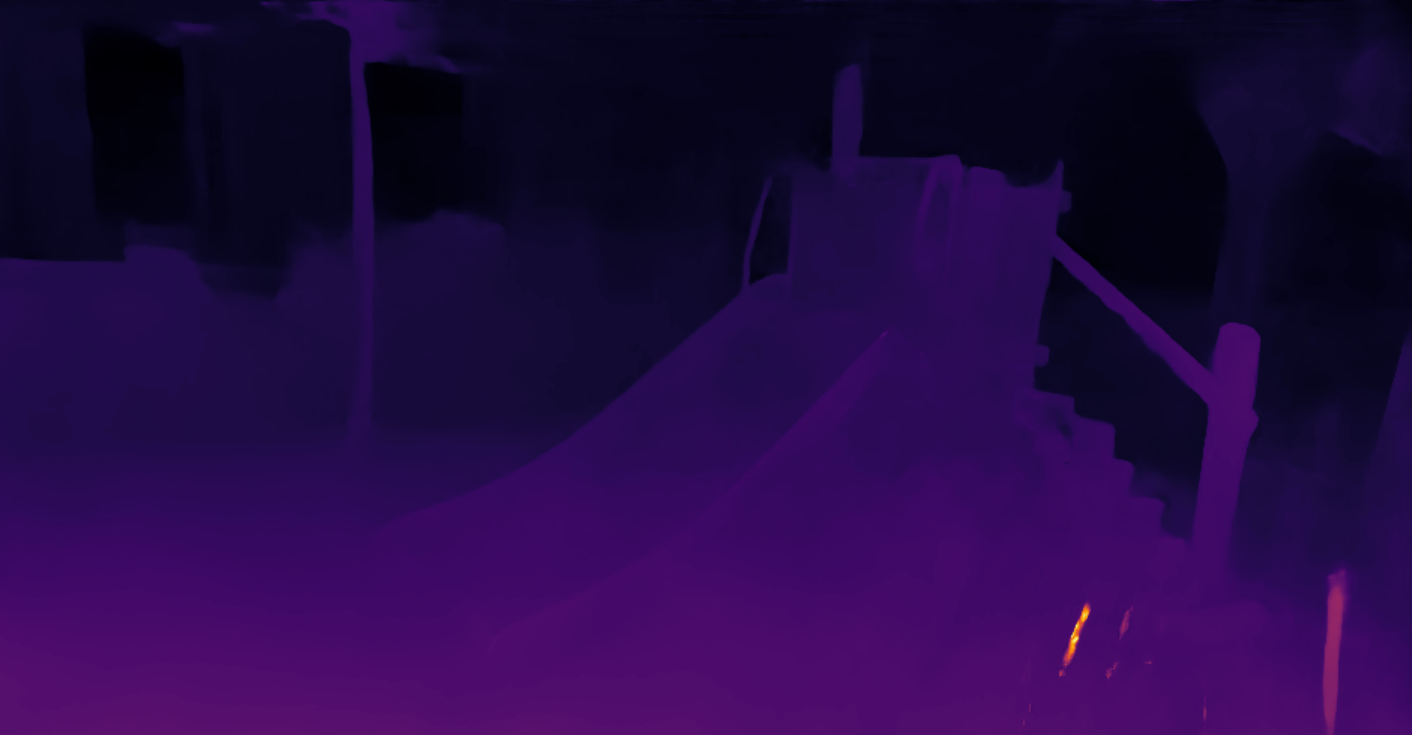}
\includegraphics[width=1.0\textwidth]{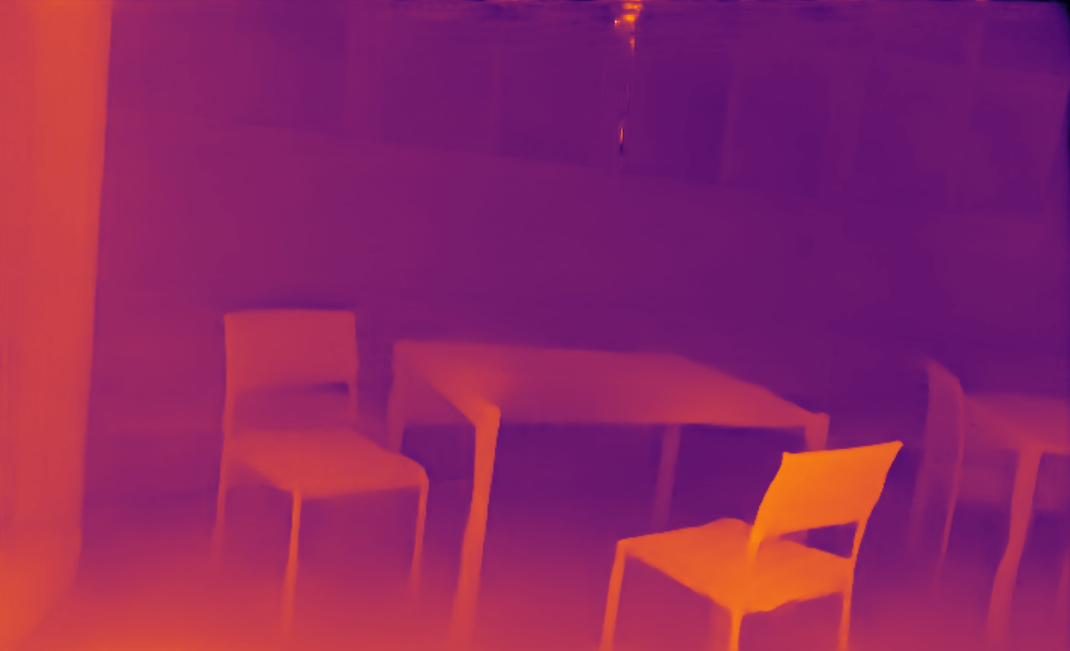}
\end{minipage}
}
\subfigure[Ours]{
\begin{minipage}[b]{0.23\textwidth}
\includegraphics[width=1.0\textwidth]{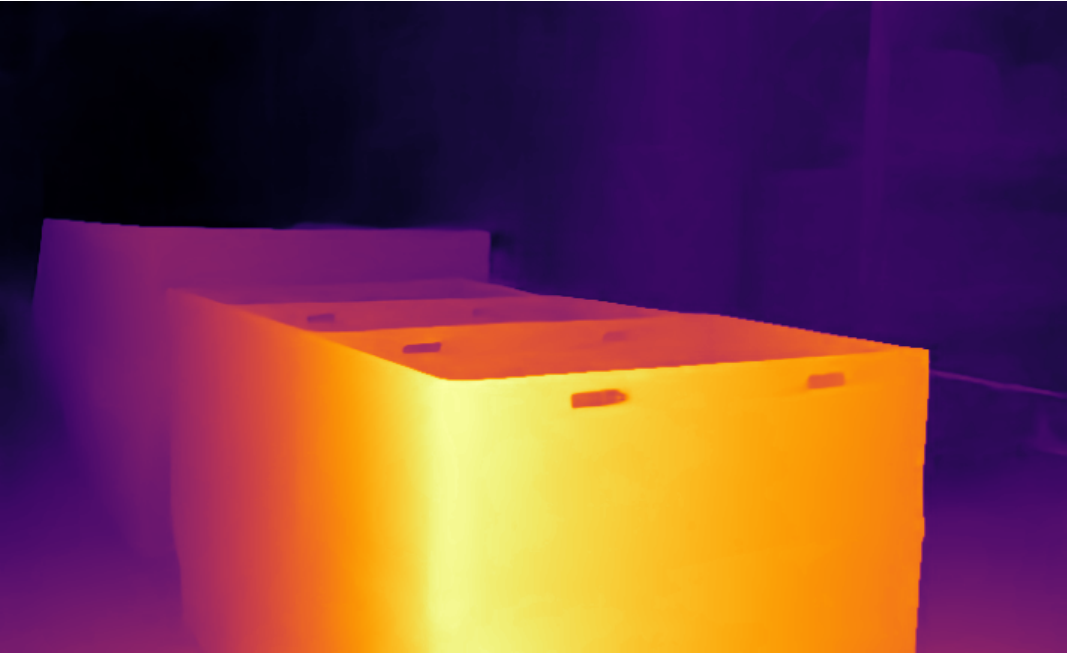}
\includegraphics[width=1.0\textwidth]{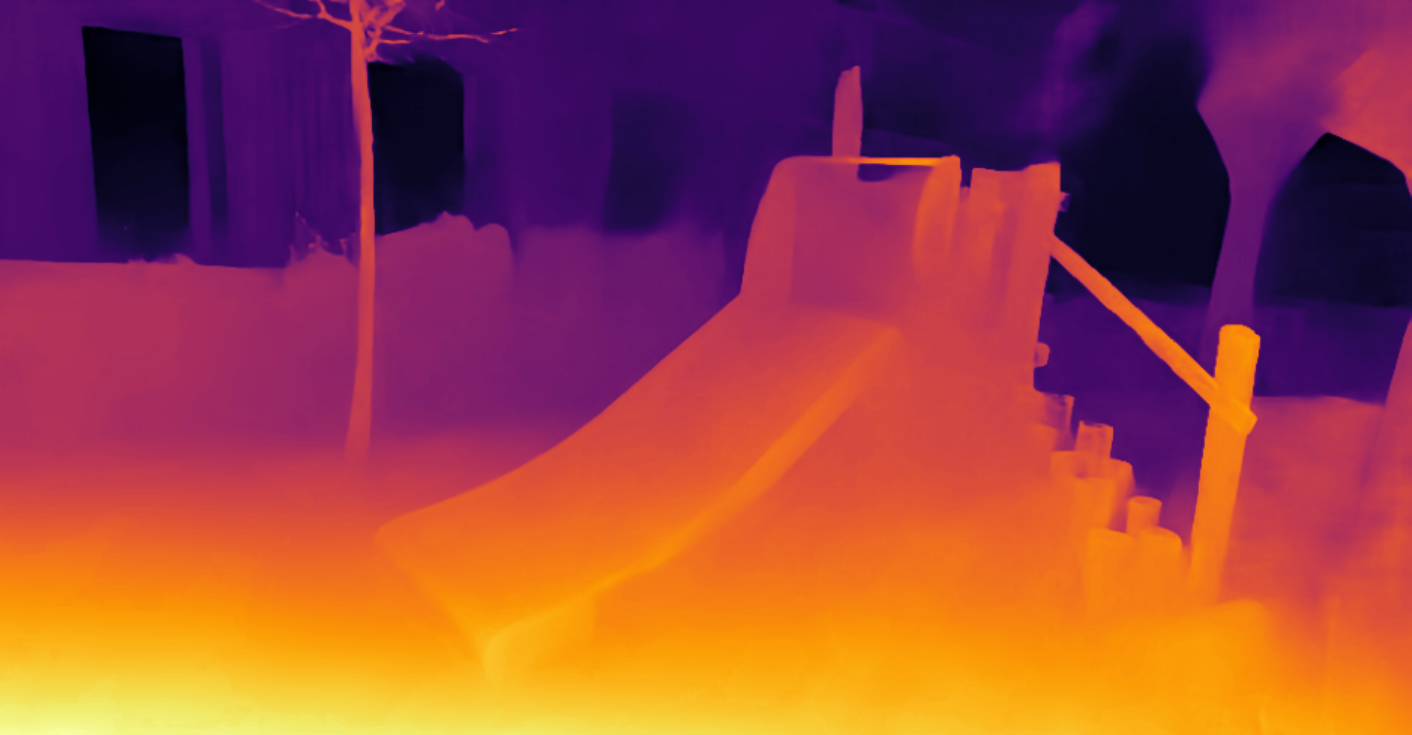}
\includegraphics[width=1.0\textwidth]{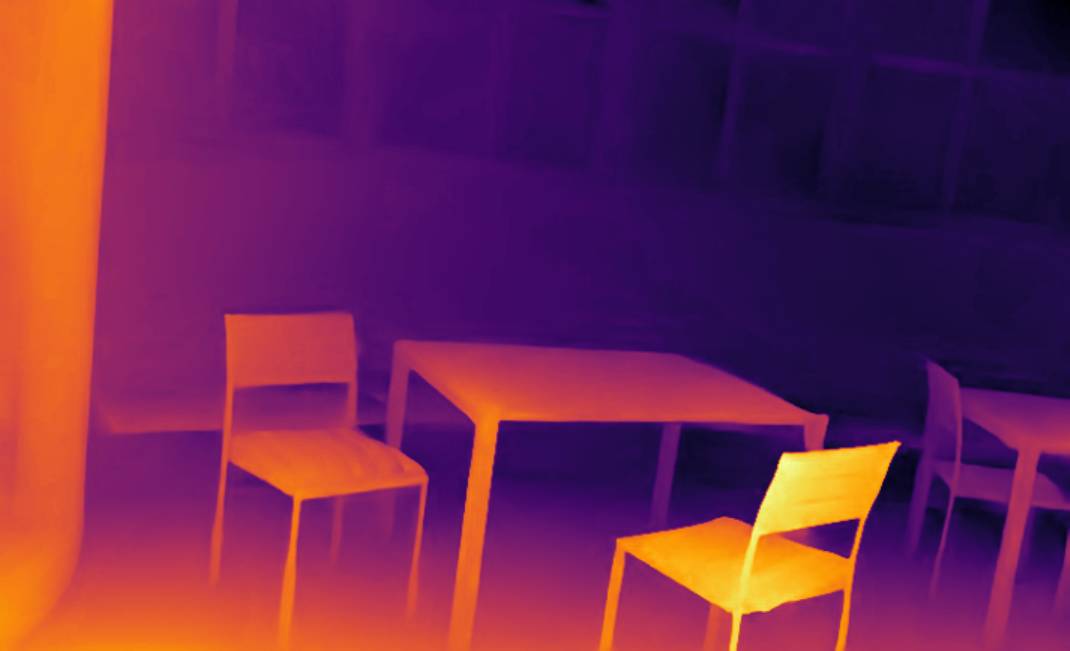}
\end{minipage}
}
\caption{\small \textbf{Qualitative Comparison.} We compare our method with recent state-of-the-art methods such as IGEV~\citep{xu2023iterative}, PCWNet~\citep{shen2022pcw} on ETH 3D~\citep{schoeps2017cvpr}. All methods are trained only on SceneFlow~\citep{mayer2016large}.}
\label{fig:sup_qual_eth}
\end{figure}

\begin{figure}
\centering
\subfigure[Image]{
\begin{minipage}[b]{0.23\textwidth}
\includegraphics[width=1.0\textwidth]{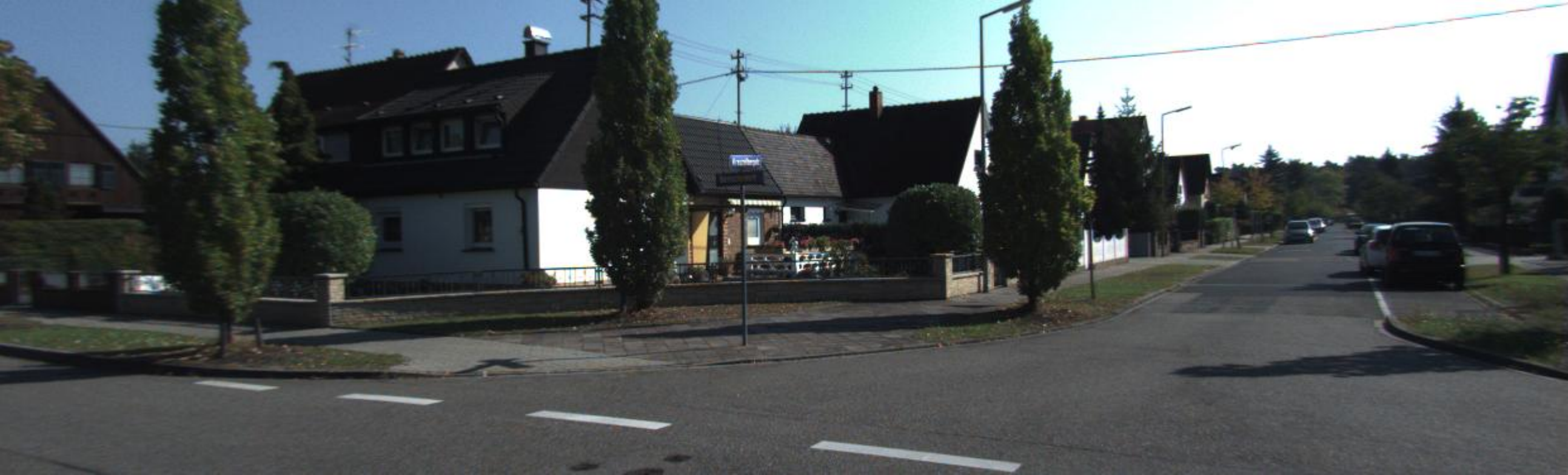}
\includegraphics[width=1.0\textwidth]{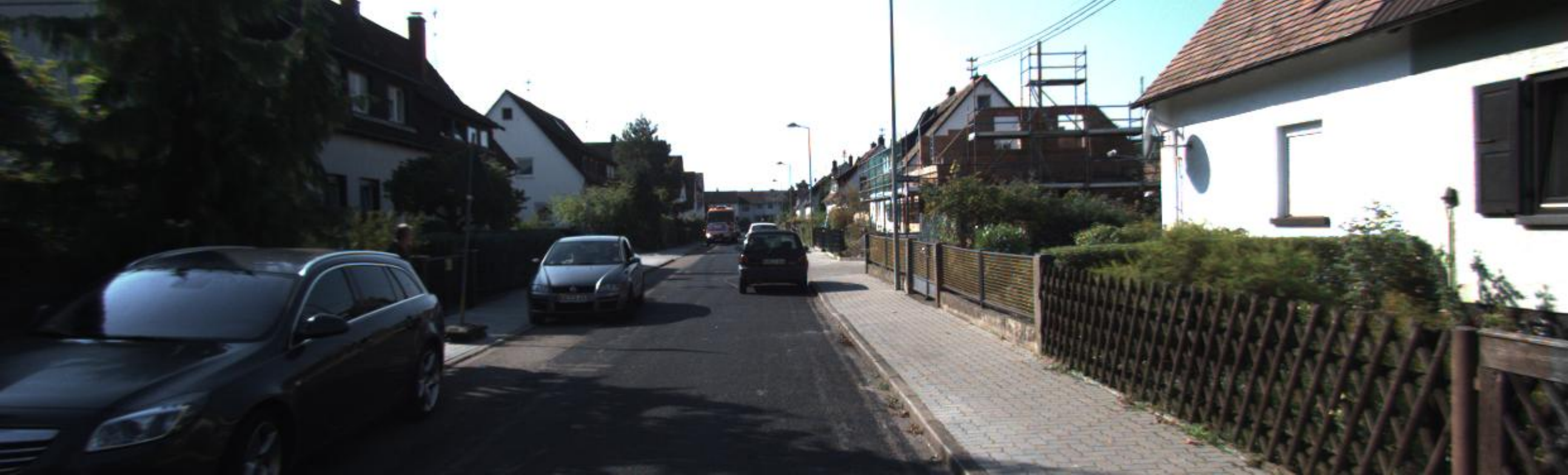}
\includegraphics[width=1.0\textwidth]{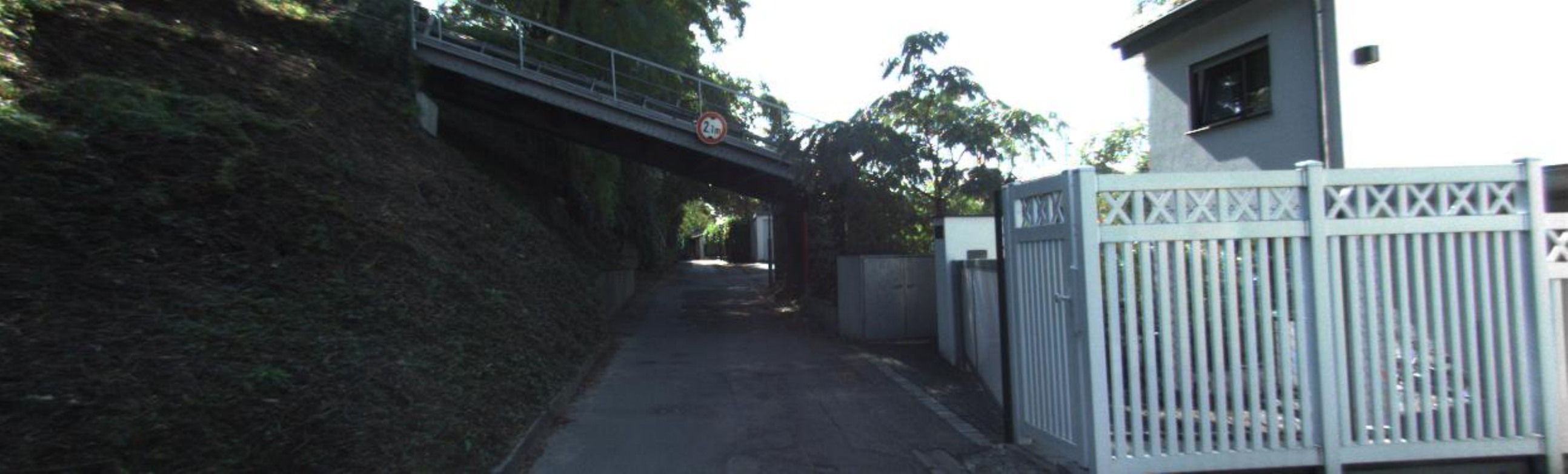}
\includegraphics[width=1.0\textwidth]{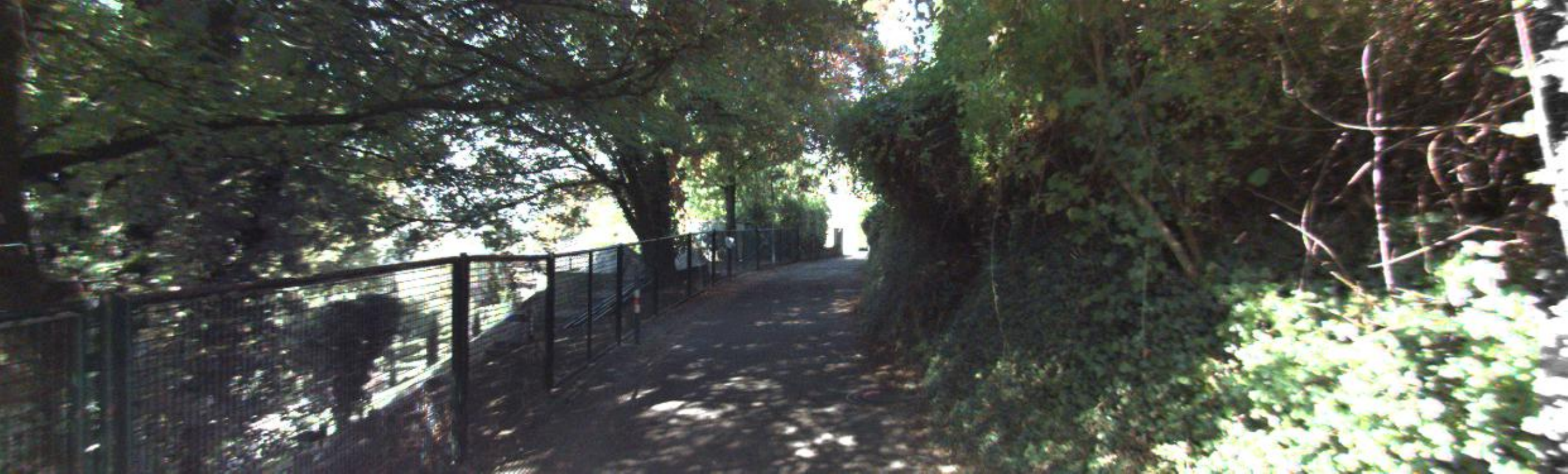}
\end{minipage}
}
\subfigure[IGEV]{
\begin{minipage}[b]{0.23\textwidth}
\includegraphics[width=1.0\textwidth]{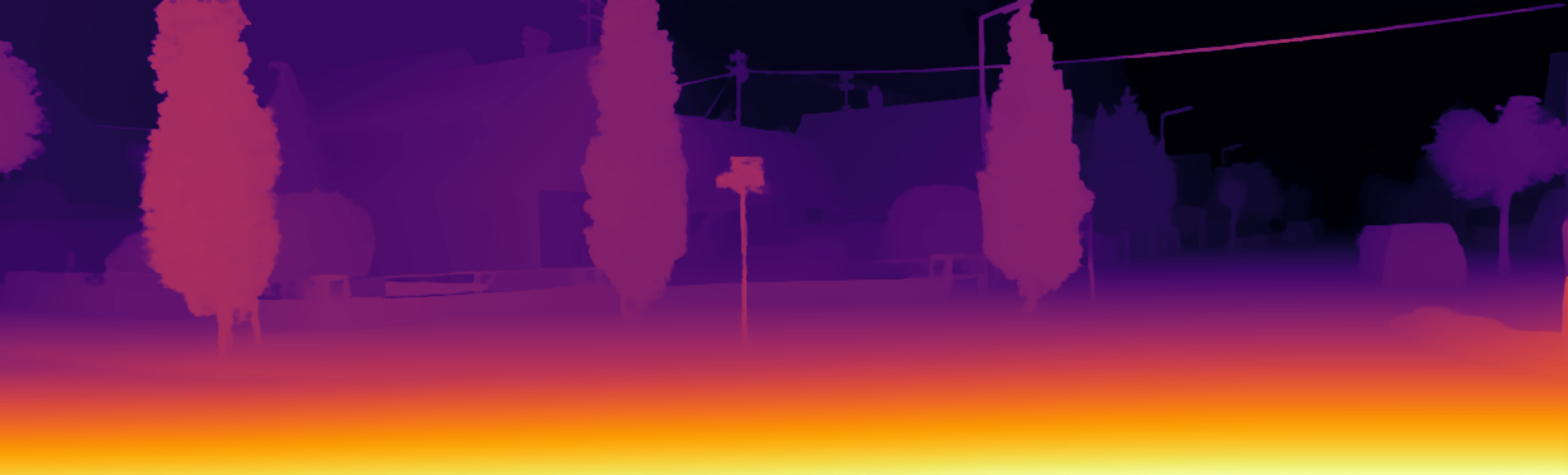}
\includegraphics[width=1.0\textwidth]{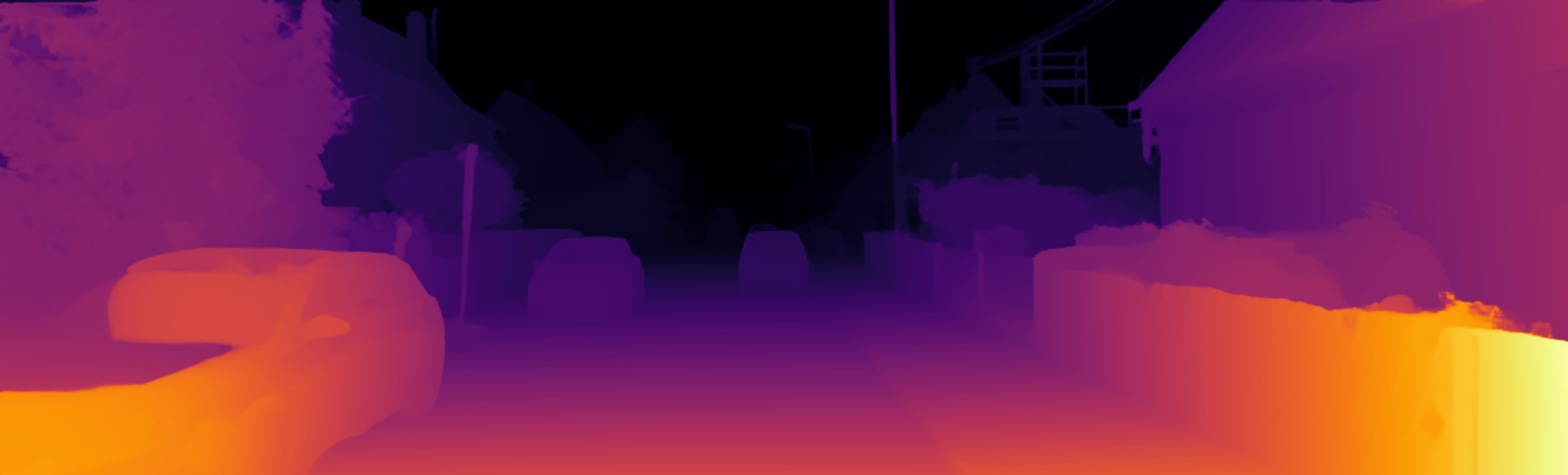}
\includegraphics[width=1.0\textwidth]{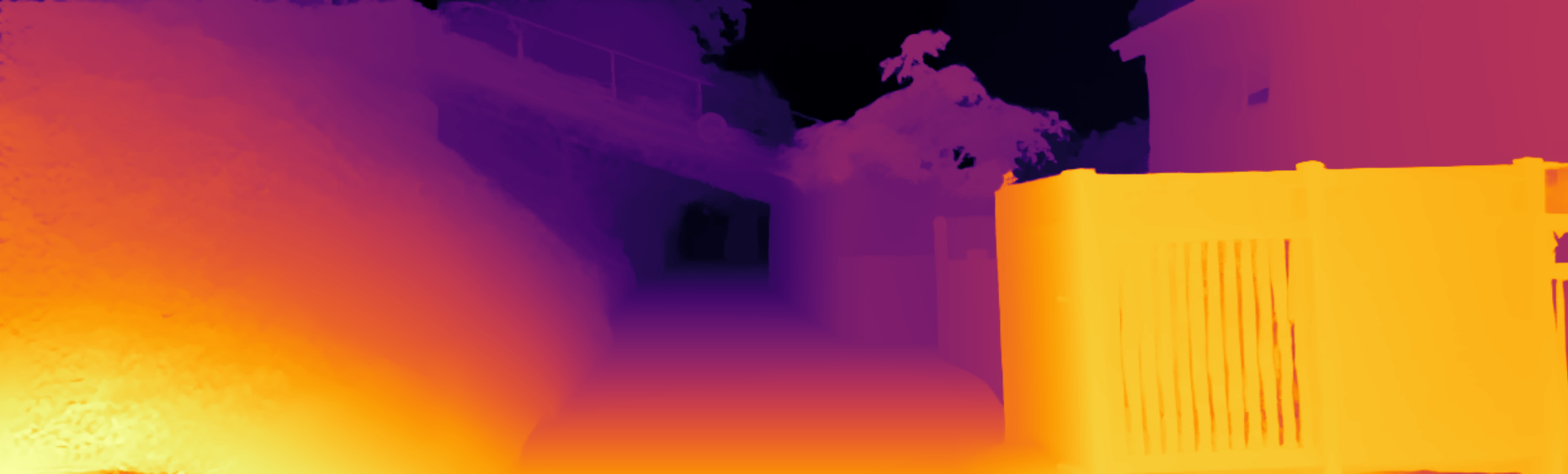}
\includegraphics[width=1.0\textwidth]{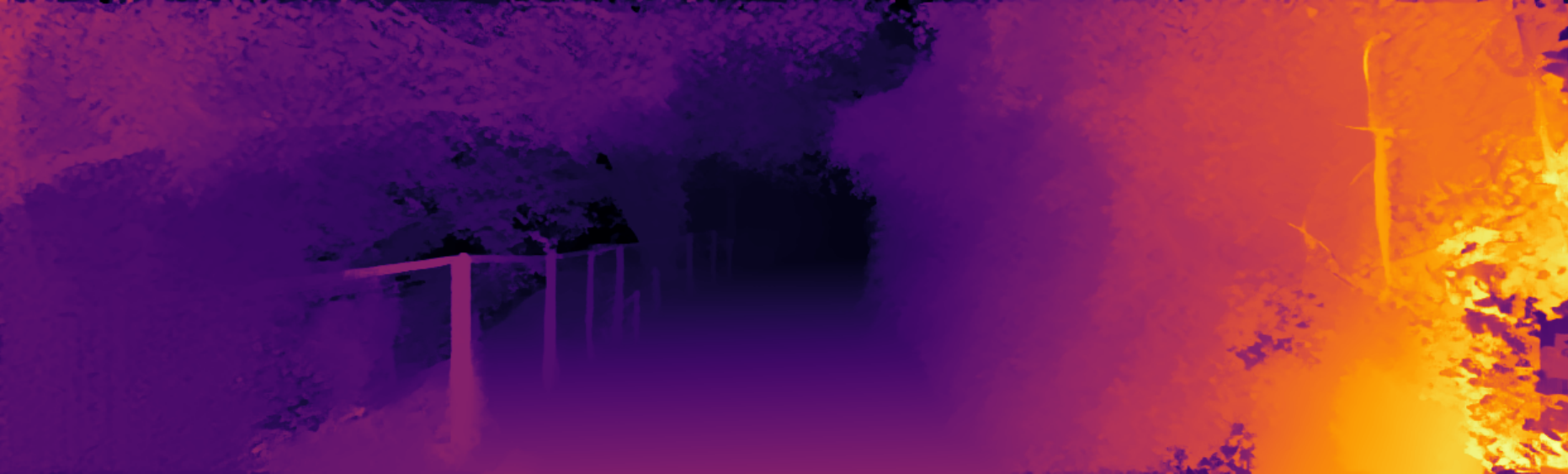}
\end{minipage}
}
\subfigure[PCWNet]{
\begin{minipage}[b]{0.23\textwidth}
\includegraphics[width=1.0\textwidth]{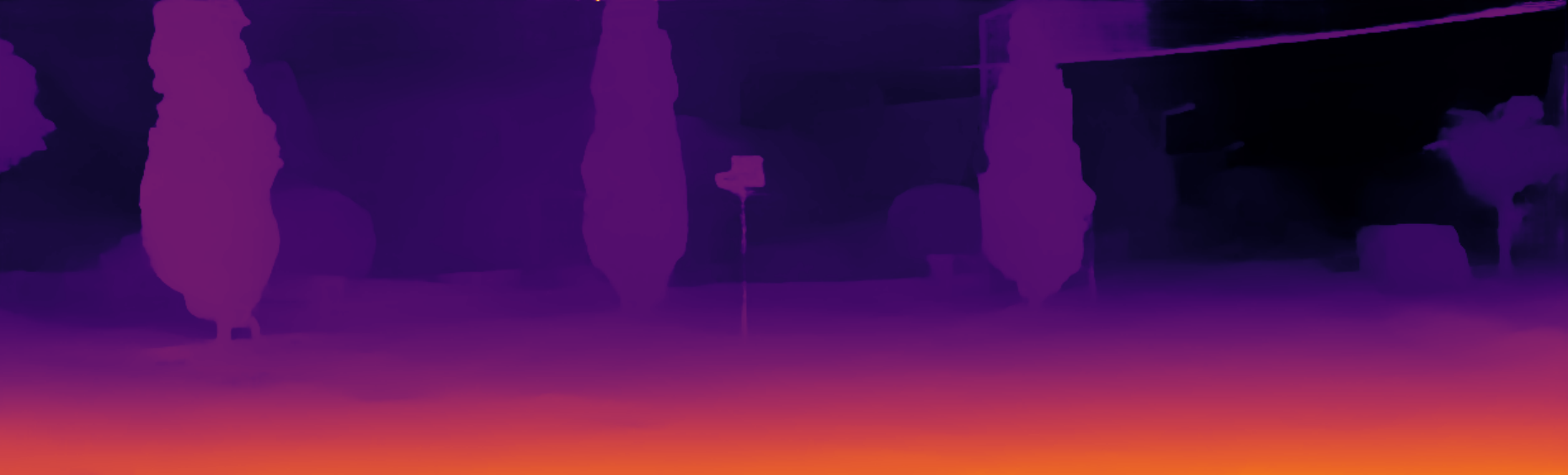}
\includegraphics[width=1.0\textwidth]{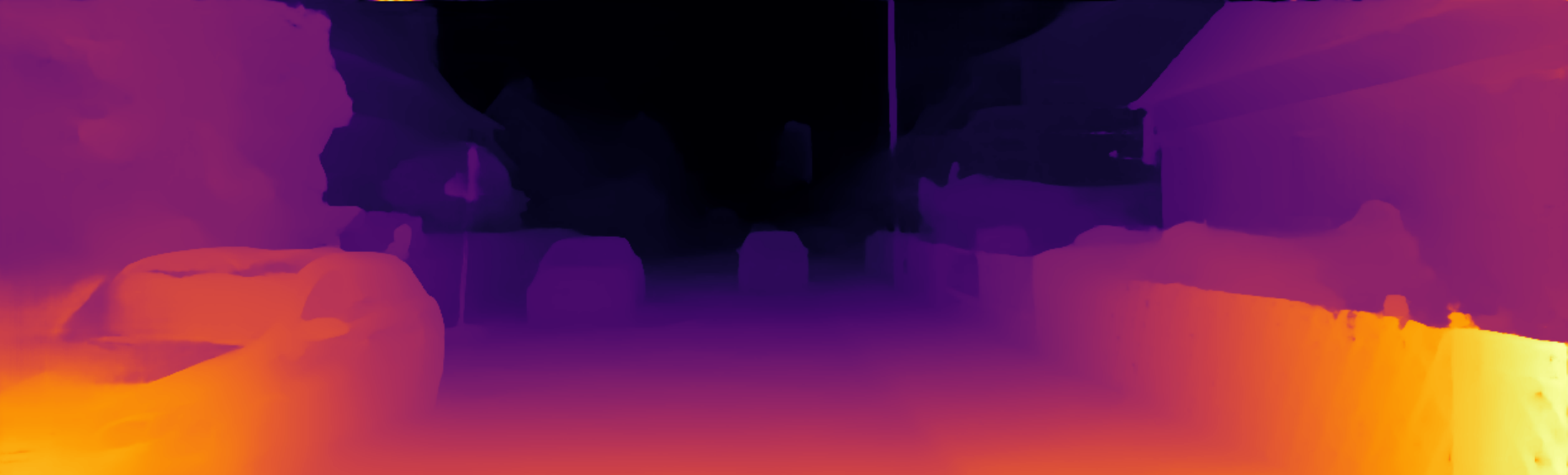}
\includegraphics[width=1.0\textwidth]{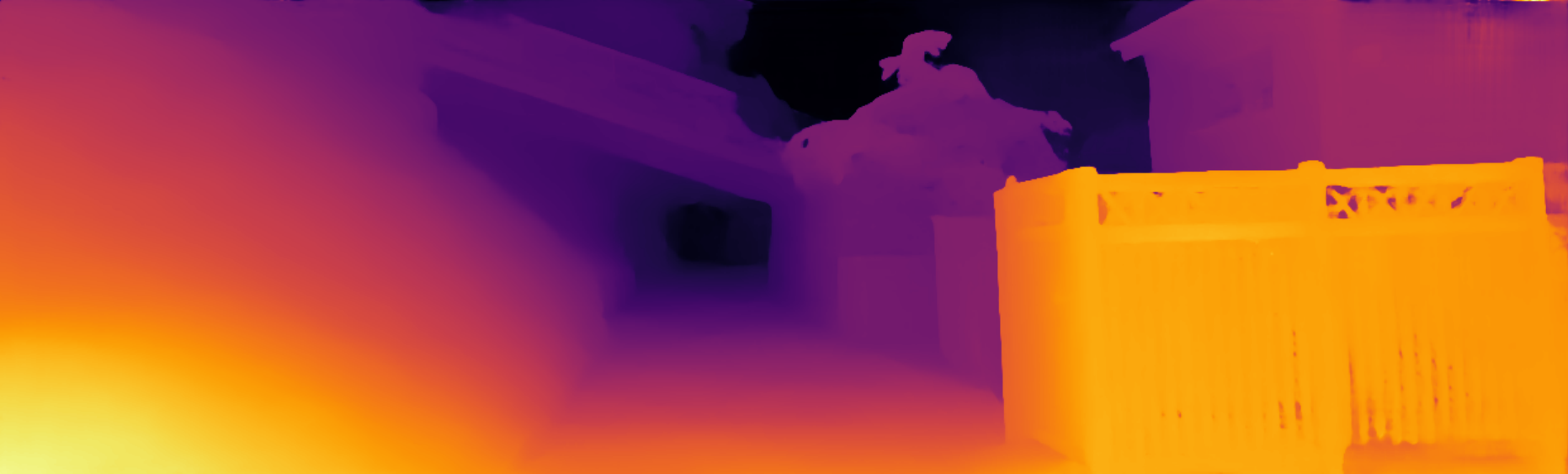}
\includegraphics[width=1.0\textwidth]{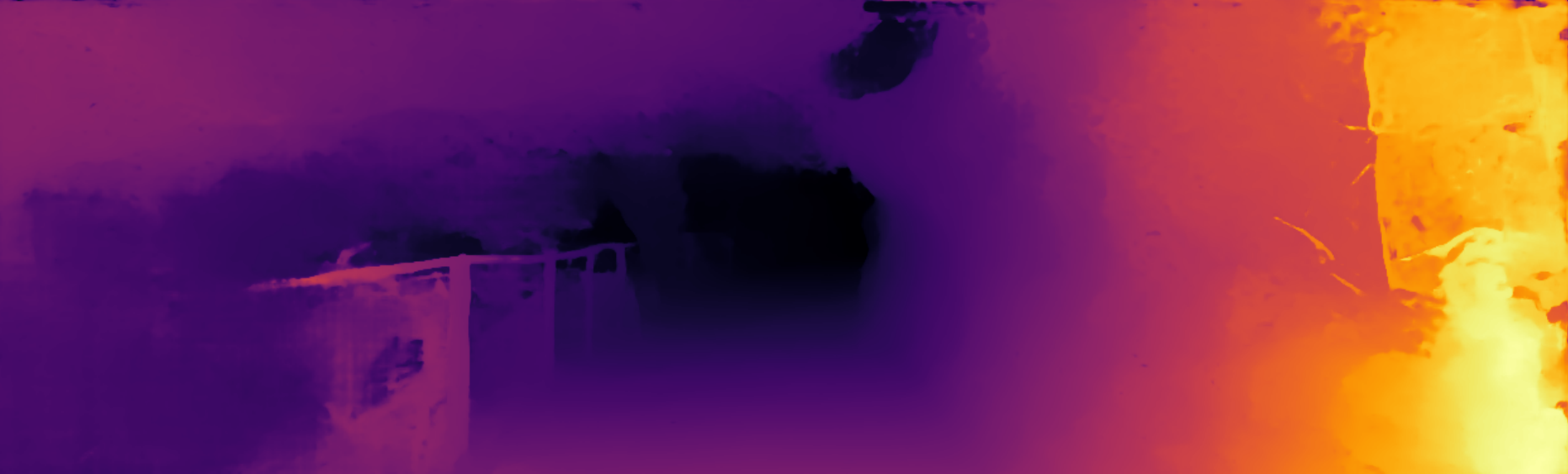}
\end{minipage}
}
\subfigure[Ours]{
\begin{minipage}[b]{0.23\textwidth}
\includegraphics[width=1.0\textwidth]{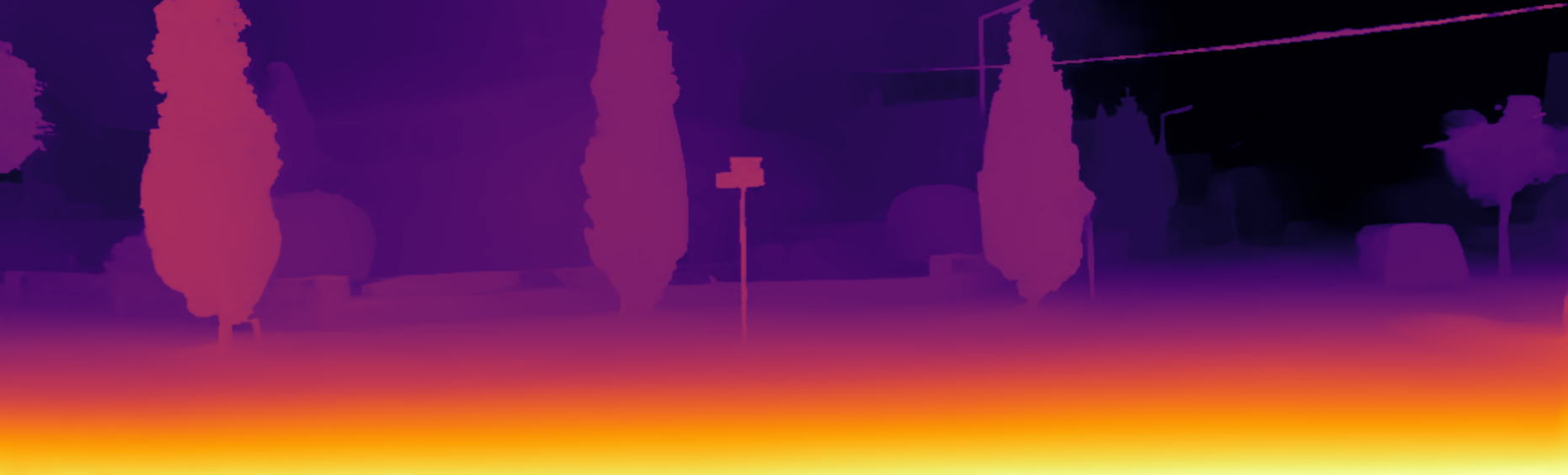}
\includegraphics[width=1.0\textwidth]{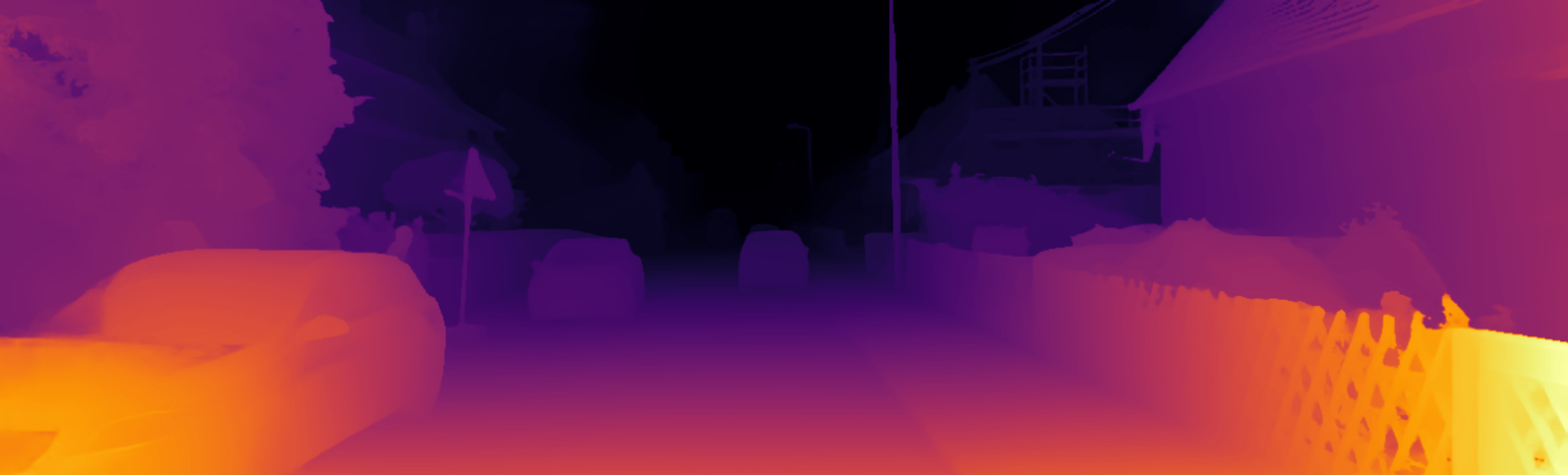}
\includegraphics[width=1.0\textwidth]{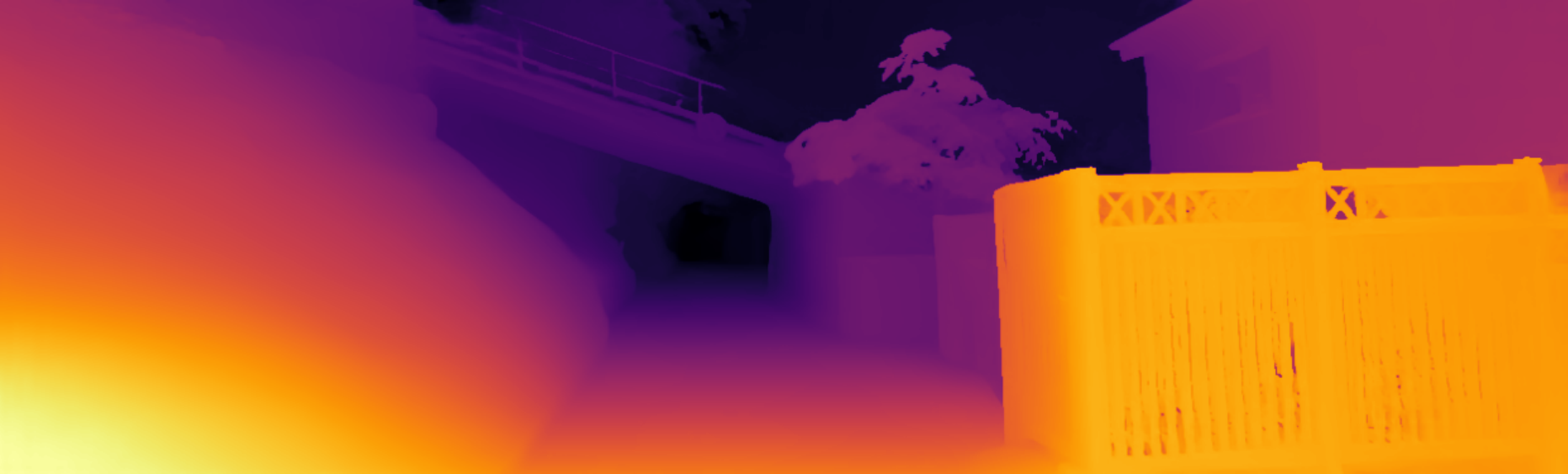}
\includegraphics[width=1.0\textwidth]{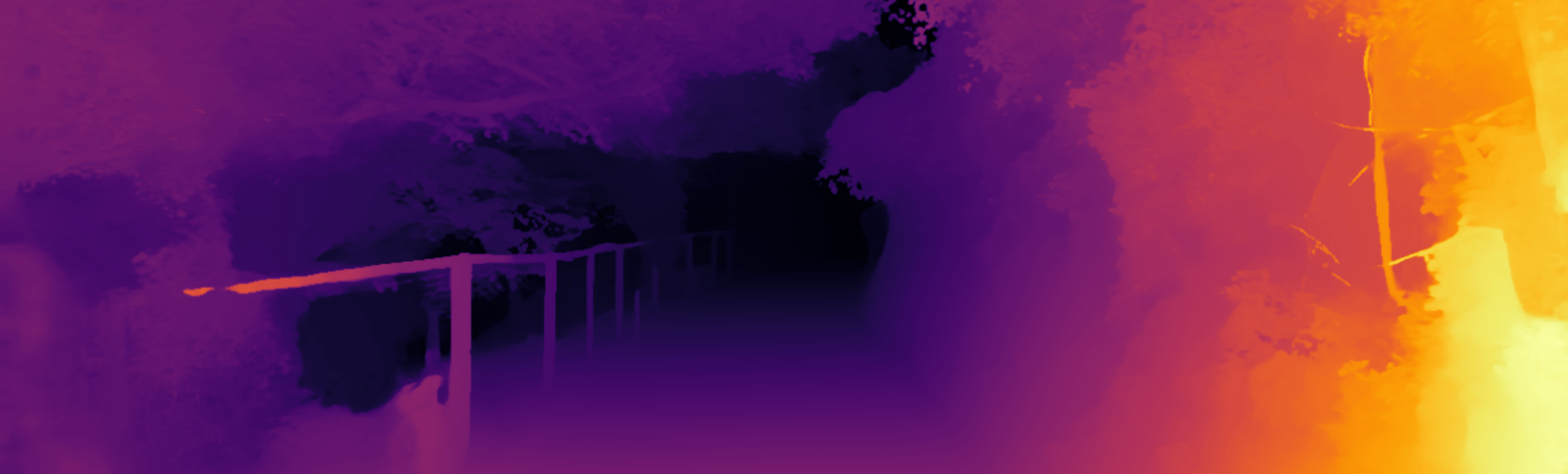}
\end{minipage}
}
\caption{\small \textbf{Qualitative Comparison.} We compare our method with recent state-of-the-art methods such as IGEV~\citep{xu2023iterative}, PCWNet~\citep{shen2022pcw} on KITTI 2012~\citep{geiger2012we}. All methods are trained only on SceneFlow~\citep{mayer2016large}.}
\label{fig:sup_qual_kitti}
\end{figure}

In this section, we present more qualitative results on real-world datasets in Fig. \ref{fig:sup_qual_mid}, Fig. \ref{fig:sup_qual_eth} and Fig. \ref{fig:sup_qual_kitti}. It can be observed that in general our method generalizes and predicts high-frequency details better than other recent methods.

\section{Solution for Squared $L^2$ Norm Loss}
In this part, we present the optimal solution when using the squared $L^2$ norm loss, i.e., $\mathcal{L}(y,x) = (y-x)^2$. 

\begin{equation}
    \mathtt{argmin}_y F(y, \mathbf{p}^m) = \mathtt{argmin}_y \int (y-x)^2 p(x; \mathbf{p}^m) dx.
\end{equation}
Firstly, we found the function $F(y, \mathbf{p}^m)=\int (y-x)^2 p(x; \mathbf{p}^m) dx$ is convex with respect to $y$, because
\begin{align}
\int (\lambda y_1 + (1-\lambda) y_2 -x)^2 p(x; \mathbf{p}^m) dx 
& \leq  \int (\lambda (y_1 -x)^2 + (1-\lambda) (y_2-x)^2) p(x; \mathbf{p}^m) dx \\
& = \lambda \int (y_1 -x)^2 p(x; \mathbf{p}^m) dx + (1-\lambda)  \int (y_2-x)^2 p(x; \mathbf{p}^m) dx
\end{align}
Secondly, the optimal solution for the function $F(y, \mathbf{p}^m)$ can be obtained where $\partial F / \partial y = 0$, i.e., 
\begin{equation}
   \frac{\partial F(y, \mathbf{p}^m)}{\partial y} = 2\int (y-x) p(x; \mathbf{p}^m) dx  = 2y - 2\int x p(x; \mathbf{p}^m) dx = 0
\end{equation}
Therefore the optimal solution is $y=\int x p(x; \mathbf{p}^m) dx$.

\section{Evaluation Metrics}
The definition of evaluation metrics \citep{geiger2012we, menze2015object} is below:\\
D1: Percentage of stereo disparity outliers in first frame.\\
BG: Percentage of outliers averaged only over background regions.\\
FG: Percentage of outliers averaged only over foreground regions.\\
ALL: Percentage of outliers averaged over all ground truth pixels.

\end{document}